\documentclass[jair,twoside,11pt,theapa]{article}
\usepackage{jair, theapa, rawfonts}

\jairheading{}{}{}{}{}
\usepackage{hyperref}       % hyperlinks
\usepackage{url}            % simple URL typesetting
\usepackage{booktabs}       % professional-quality tables
\usepackage{amsfonts}       % blackboard math symbols
\usepackage{nicefrac}       % compact symbols for 1/2, etc.
\usepackage{microtype}      % microtypography

\usepackage{times}
\usepackage{calc}
\usepackage{mathtools}
\usepackage{color}
\usepackage{graphicx} % more modern
\usepackage{subfigure}
\usepackage{algorithm}
\usepackage{algorithmic}
\usepackage{booktabs}
\usepackage{enumitem}
\usepackage{bbm}
\usepackage{multirow}
\usepackage{pifont}
\usepackage{wrapfig}
\usepackage{longtable}
\usepackage{natbib}
\usepackage{amssymb}
\usepackage{kky}
\usepackage{laiDefns}
\usepackage{jlaiDefns}

\ShortHeadings{Multi-fidelity Gaussian Process Bandit Optimisation}
{Kandasamy, Dasarathy, Oliva, Schneider, P\'oczos}
\firstpageno{1}

\begin{document}

\title{Multi-fidelity Gaussian Process Bandit Optimisation}

\newcommand{\cmusymbol}{{\text{\incmtt{1}}}}
\newcommand{\ricesymbol}{{\text{\incmtt{2}}}}
\newcommand{\uncsymbol}{{\text{\incmtt{3}}}}
% \newcommand{\ricesymbol}{\flat}

% \author{\name Kirthevasan Kandasamy$\,^\cmusymbol$  \email kandasamy@cs.cmu.edu \\
%        \addr $^\cmusymbol\,$Machine Learning Department\\
%        Carnegie Mellon University\\
%        Pittsburgh, PA 15217, USA
%        \AND
%        \name Gautam Dasarathy \email gautamd@rice.edu \\
%        \addr Electrical and Computer Engineering\\
%        Rice University\\
%        Houston, TX 77005, USA
%        \AND
%        \name Junier Oliva \email joliva@cs.cmu.edu 
%        \AND
%        \name Jeff Schneider \email schneide.cmu.edu 
%        \AND
%        \name Barnab\'as P\'oczos \email bapoczos@cs.cmu.edu 
%       }
\author{\name Kirthevasan Kandasamy$\,^\cmusymbol$  \email kandasamy@cs.cmu.edu \\
%        \AND
       \name Gautam Dasarathy$\,^\ricesymbol$ \email gautamd@asu.edu \\
%        \AND
       \name Junier Oliva$\,^\uncsymbol$ \email joliva@cs.cmu.edu  \\
%        \AND
       \name Jeff Schneider$\,^\cmusymbol$ \email schneide@cs.cmu.edu  \\
%        \AND
       \name Barnab\'as P\'oczos$\,^\cmusymbol$ \email bapoczos@cs.cmu.edu \\
       \addr \emph{$^\cmusymbol\,$}
       Carnegie Mellon University, 
       Pittsburgh, PA, USA \\
       \addr \emph{$^\ricesymbol\,$}
       Arizona State University,
       Tempe, AZ, USA \\
       \addr \emph{$^\uncsymbol\,$}
       University of North Carolina at Chapel Hill,
       Chapel Hill, NC, USA
      }

\maketitle

\newcommand{\insertAlgoMFGPUCB}{
% \vspace{-0.05in}
\begin{algorithm}
% \textbf{Input: } 
% kernel $\kernel$, bounds $(\zetam)_{m=1}^M$, 
%   parameters $\gammam$.
\textbf{Inputs:}
kernel $\kernel$, bounds $\{\zetam\}_{m=1}^M$, 
  thresholds $\{\gammam\}_{m=1}^M$.
\vspace{-0.05in}
\begin{itemize}[leftmargin=0.2in]
\item For $m=1,\dots,M$:   \vphantom{T} \hspace{0.1in} 
  $\Dcalmmtt{m}{0} \leftarrow \emptyset$, \hspace{0.1in}
%   $(\mummtt{m}{0}, {\sigmammtt{m}{0}}^2) \leftarrow (\zero, \kernel)$.
  $(\mummtt{m}{0}, \sigmammtt{m}{0}) \leftarrow (\zero, \kernel^{1/2})$.
\vspace{-0.03in}
\item for $t = 1, 2,\dots$
\vspace{-0.05in}
  \begin{enumerate}
%   \item $\xt \leftarrow \argmax_{x\in\Xcal} \,\ucbtx$. 
%   \hspace{0.3in}(See Equation~\eqref{eqn:mfgpucbDefn})
%   \item $\mt = \min_m\{\, m\, | \betath\sigmamtmo(\xt) \geq \gammam 
%       \;\;\textrm{or}\;\; m = M\}$.
%       \hspace{0.3in}
%       (See Sections~\ref{sec:analysis},~\ref{sec:implementation} for $\betat$)
  \item $\xt \leftarrow \argmax_{x\in\Xcal} \,\ucbtx$. 
  \hspace{0.5in}\hfill{See~\eqref{eqn:mfgpucbDefn} for $\utilt$ and
                Sections~\ref{sec:analysis},~\ref{sec:implementation} for $\betat$.}
  \item $\mt = \min\,\{ \,m\, | \betath\sigmamtmo(\xt) \geq \gammam 
      \;\;\textrm{or}\;\; m = M\}$.
  \item $\yt \leftarrow$ Query $\funcmm{\mt}$ at $\xt$. 
   \item   Update $\Dcalmtt \leftarrow \Dcalmmtt{\mt}{t-1}\cup\{(\xt,\yt)\}$.
  \hspace{0.05in}
  Obtain $\mumtt,\sigmamtt$ conditioned on $\Dcalmtt$~\eqref{eqn:gpPostEqns}. \newline
  Set $\;\Dcalmmtt{m}{t}\leftarrow\Dcalmmtt{m}{t-1}$,
      $\;\mummtt{m}{t}\leftarrow\mummtt{m}{t-1}$,
      $\;\sigmammtt{m}{t}\leftarrow\sigmammtt{m}{t-1}$
    for $m\neq \mt$.
%   Let $(\Dcalmmtt{m}{t},\mummtt{m}{t},\sigmammtt{m}{t})\leftarrow
%         (\Dcalmmtt{m}{t-1},\mummtt{m}{t-1},\sigmammtt{m}{t-1})$ for $m\neq \mt$.
  \end{enumerate}
% \vspace{-0.1in}
\end{itemize}
% \vspace{-0.05in}
\caption{\hspace{0.05in}\mfgpucb \label{alg:mfgpucb}
% \hspace{0.3in}
}
% \vspace{-0.06in}
\end{algorithm}
}

\newcommand{\insertTmnxTable}{
\newcommand{\diagcolour}{black}
\newcommand{\offdiagboundone}{$\frac{\eta^2}{\gammaonesq}\betan+1$}
\newcommand{\offdiagboundtwo}{$\frac{\eta^2}{\gammatwosq}\betan+1$}
\newcommand{\offdiagboundm}{$\frac{\eta^2}{\gammamsq}\betan+1$}
\newcommand{\lowtribound}{$1$}
\newcommand{\diagbound}{{\frac{5\eta^2}{\Deltamxsq}\betan+1}}
\newcommand{\diagboundone}{{\frac{5\eta^2}{\Deltaonexsq}\betan+1}}
\newcommand{\diagboundtwo}{{\frac{5\eta^2}{\Deltatwoxsq}\betan+1}}
\newcommand{\diagboundM}{{\frac{5\eta^2}{\DeltaMxsq}\betan+1}}
\begin{table}
\centering
\begin{tabular}{|c|c|c|p{0.5cm}cp{0.5cm}|c|}
\hline
 & $\Hcalmm{1}$ & $\Hcalmm{2}$ & & $\Hcalm$ & & $\HcalM\backslash\xoptset$  \\
\hline
$\Tmmttxx{1}{n}{x}$ & $\textcolor{\diagcolour}{\diagboundone}$ &
\offdiagboundone & \dots & \offdiagboundone & \dots &  
 \offdiagboundone   \\
\hline
$\Tmmttxx{2}{n}{x}$ & \multirow{7}{*}{\lowtribound} & 
  $\textcolor{\diagcolour}{\diagboundtwo}$&
\dots & \offdiagboundtwo & \dots &   \offdiagboundtwo  \\
\cline{1-1} \cline{3-7}
$\vdots$ & & &  & $\vdots$ & &  $\vdots$\\
$\Tmnx$ & & \multirow{4}{*}{\lowtribound} & \dots &
  $\textcolor{\diagcolour}{\diagbound}$& \dots & \offdiagboundm \\
$\vdots$ & & &  & $\vdots$ & &  $\vdots$\\
\cline{1-1} \cline{4-7}
$\TMnx$ & & & & \lowtribound & & 
  $\textcolor{\diagcolour}{\diagboundM}$ \\
\hline
\end{tabular}
\caption{
\label{tb:TmnxBounds}
Bounds on the number of queries for each $x\in\Hcalm$ (columns) at each
fidelity (rows). The bound for $\TMnx$ in $\HcalM$ holds for all arms except the
optimal arm $\xopt$ (note $\Deltammxx{M}{\xopt} = 0\,$).
% \vspace{-0.05in}
}
\end{table}
}

% Algorithm gpbbalgo in Appendix 
%%%%%%%%%%%%%%%%%%%%%%%%%%%%%%%%%%%%%%%%%%%%%%%%%%%%%%%%%%%%%%%%%%%%%%%%%%%%%%%%
\newcommand{\insertAlgorithmGPBALGO}{
\begin{algorithm}
\vspace{0.02in}
\textbf{Input: }
kernel $\kernel$.
\vspace{-0.10in}
\begin{itemize}%[label={}]
\item $\Dcal_0 \leftarrow \emptyset$, 
  $(\mu_0, \sigma_0) \leftarrow (\zero, \kernel^{1/2})$.
\vspace{-0.05in}
\item \textbf{for} $t=1, 2, \dots$
\vspace{-0.13in}
\begin{enumerate}
  \item $\xt \leftarrow \argmax_{x\in \Xcal} \mu_{t-1}(x) + 
          \betath\sigma_{t-1}(x)$
  \item $\yt \leftarrow \textrm{Query $\func$ at $\xt$}$.
%   \item $\Dcal_{t} = \Dcal_{t-1} \cup \{ (\xt, \yt)\}$.
  \item Perform Bayesian posterior updates to obtain $\mu_{t},
%     \sigma_{t}$ (See Equation~\eqref{eqn:gpPostEqns}).
    \sigma_{t}$. \hfill See~\eqref{eqn:gpPostEqns}.
\end{enumerate}
% \vspace{-0.15in}
\end{itemize}
\caption{$\;$\gpucb \label{alg:gpucb}\hfill\citep{srinivas10gpbandits}$\quad$}
\end{algorithm}
}

\newcommand{\imarrwthree}{1.855in}
\newcommand{\imhspthree}{-0.05in}
\newcommand{\imarrwtwo}{2.81in}
\newcommand{\imhsptwo}{0.15in}
\newcommand{\imtextspace}{-0.10in}
\newcommand{\imcaptionspace}{-0.2in}

\newcommand{\simulationText}{
Illustration of \mfgpucbs for a $2$-fidelity problem initialised with
$5$ random points at the first fidelity.
In the top figures,
the solid lines in brown and blue are $\funcone,\functwo$ respectively, and the 
dashed lines are $\ucbonet,\ucbtwot$. 
The solid green line is $\ucb_{t}=\min(\ucbonet,\ucbtwot)$.
The small crosses are queries from $1$ to $t-1$ and the red star is the
maximiser of $\ucbt$, i.e. the next query $\xt$.
$\xopt$, the optimum of $\functwo$ is shown in magenta.
In the bottom figures, the solid orange line is $\betath\sigmaonetmo$ and 
the dashed black line is $\gammaone$. 
When $\betath\sigmaonetmo(\xt) \leq\gammaone$ we play at fidelity $\mt=2$ and otherwise at
$\mt=1$.
}

\newcommand{\imSimWidth}{2.84in}
\newcommand{\imSimHspace}{0.02in}
\newcommand{\imSimVspace}{-0.10in}
\newcommand{\imSimVGap}{0.12in}

\newcommand{\insertSimulationVLong}{
\begin{figure*}
\centering
\subfigure{
  \includegraphics[width=\imSimWidth]{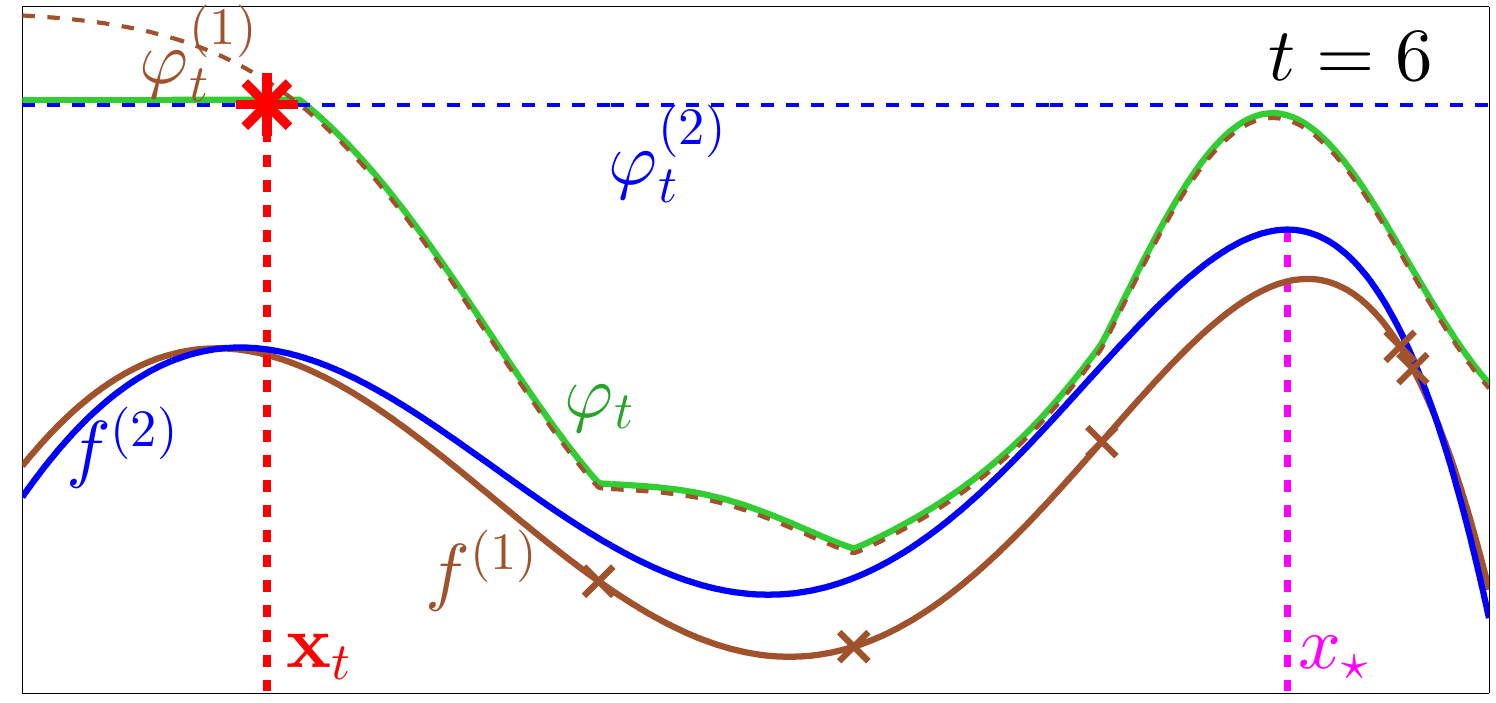} \hspace{\imSimHspace}
}
\subfigure{
  \includegraphics[width=\imSimWidth]{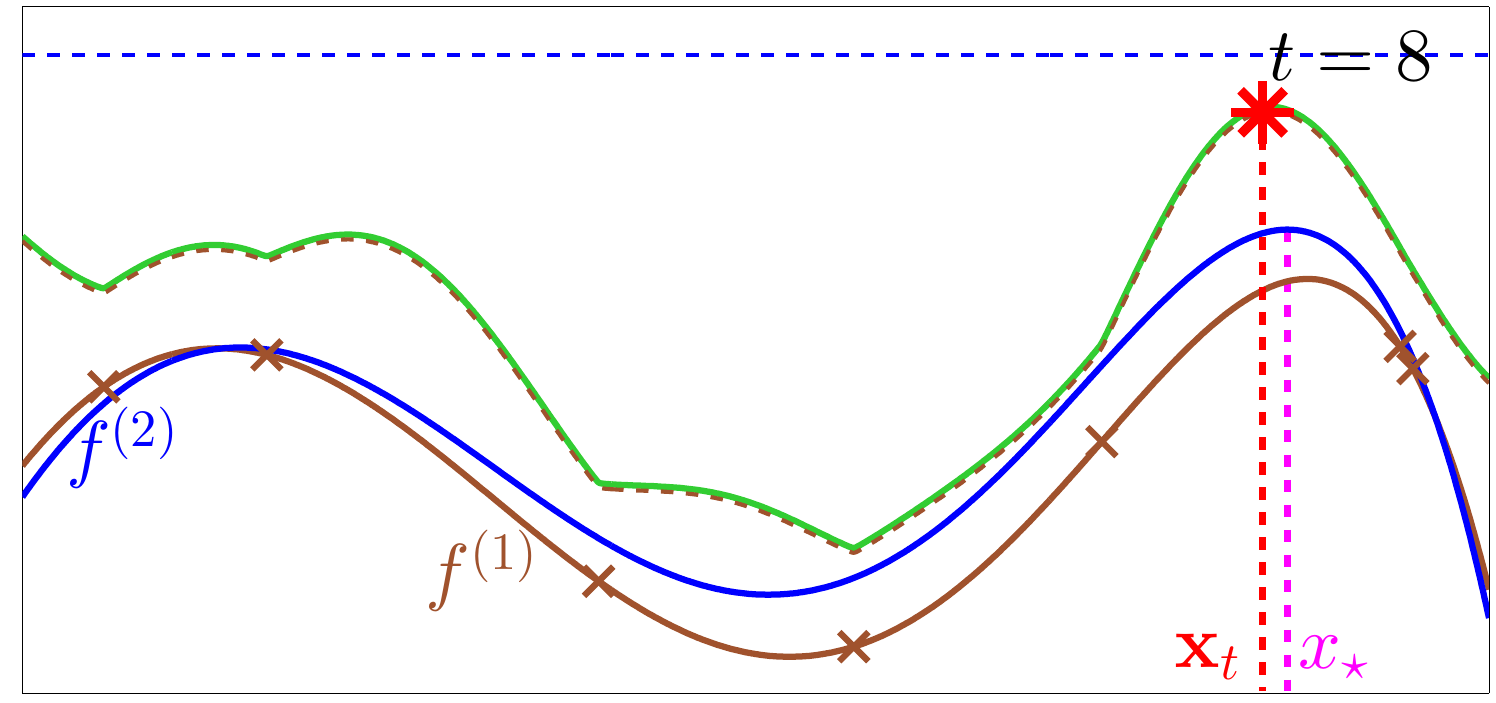} \hspace{\imSimHspace}
} \\
\vspace{\imSimVspace}
\subfigure{
  \includegraphics[width=\imSimWidth]{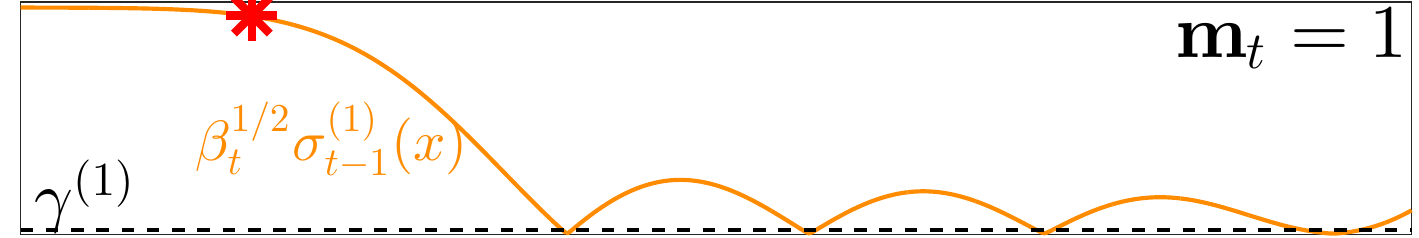} \hspace{\imSimHspace}
}
\subfigure{
  \includegraphics[width=\imSimWidth]{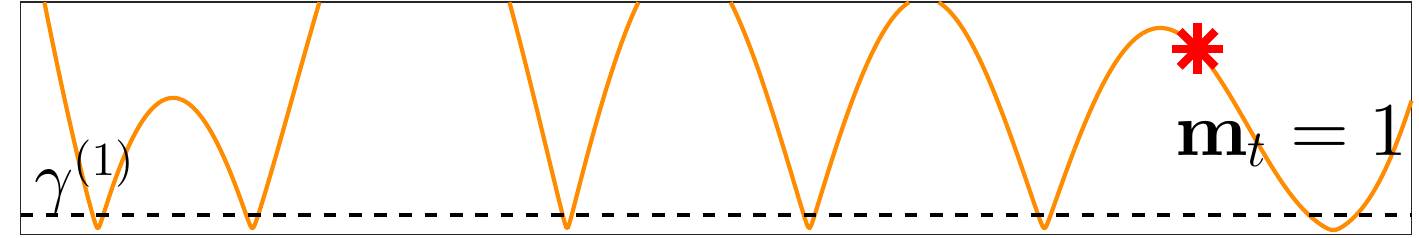} \hspace{\imSimHspace}
} \\

\vspace{\imSimVGap}
\subfigure{
  \includegraphics[width=\imSimWidth]{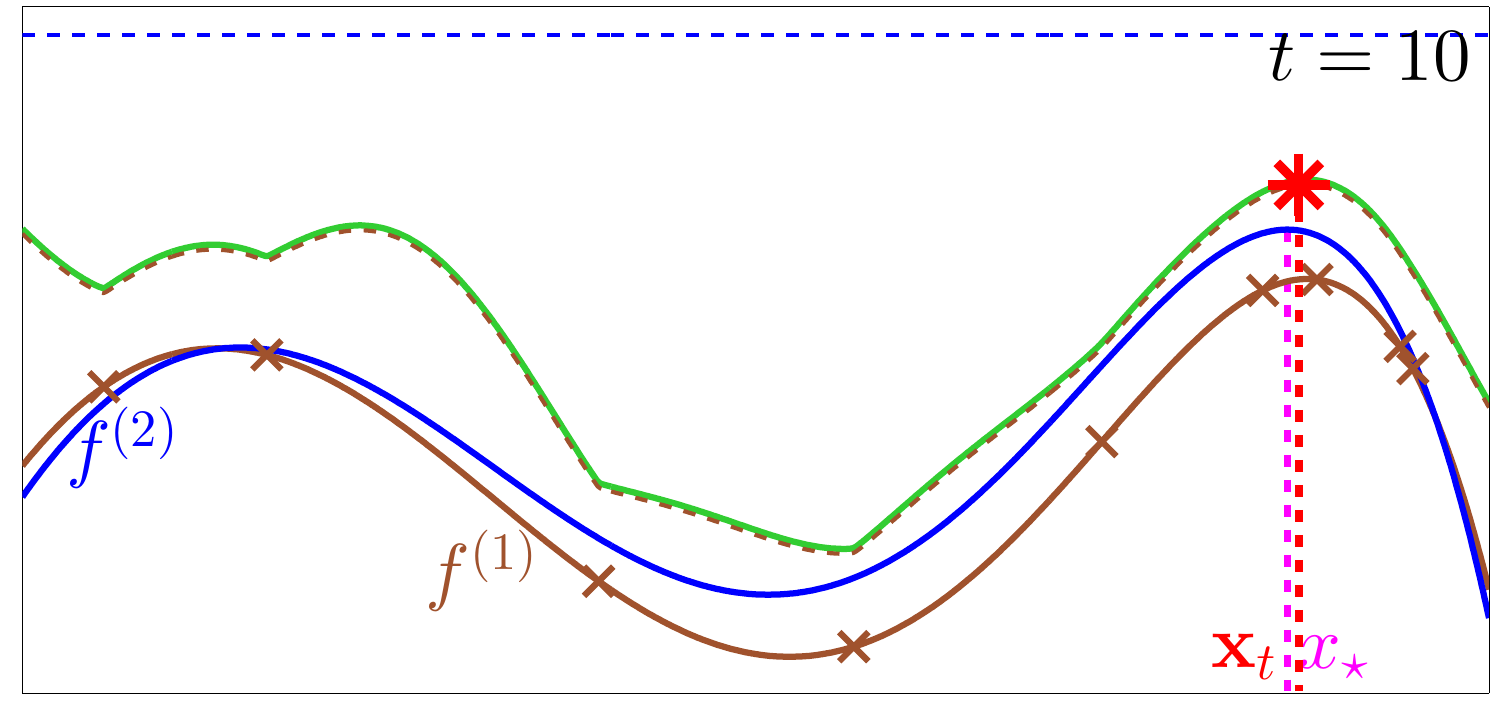} \hspace{\imSimHspace}
}
\subfigure{
  \includegraphics[width=\imSimWidth]{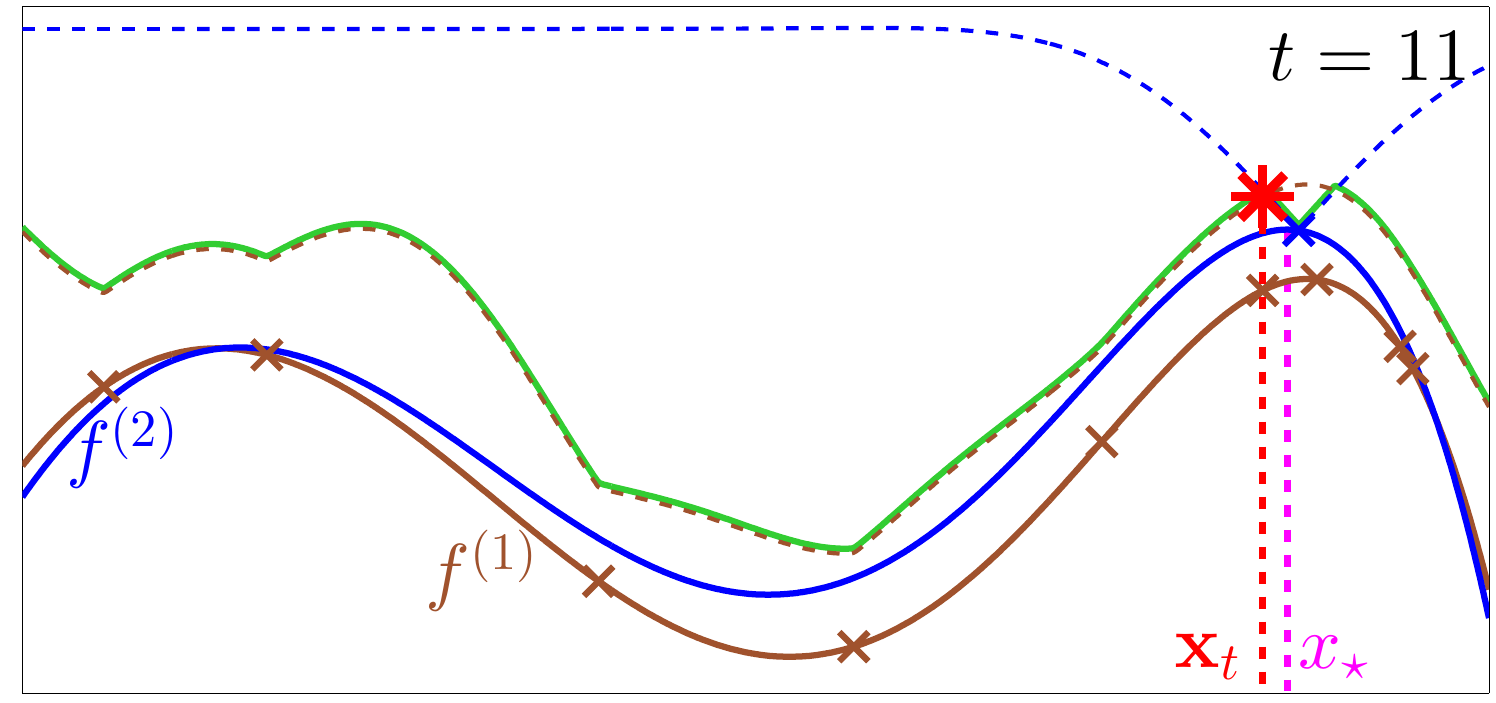} \hspace{\imSimHspace}
} \\
\vspace{\imSimVspace}
\subfigure{
  \includegraphics[width=\imSimWidth]{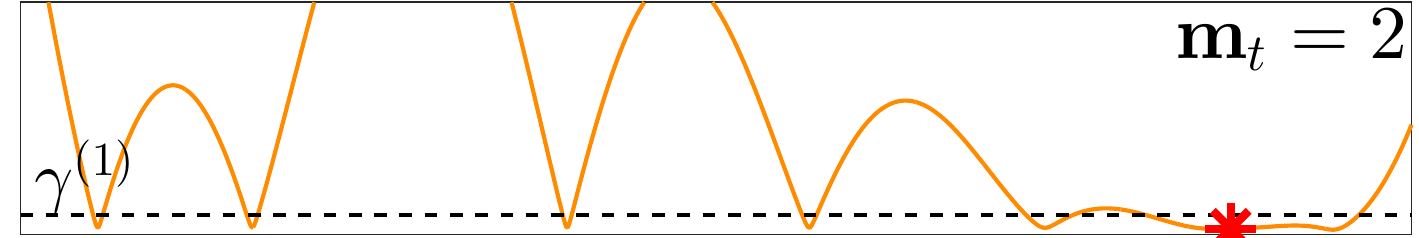} \hspace{\imSimHspace}
}
\subfigure{
  \includegraphics[width=\imSimWidth]{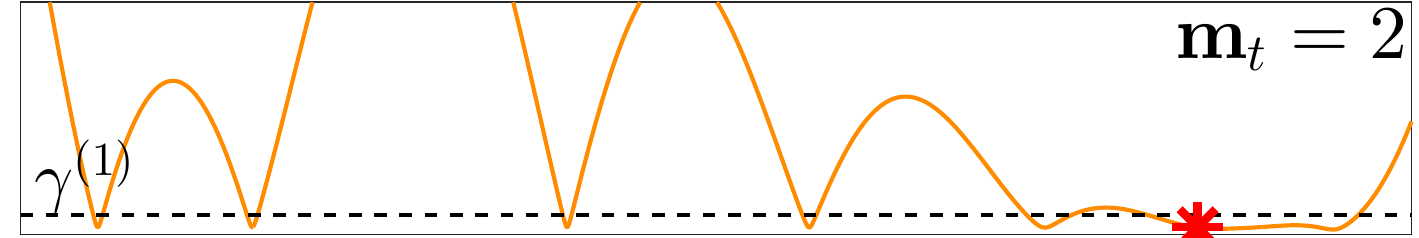} \hspace{\imSimHspace}
} \\

\vspace{\imSimVGap}
\subfigure{
  \includegraphics[width=\imSimWidth]{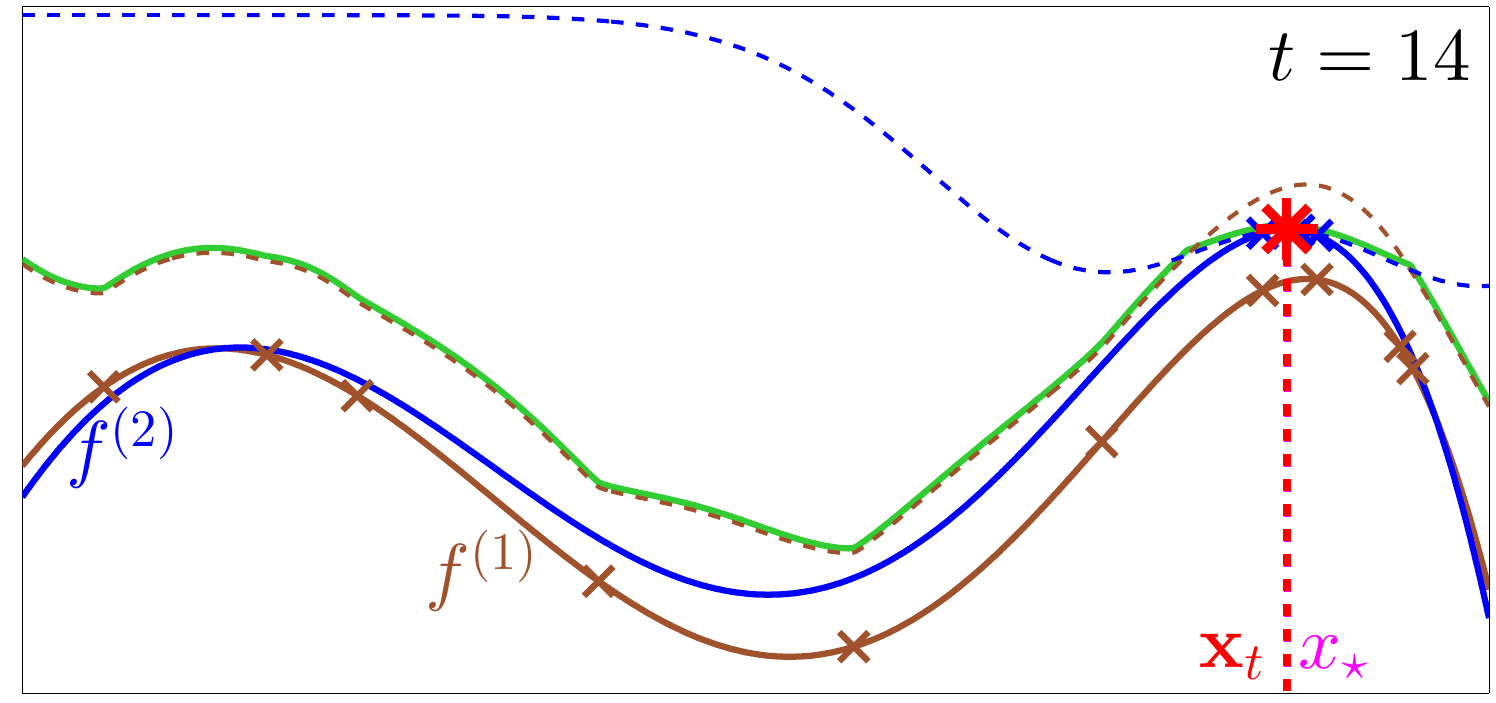} \hspace{\imSimHspace}
}
\subfigure{
  \includegraphics[width=\imSimWidth]{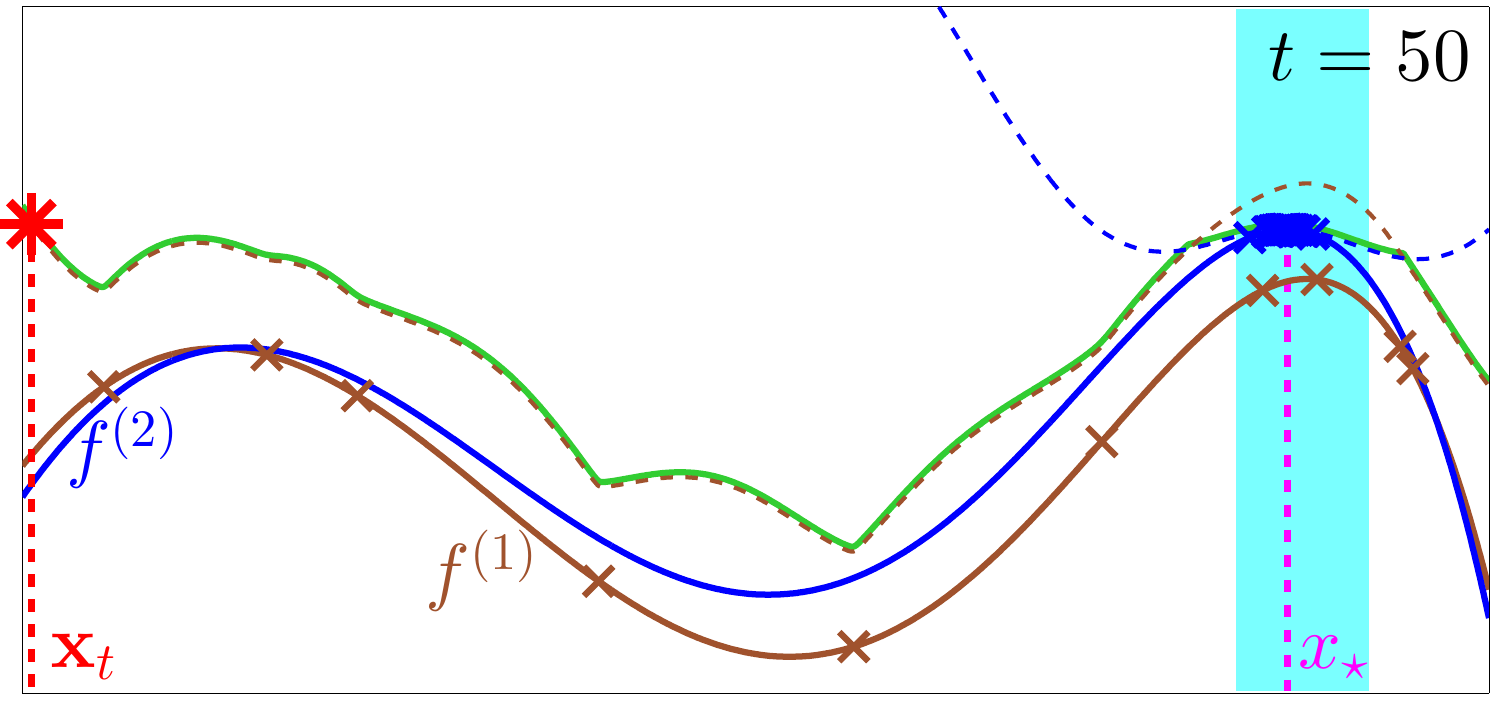} \hspace{\imSimHspace}
} \\
\vspace{\imSimVspace}
\subfigure{
  \includegraphics[width=\imSimWidth]{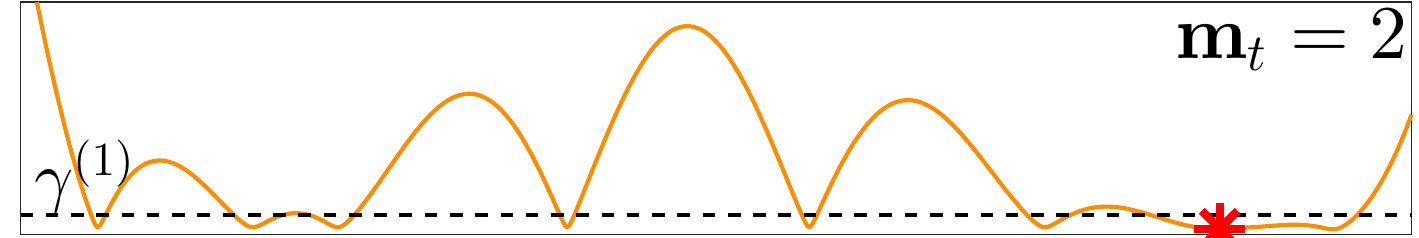} \hspace{\imSimHspace}
}
\subfigure{
  \includegraphics[width=\imSimWidth]{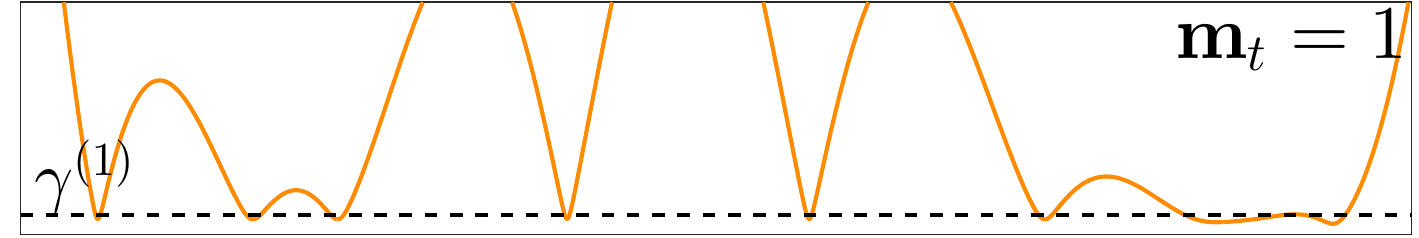} \hspace{\imSimHspace}
} \\
\vspace{\imcaptionspace}
\caption[]{ %\small
\simulationText
\vspace{0.1in}

At the initial stages, \mfgpucbs is mostly exploring $\Xcal$ in the first
fidelity. $\betath\sigmaonetmo$ is large and we are yet to constrain $\funcone$ well
to proceed to $m=2$.
At $t=10$, we have constrainted $\funcone$ sufficiently well at a region around the 
optimum. $\betath\sigmaonetmo(\xt)$ falls below $\gammaone$ and we query at $\mt=2$.
Notice that once we do this (at $t=11$), $\ucbtwot$ dips to change $\ucbt$ in that
region.
At $t=14$, \mfgpucbs has identified the maximum $\xopt$ with just $4$ queries to
$\functwo$.
In the last figure, at $t=50$, the algorithm decides to explore at a point far away
from the optimum. However, this query occurs in the first fidelity since we have not
sufficiently constrained $\funcone(\xt)$ in this region.
The key idea is that it is \emph{not necessary} to query such regions at the second
fidelity as the first fidelity alone is enough to conclude that it is suboptimal.
Herein lies the crux of our method.
% \toworkon{State the last sentence somewhere in the main text.}
The region shaded in cyan in the last figure is the good set
$\Xcalg = \{x; \functwo(\xopt) - \funcone(x) \leq \zetaone\}$ discussed in
Section~\ref{sec:analysis}.
% As we will see, this set is fundamental to the $2$-fidelity problem. 
Our analysis predicts that most second fidelity queries in \mfgpucbs will be
confined to this set with high probability and the simulation corroborates this claim.
In addition, observe that in a large portion of $\Xcal$, $\ucbt$ is given by
$\ucbonet$ except in a small neighborhood around $\xopt$, where it is given
by $\ucbtwot$.

% The cyan region in the last figure is the good set
% $\Xcalg = \{x:\functwo(x)-\funcone(x) \leq \zetaone\}$.
\vspace{-0.4in}
\label{fig:mfSim}
}
\end{figure*}
}

\newcommand{\insertSimulationShortFigure}{
\begin{figure*}
\centering
\subfigure{
  \includegraphics[width=\imSimWidth]{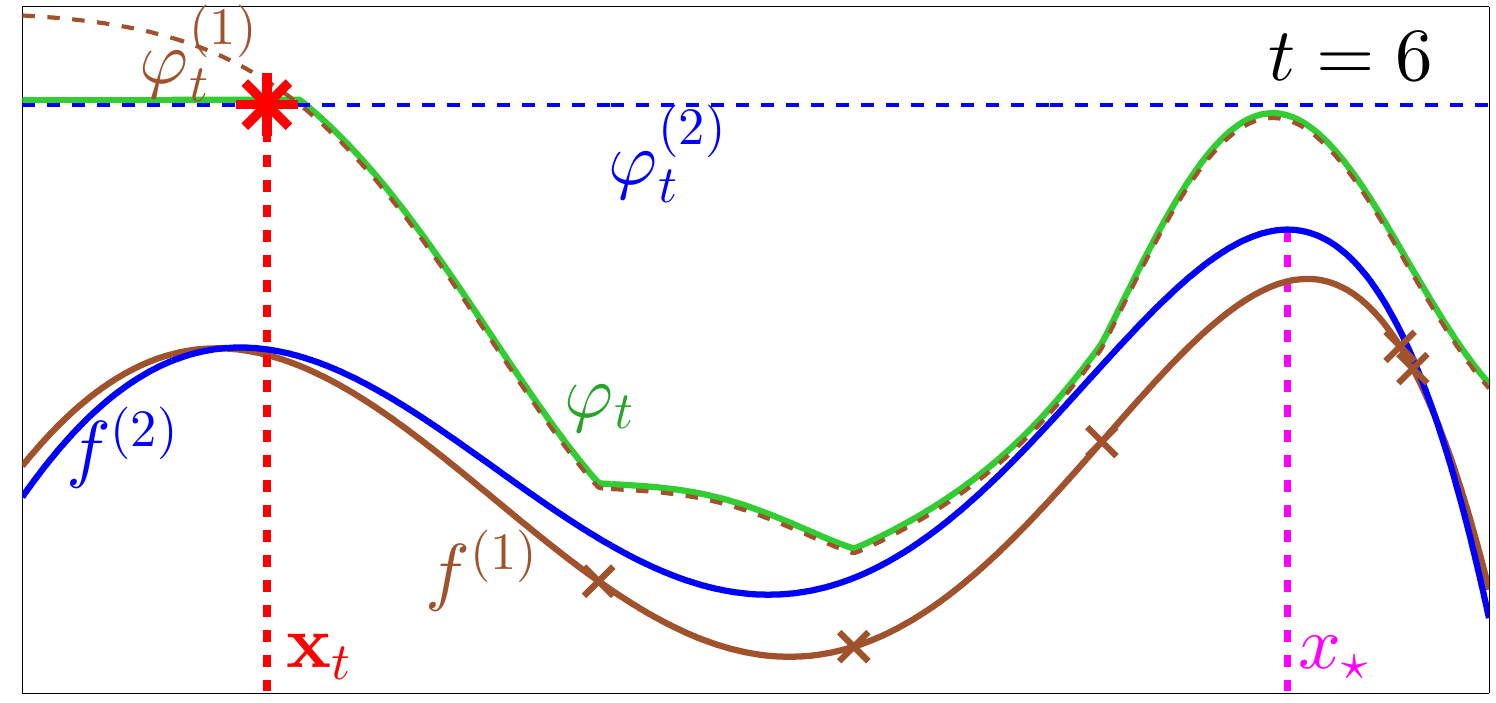} \hspace{\imSimHspace}
}
\subfigure{
  \includegraphics[width=\imSimWidth]{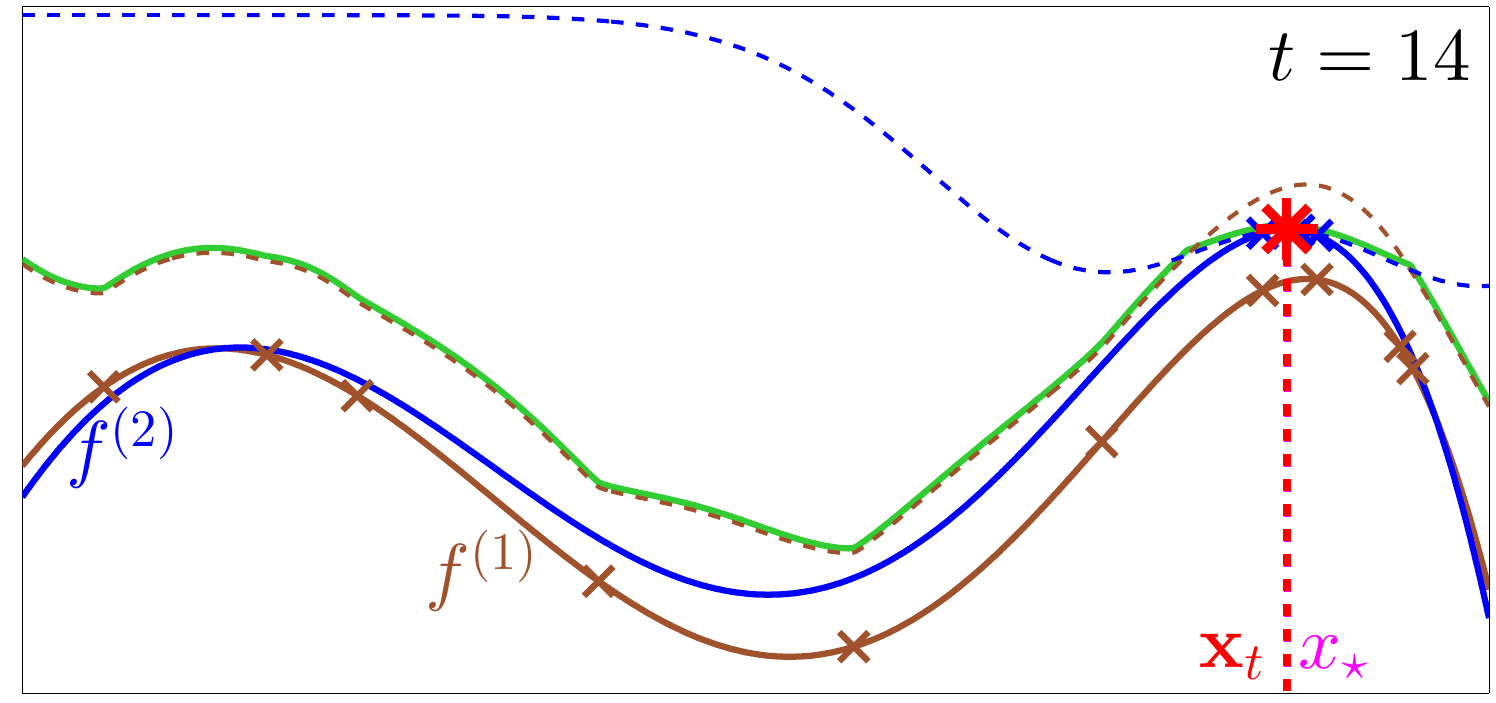} \hspace{\imSimHspace}
} \\
\vspace{\imSimVspace}
\subfigure{
  \includegraphics[width=\imSimWidth]{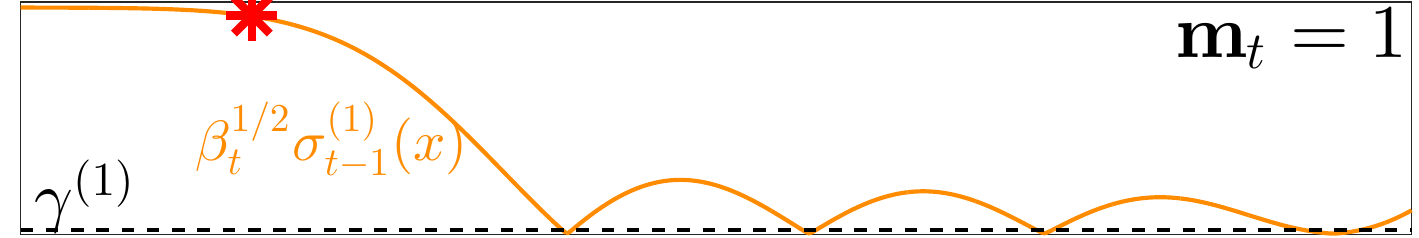} \hspace{\imSimHspace}
}
\subfigure{
  \includegraphics[width=\imSimWidth]{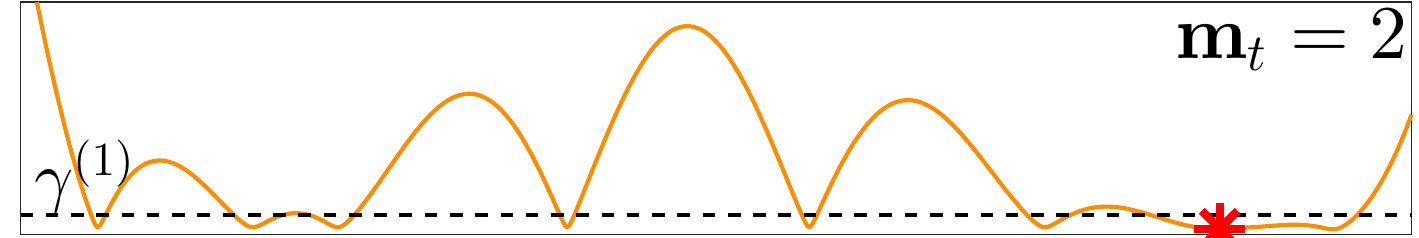} \hspace{\imSimHspace}
}\\[-0.02in]
\vspace{\imcaptionspace}
\caption[]{ \small
\simulationText
See Fig.~\ref{fig:mfSim} in Appendix~\ref{app:intuition} for an extended simulation.
\vspace{\imtextspace}
\label{fig:mfSimShort}
}
\end{figure*}
}

\newcommand{\insertSimulationLong}{
\begin{figure*}
\centering
\subfigure{
  \includegraphics[width=\imSimWidth]{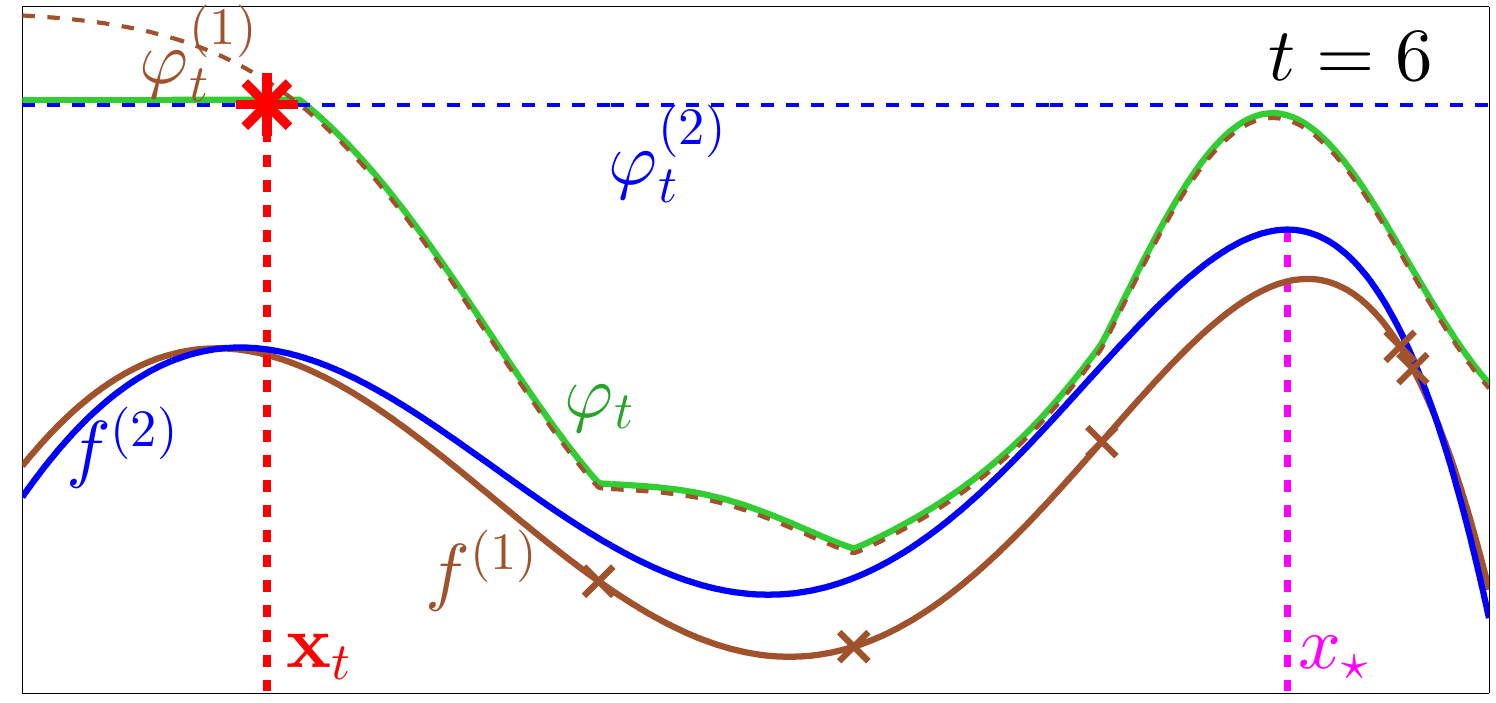} \hspace{\imSimHspace}
}
\subfigure{
  \includegraphics[width=\imSimWidth]{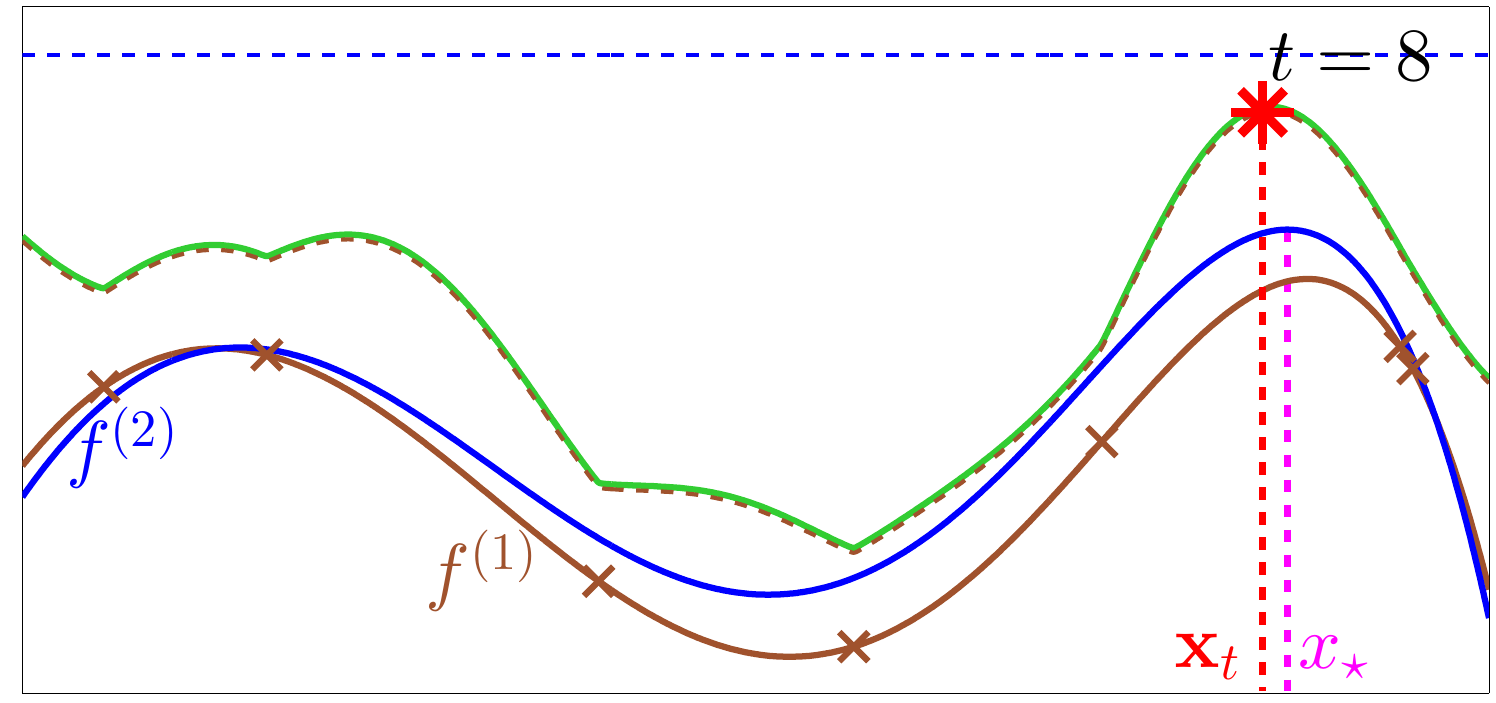} \hspace{\imSimHspace}
} \\
\vspace{\imSimVspace}
\subfigure{
  \includegraphics[width=\imSimWidth]{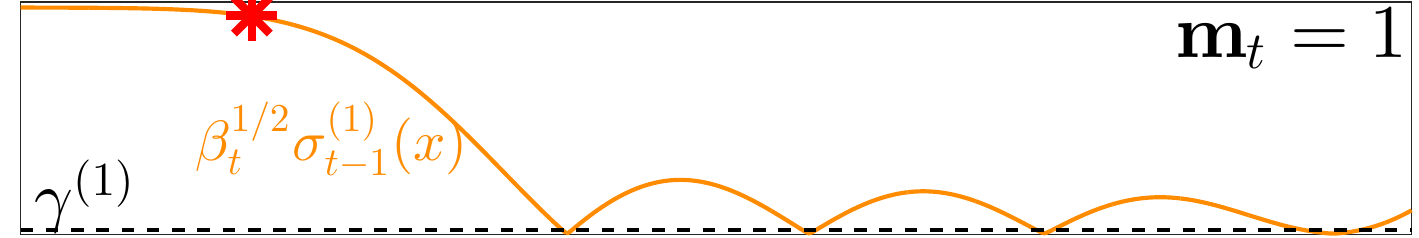} \hspace{\imSimHspace}
}
\subfigure{
  \includegraphics[width=\imSimWidth]{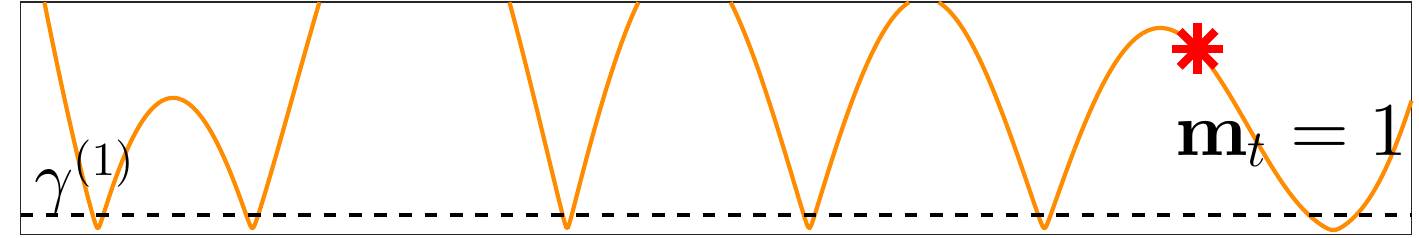} \hspace{\imSimHspace}
} \\

\vspace{\imSimVGap}
\subfigure{
  \includegraphics[width=\imSimWidth]{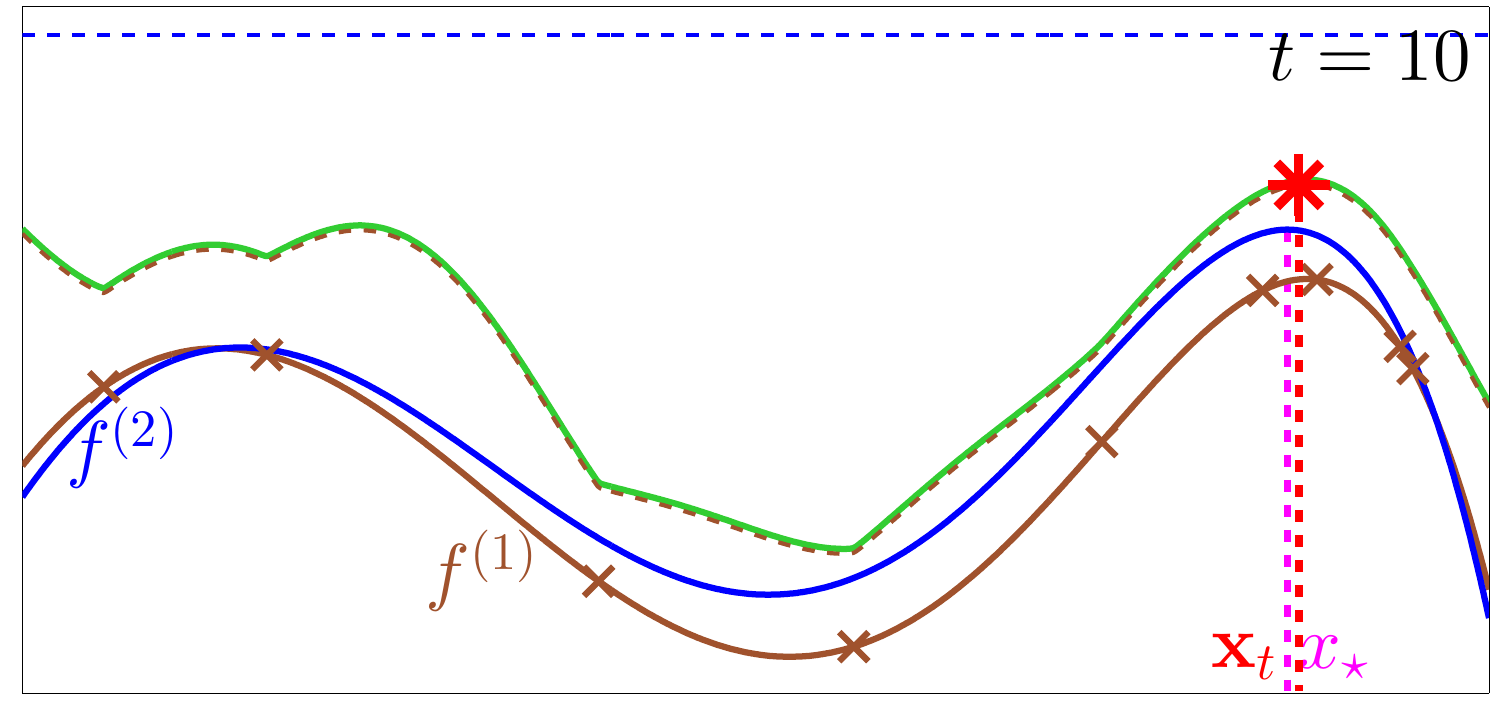} \hspace{\imSimHspace}
}
\subfigure{
  \includegraphics[width=\imSimWidth]{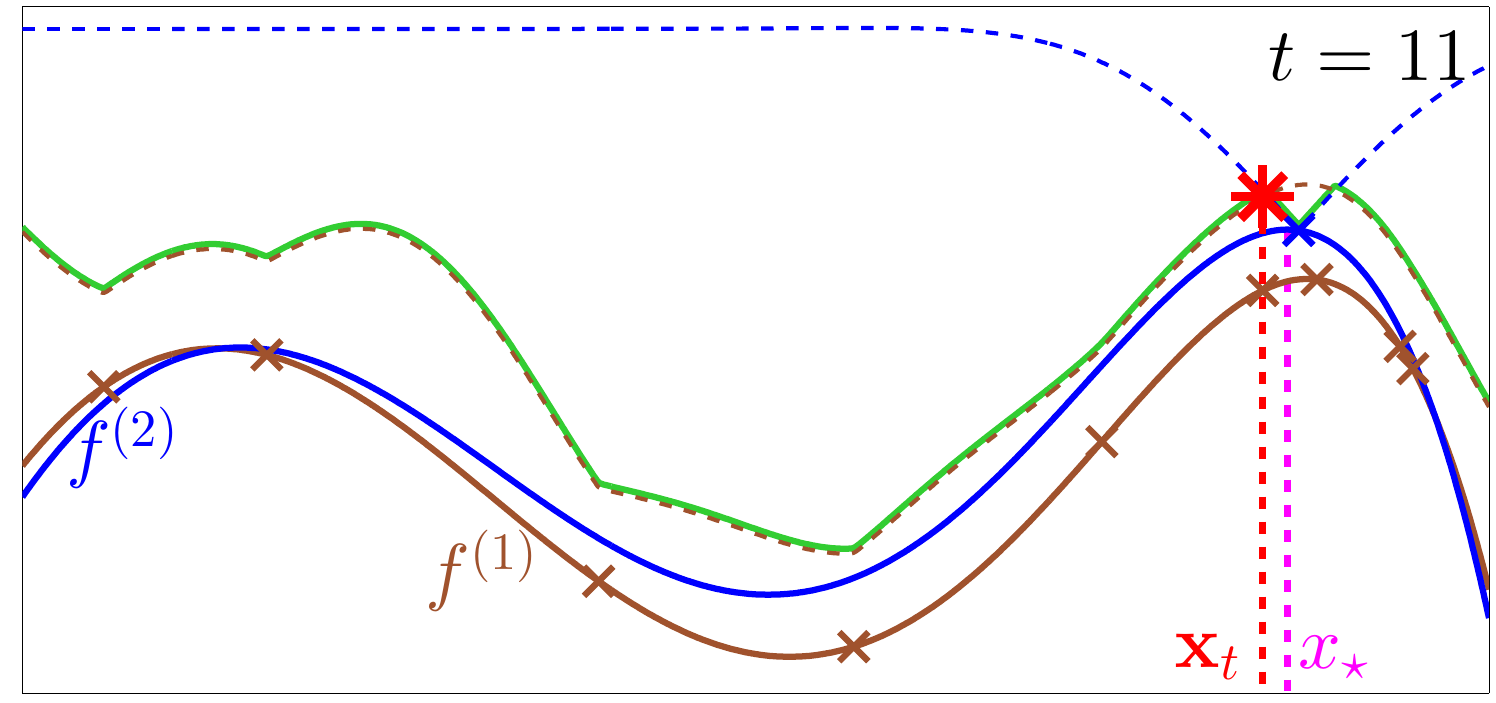} \hspace{\imSimHspace}
} \\
\vspace{\imSimVspace}
\subfigure{
  \includegraphics[width=\imSimWidth]{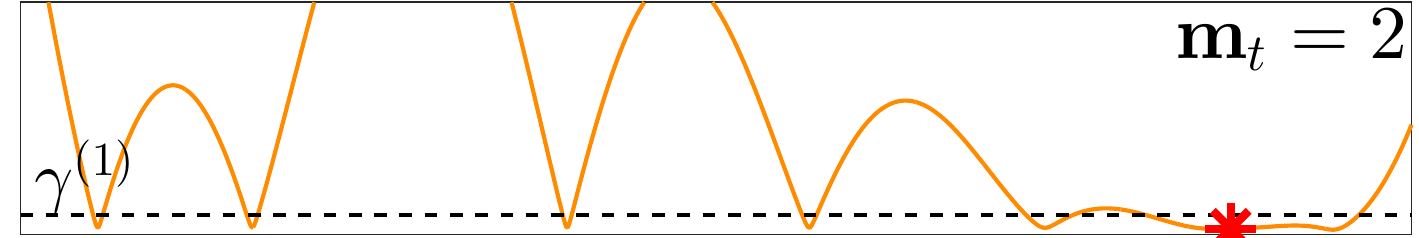} \hspace{\imSimHspace}
}
\subfigure{
  \includegraphics[width=\imSimWidth]{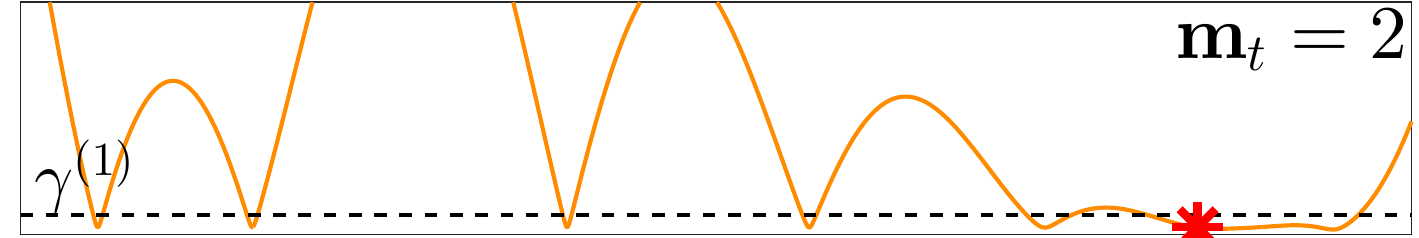} \hspace{\imSimHspace}
} \\

\vspace{\imSimVGap}
\subfigure{
  \includegraphics[width=\imSimWidth]{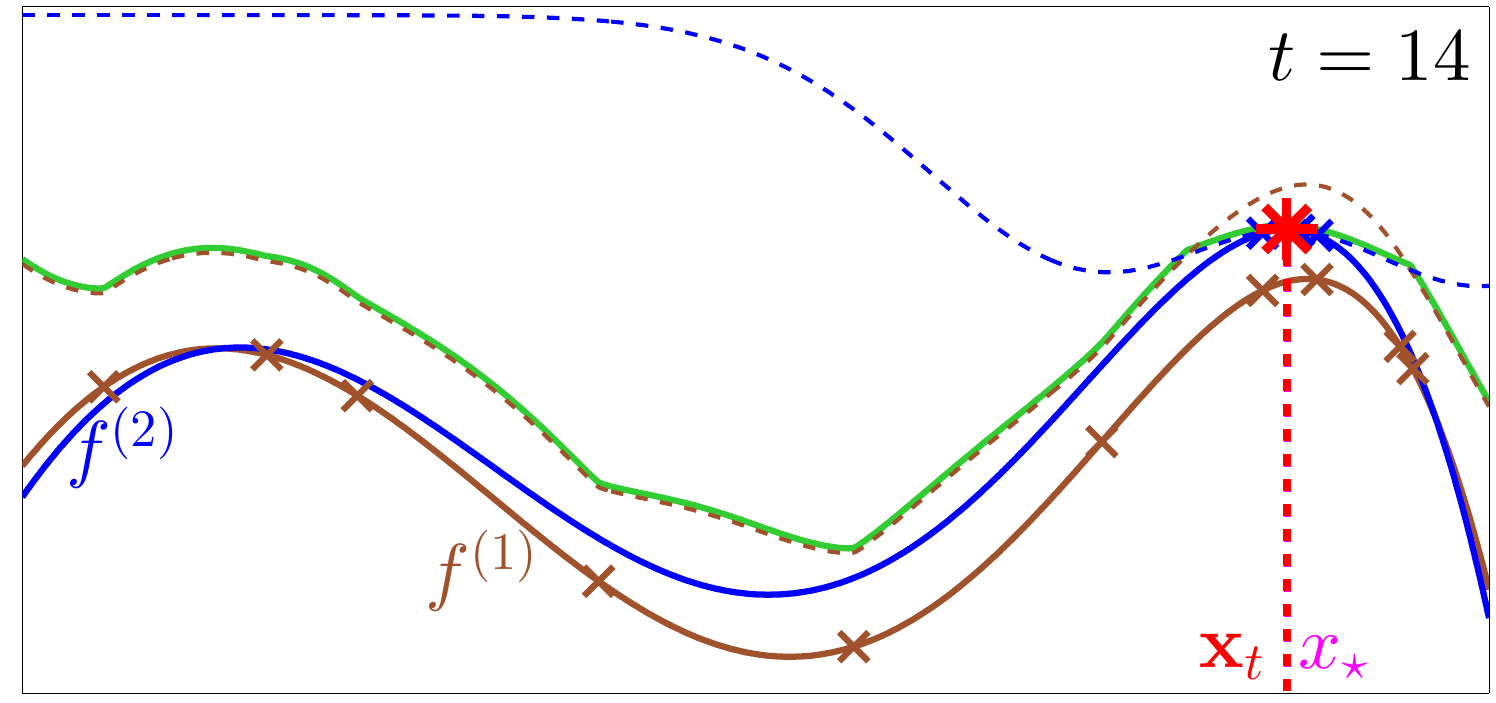} \hspace{\imSimHspace}
}
\subfigure{
  \includegraphics[width=\imSimWidth]{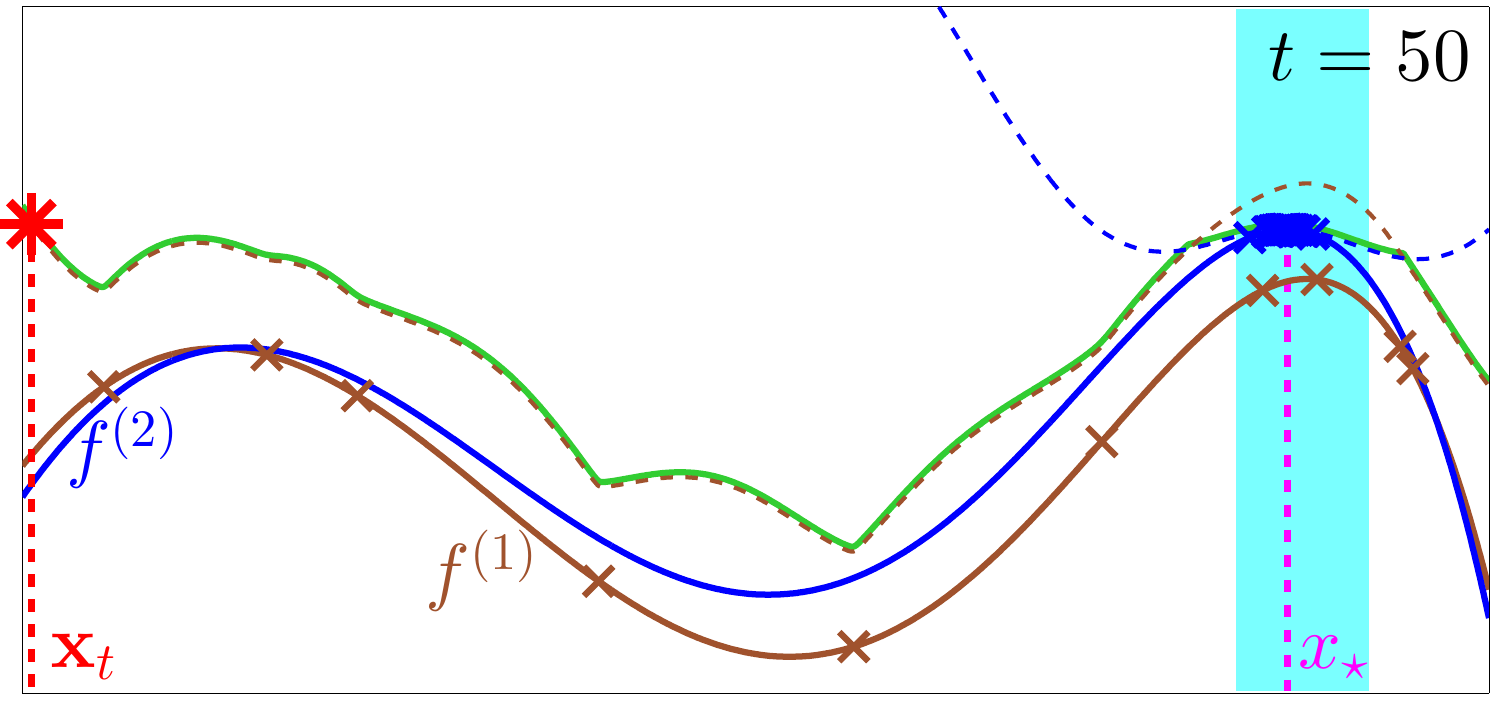} \hspace{\imSimHspace}
} \\
\vspace{\imSimVspace}
\subfigure{
  \includegraphics[width=\imSimWidth]{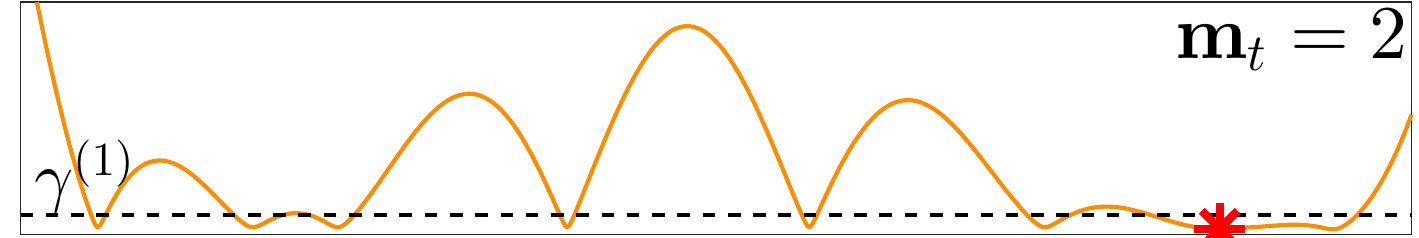} \hspace{\imSimHspace}
}
\subfigure{
  \includegraphics[width=\imSimWidth]{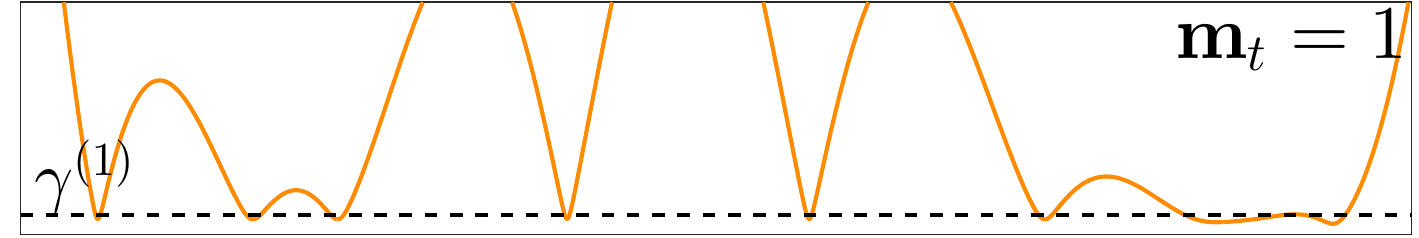} \hspace{\imSimHspace}
} \\
\vspace{0.1in}
\caption[]{ %\small
The $6$ panels illustrate an execution of \mfgpucbs in $2$ fidelities at times
$t=6,8,10,11,14,50$.
In each panel, the top figure illustrates the upper bounds and selection of $\xt$
while the bottom figure illustrates the selection of $\mt$.
% Illustration of \mfgpucbs for a $2$-fidelity 
We have initialised \mfgpucbs with $5$ random points at the first fidelity.
In the top figures,
the solid lines in brown and blue are $\funcone,\functwo$ respectively, and the 
dashed lines are $\ucbonet,\ucbtwot$. 
The solid green line is $\ucb_{t}=\min(\ucbonet,\ucbtwot)$.
The small crosses are queries from $1$ to $t-1$ and the red star is the
maximiser of $\ucbt$, i.e. the next query $\xt$.
$\xopt$, the optimum of $\functwo$ is shown in magenta.
In the bottom figures, the solid orange line is $\betath\sigmaonetmo$ and 
the dashed black line is $\gammaone$. 
When $\betath\sigmaonetmo(\xt) \leq\gammaone$ we play at fidelity $\mt=2$ and otherwise at
$\mt=1$.
The cyan region in the last panel is the good set $\Xcalg$ described in
Section~\ref{sec:analysisPreview}.
\label{fig:mfSimLong}
}
\end{figure*}
}

\newcommand{\insertSimulationFigure}{
\begin{figure*}
\centering
\subfigure{
  \includegraphics[width=\imSimWidth]{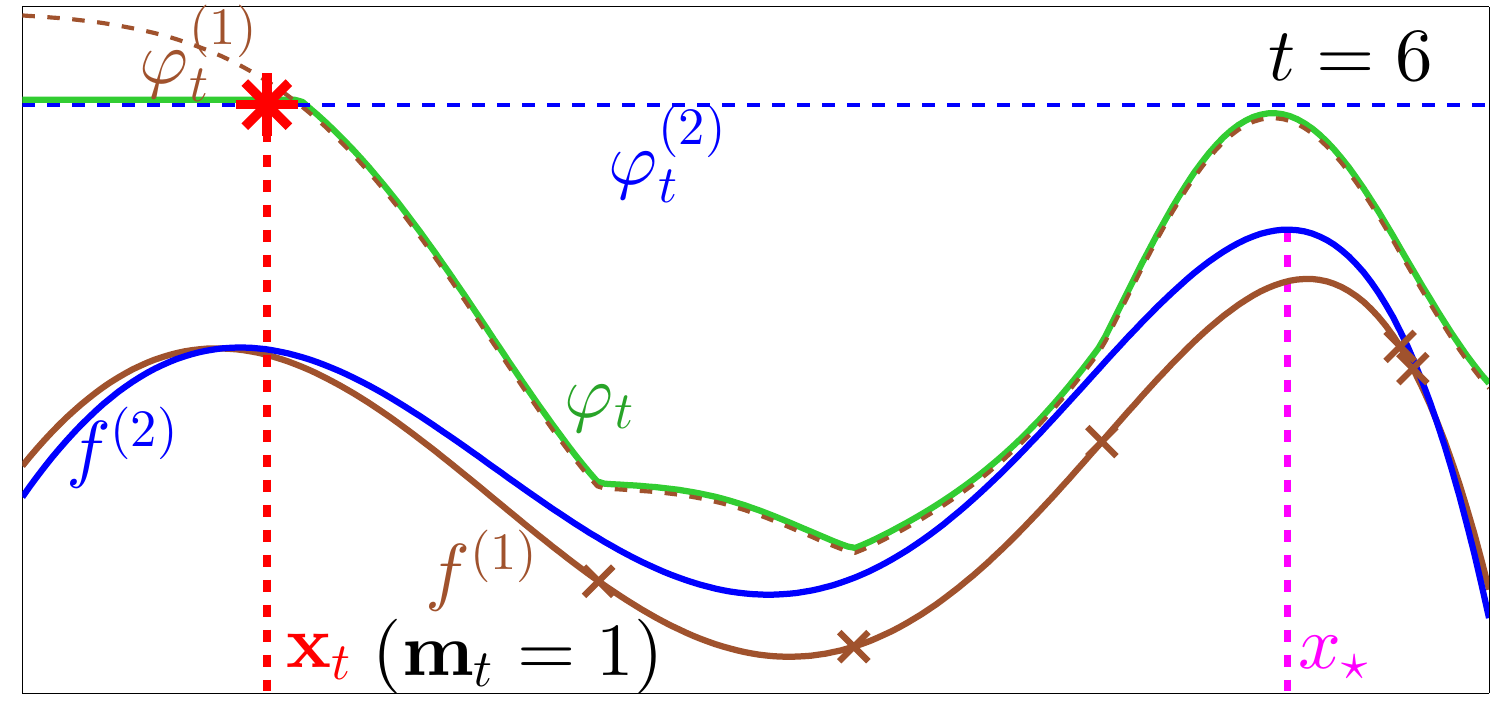} \hspace{\imSimHspace}
}
\subfigure{
  \includegraphics[width=\imSimWidth]{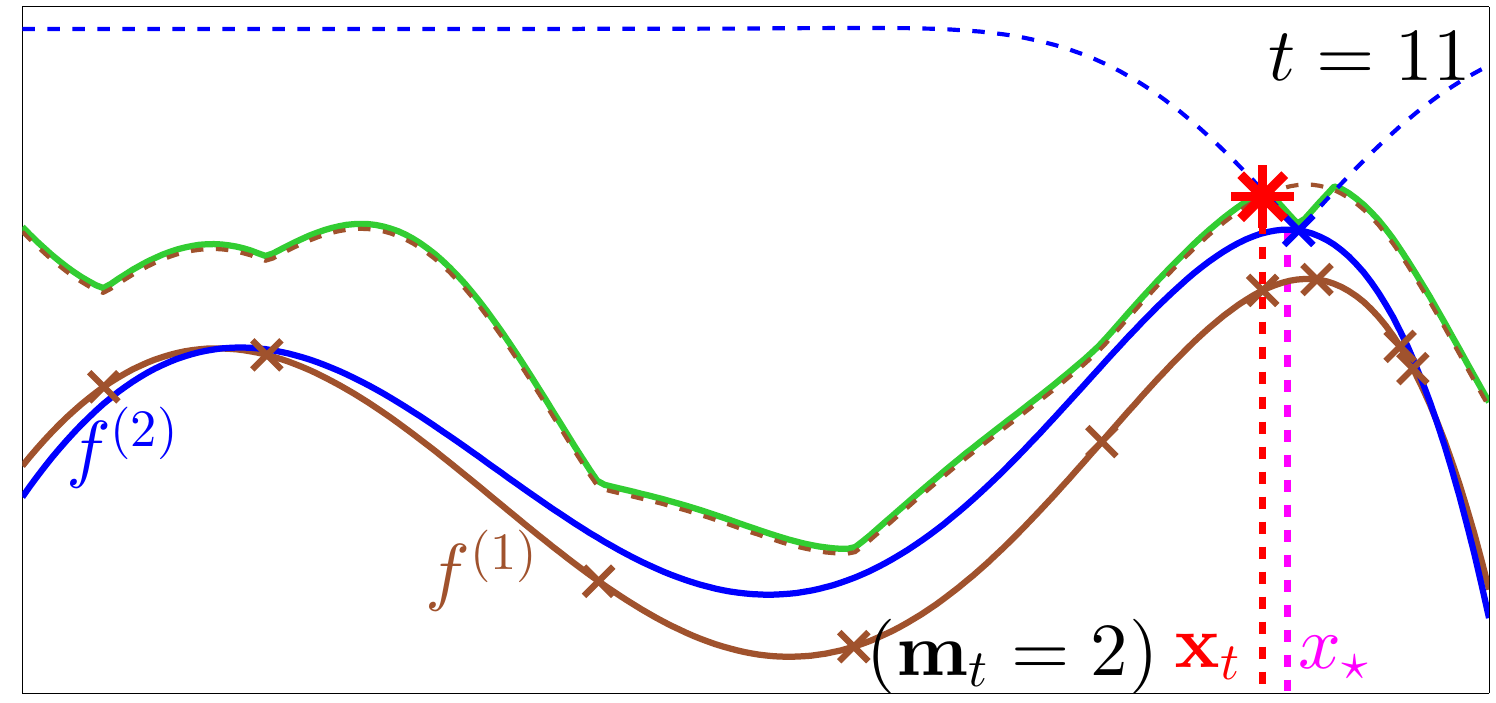} \hspace{\imSimHspace}
} \\
\vspace{\imSimVspace}
\subfigure{
  \includegraphics[width=\imSimWidth]{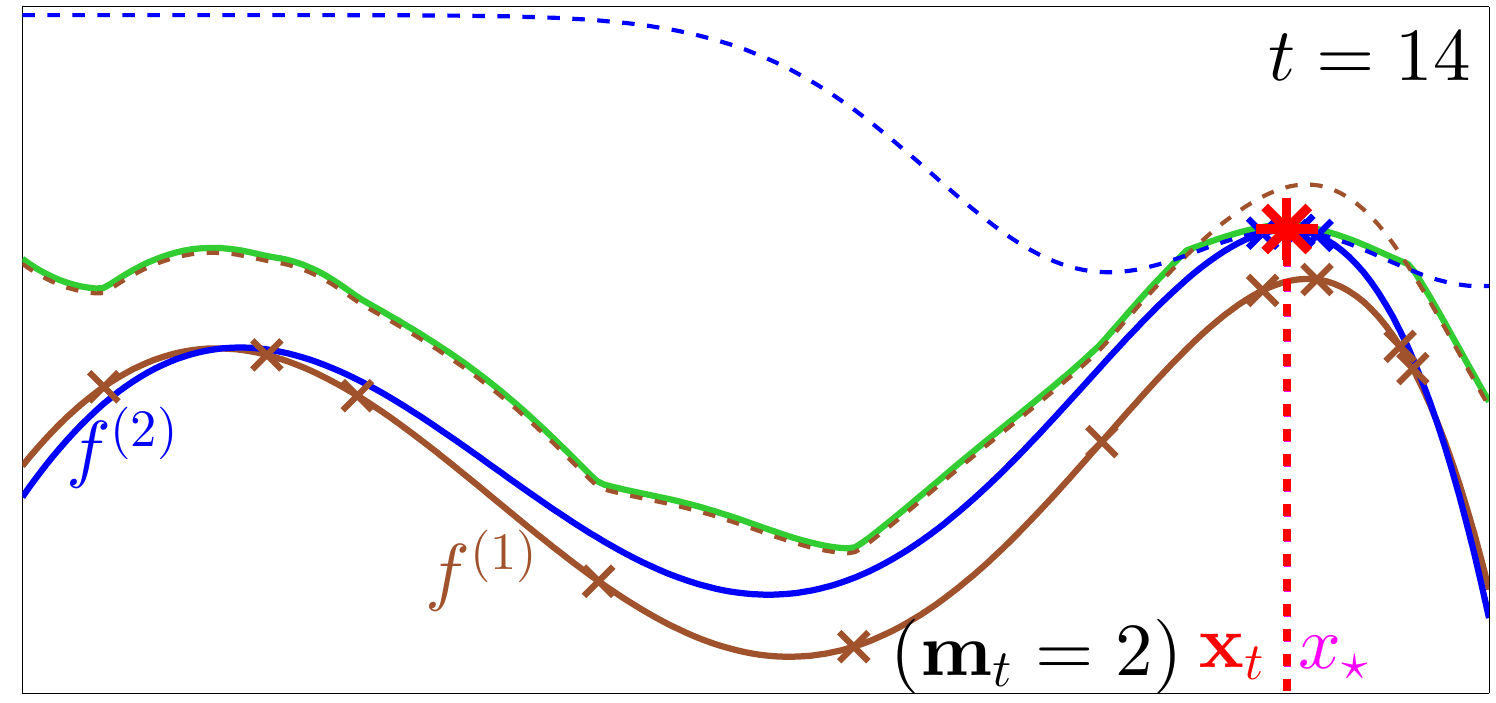} \hspace{\imSimHspace}
}
\subfigure{
  \includegraphics[width=\imSimWidth]{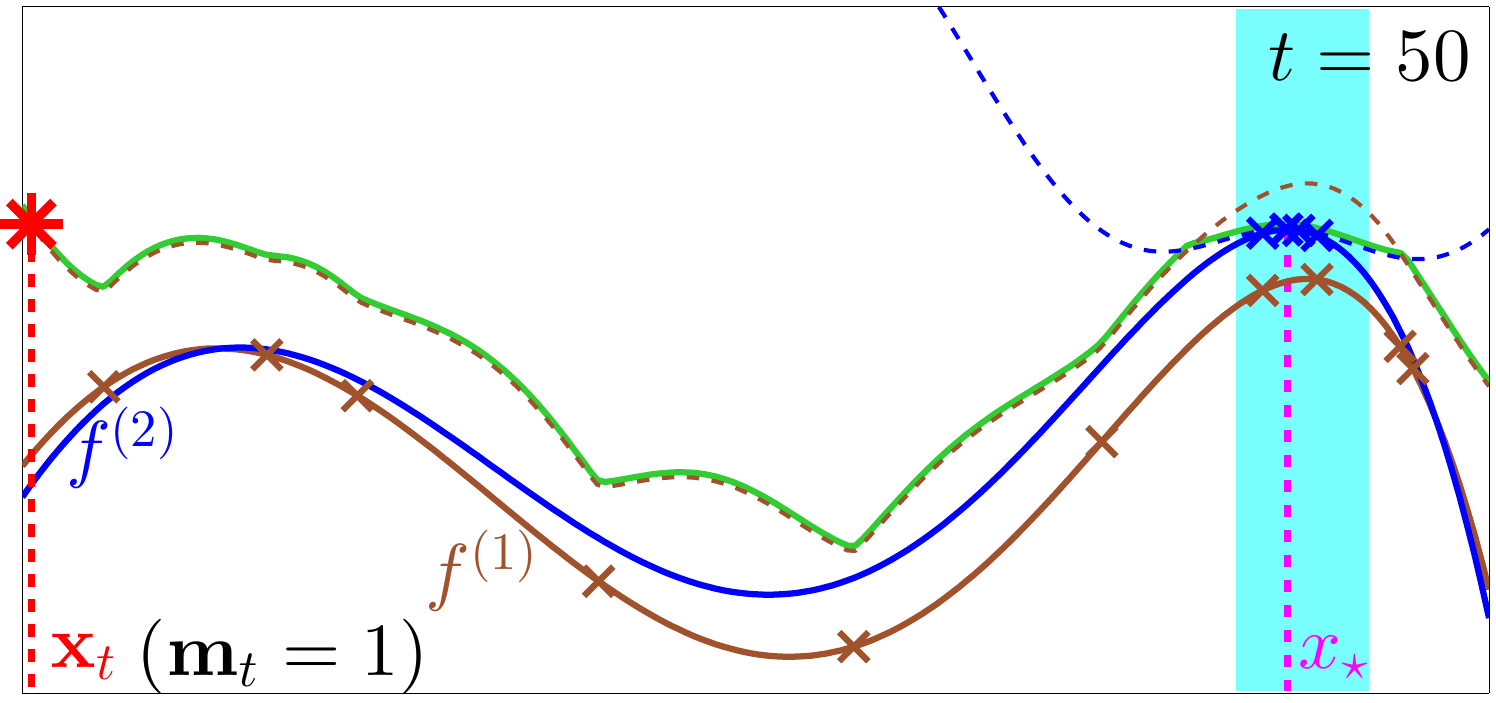} \hspace{\imSimHspace}
} \\
\vspace{\imcaptionspace}
\caption[]{ \small
\simulationText
The cyan region in the last figure is the good set
$\Xcalg = \{x:\functwo(x)-\funcone(x) \leq \zetaone\}$.
\vspace{-0.4in}
\label{fig:mfSim}
}
\end{figure*}
}

\newcommand{\syntheticResultsCommon}{
The title states the function, its dimensionality, the number of
fidelities and the costs we used for each fidelity in the experiment.
All curves barring \directs (which is a deterministic), 
were produced by averaging over 20 experiments.
The error bars indicate one standard error.
}

\newcommand{\insertSRToyOne}{
\begin{figure*}
\centering
\hspace{\imhspthree}
\subfigure{
  \includegraphics[width=\imarrwthree]{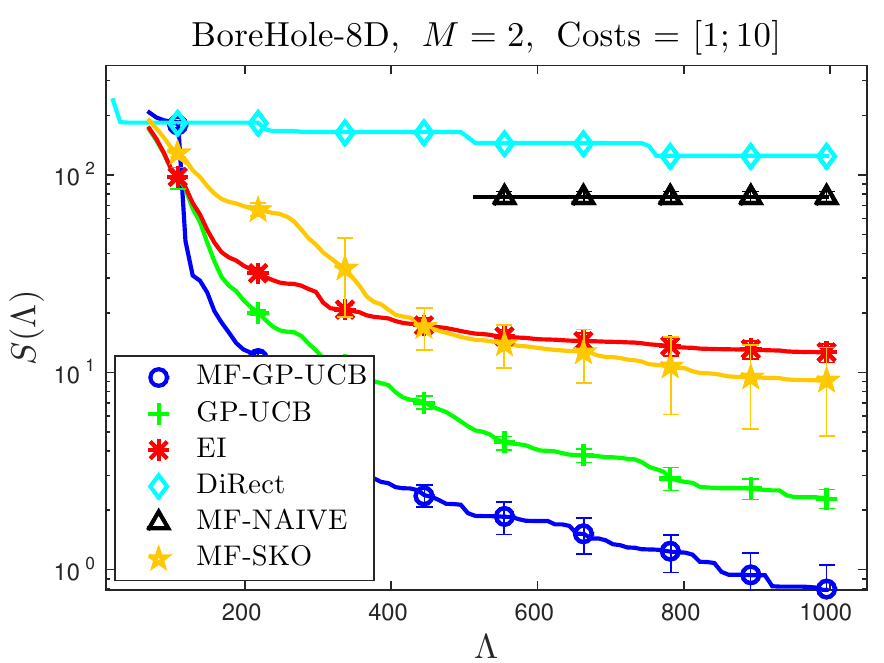} \hspace{\imhspthree}
} 
\subfigure{
  \includegraphics[width=\imarrwthree]{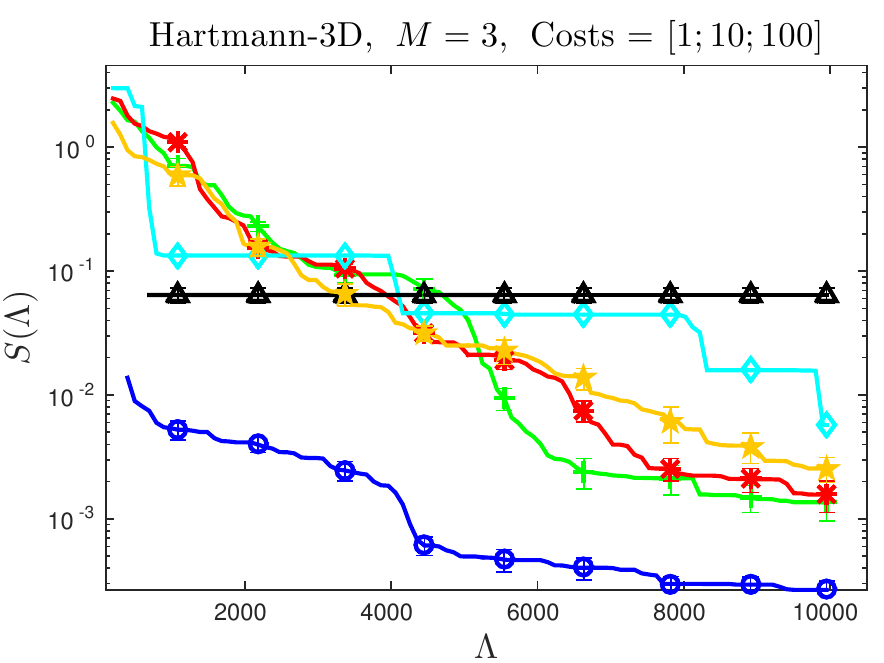} \hspace{\imhspthree}
}
\subfigure{
  \includegraphics[width=\imarrwthree]{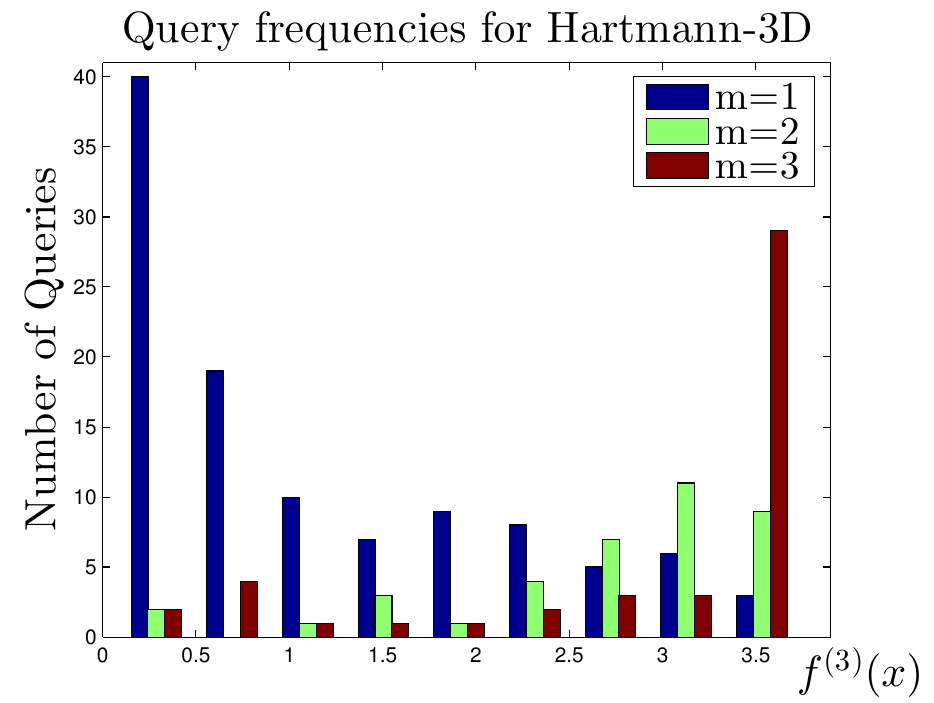} \hspace{\imhspthree}
} \\[-0.2in]
\caption[]{\small
The simple regret $S(\COST)$ against the spent capital $\COST$ on synthetic
functions.
\syntheticResultsCommon
See Figures~\ref{fig:toySRThree},~\ref{fig:badCurrin}~\ref{fig:toyCR} in 
Appendix~\ref{app:experiments} 
for more synthetic results.
The last panel shows the number of queries at different function values at each
fidelity for the Hartmann-3D example. \hspace{-0.1in}
\vspace{\imtextspace}
\label{fig:toySROne}
}
\end{figure*}
}

\newcommand{\insertBadCurrin}{
\begin{figure*}
\centering
\subfigure[]{
  \includegraphics[width=\imarrwtwo]{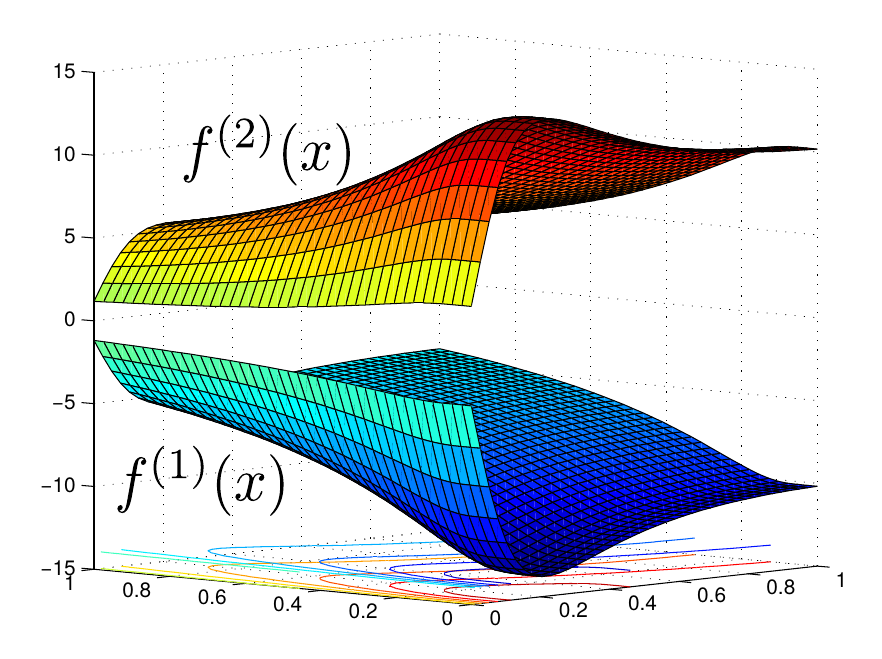} \hspace{-0.0in}
  \label{fig:badCurrinFunctions}
}
\subfigure[]{
  \includegraphics[width=\imarrwtwo]{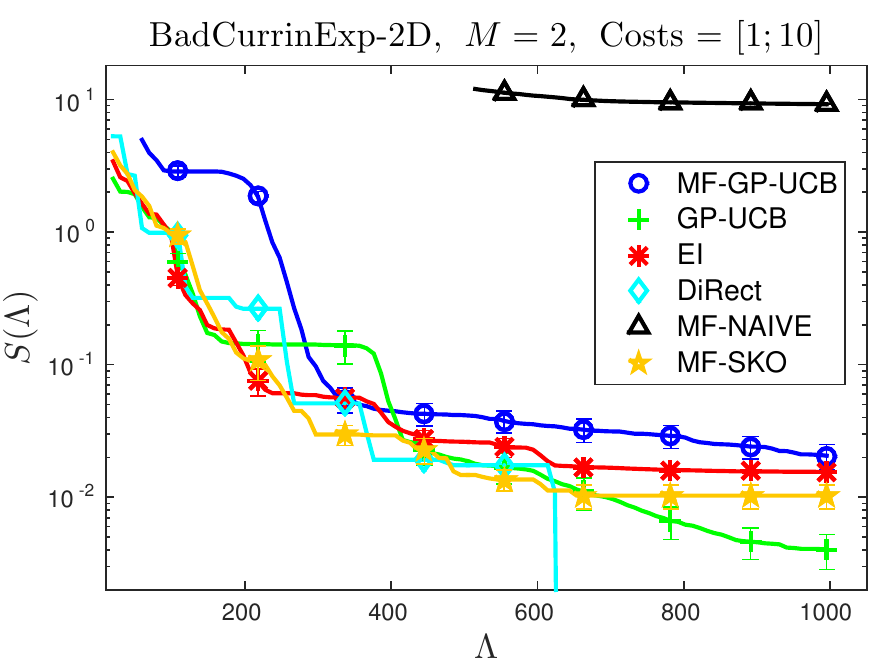}
  \label{fig:badCurrin}
}\\[-0.1in] 
\caption[]{%\small
\subref{fig:badCurrinFunctions}: the functions used in the Bad Currin Exponential
experiment where $\funcone = -\functwo$.
\subref{fig:badCurrin}: the simple regret for this experiment.
See caption under Figure~\ref{fig:toySRAll} for more details.
% \vspace{\imtextspace}
\label{fig:badCurrin}
}
\end{figure*}
}

\newcommand{\insertSRToyAll}{
\begin{figure*}
\centering
% \hspace{\imhsptwo}
\subfigure{
  \includegraphics[width=\imarrwtwo]{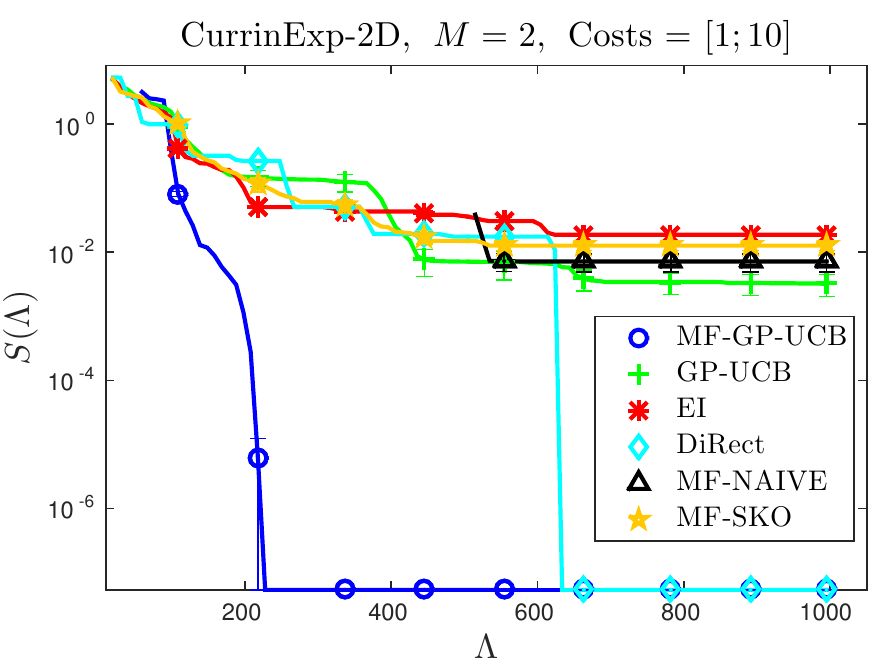} \hspace{\imhsptwo}
}
\subfigure{
  \includegraphics[width=\imarrwtwo]{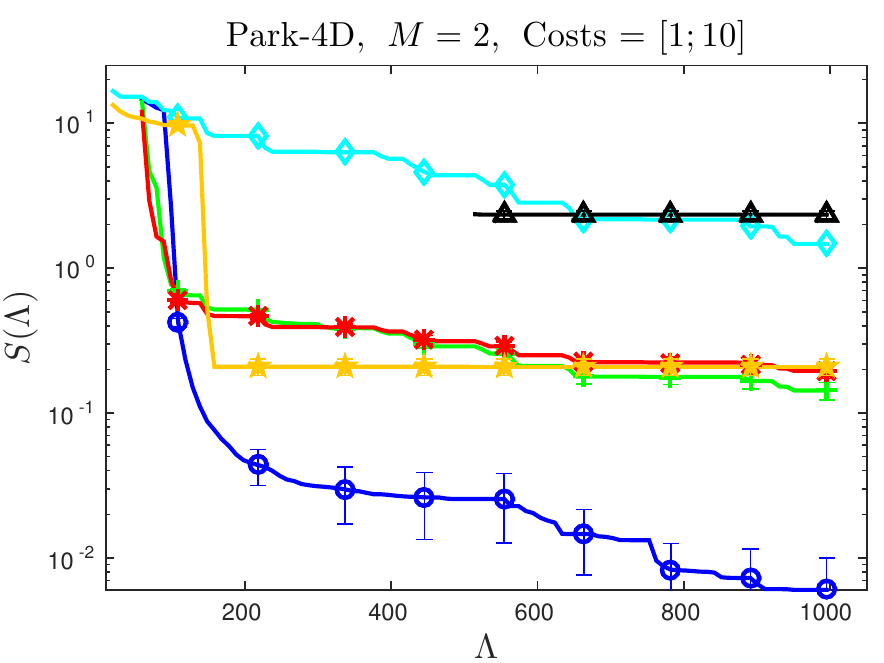} %\hspace{\imhsptwo}
} \\
\subfigure{
  \includegraphics[width=\imarrwtwo]{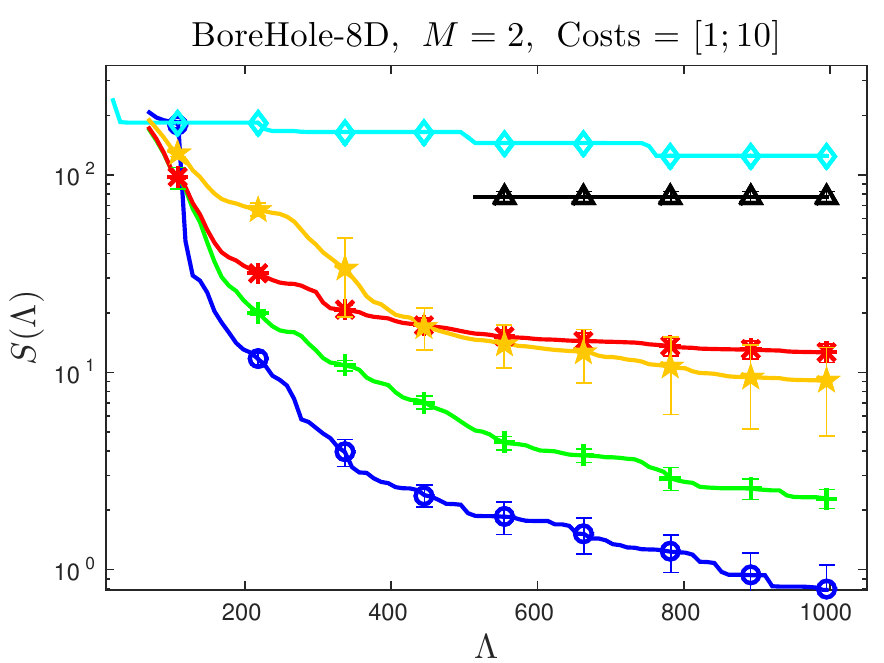} \hspace{\imhsptwo}
} 
\subfigure{
  \includegraphics[width=\imarrwtwo]{figs/Hartmann3-SR} %\hspace{\imhsptwo}
} \\
\subfigure{
  \includegraphics[width=\imarrwtwo]{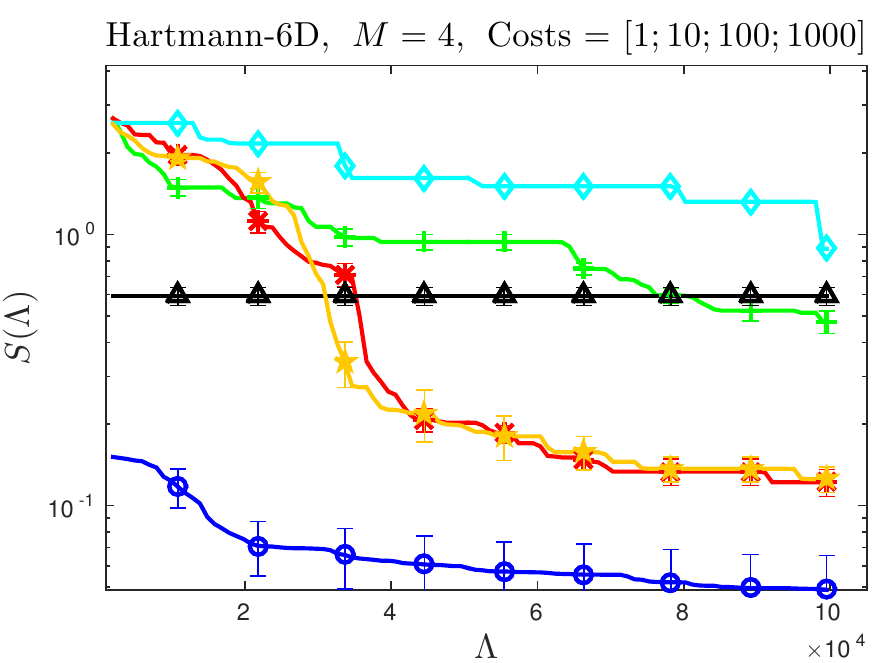} \hspace{\imhsptwo}
}
\subfigure{
  \includegraphics[width=\imarrwtwo]{figs/fvalHist} %\hspace{\imhsptwo}
} 
\vspace{\imcaptionspace}
\caption[]{\small
The simple regret $S(\COST)$~\eqref{eqn:defnMFSR} against the spent capital $\COST$ on
the synthetic functions. 
% \syntheticResultsCommon
The title states the function, its dimensionality, the number of
fidelities and the costs we used for each fidelity in the experiment;
for example, in the fourth panel, we used $M=3$ fidelities, with costs
$\costone=1, \costtwo=10, \costmm{3}=100$ on the 3 dimensional Hartmann function.
All curves barring \directs (which is a deterministic), 
were produced by averaging over 20 experiments.
The error bars indicate one standard error.
All figures follow the legend in the first figure for the Currin exponential function.
The last panel shows the number of queries at different function values at each
fidelity for the Hartmann-3D example. \hspace{-0.1in}
% \vspace{\imtextspace}
\label{fig:toySRAll}
}
\end{figure*}
}

\newcommand{\insertSRToyThree}{
\begin{figure*}
\centering
\hspace{\imhspthree}
\subfigure{
  \includegraphics[width=\imarrwthree]{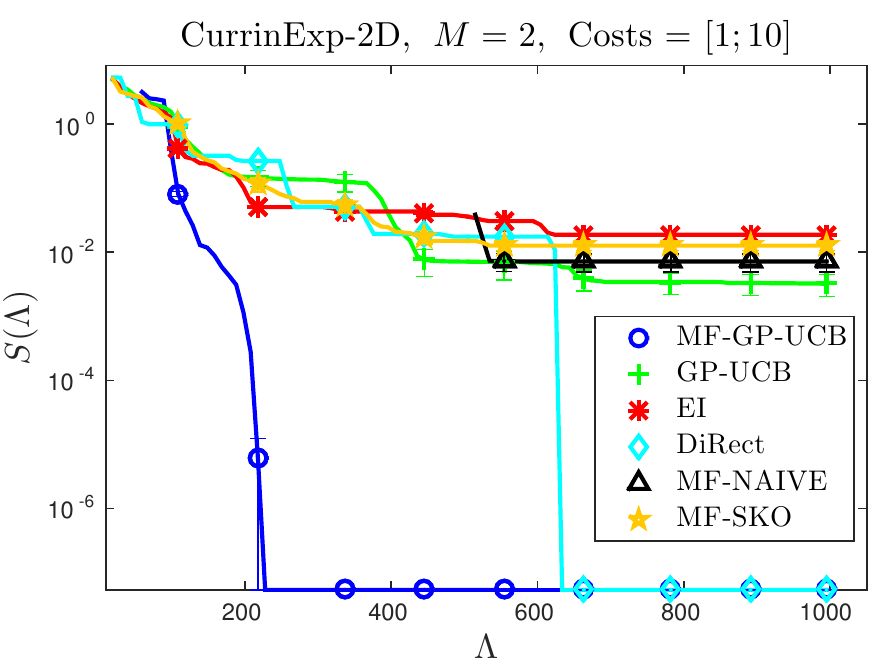} \hspace{\imhspthree}
  \label{fig:currinExp}
}
\subfigure{
  \includegraphics[width=\imarrwthree]{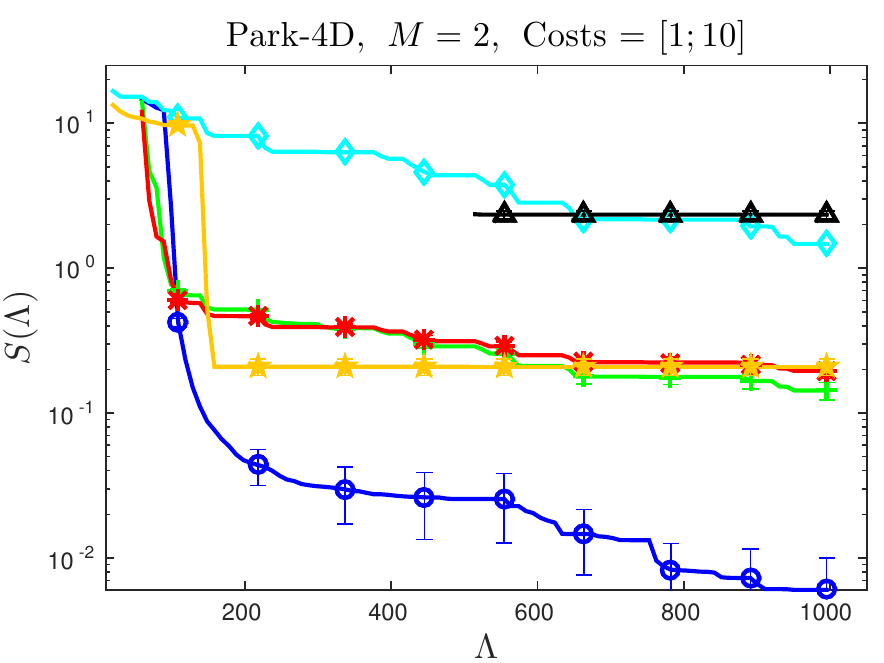} \hspace{\imhspthree}
  \label{fig:park}
}
\subfigure{
  \includegraphics[width=\imarrwthree]{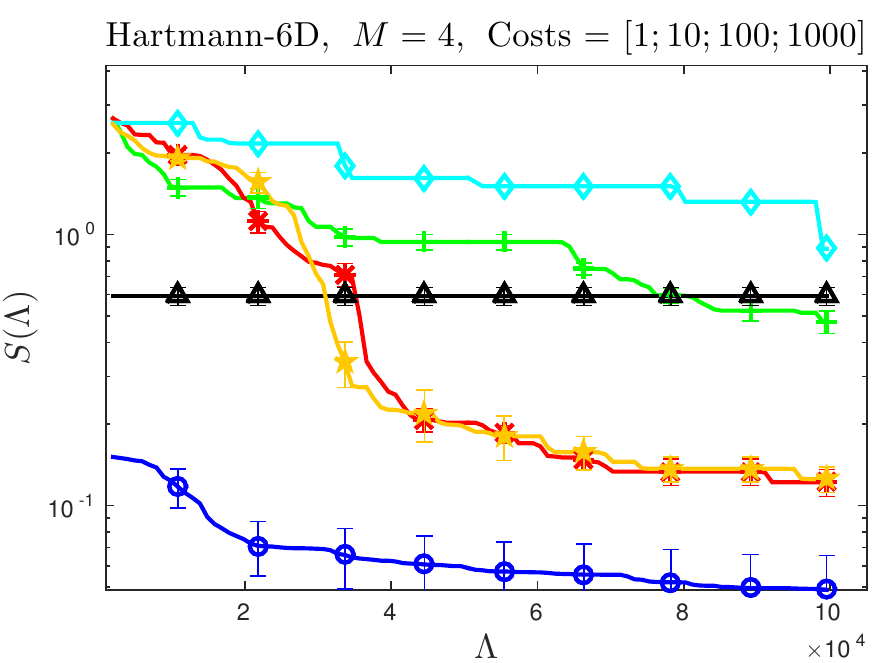} \hspace{\imhspthree}
  \label{fig:hartmann}
}\\[-0.1in] 
\caption[]{\small
The simple regret $S(\COST)$ against the spent capital $\COST$ on
the synthetic functions. 
\syntheticResultsCommon
\label{fig:toySRThree}
}
\end{figure*}
}

\newcommand{\insertToyResultsSRFigure}{
\begin{figure*}
\centering
\hspace{\imhspthree}
\subfigure{
  \includegraphics[width=\imarrwthree]{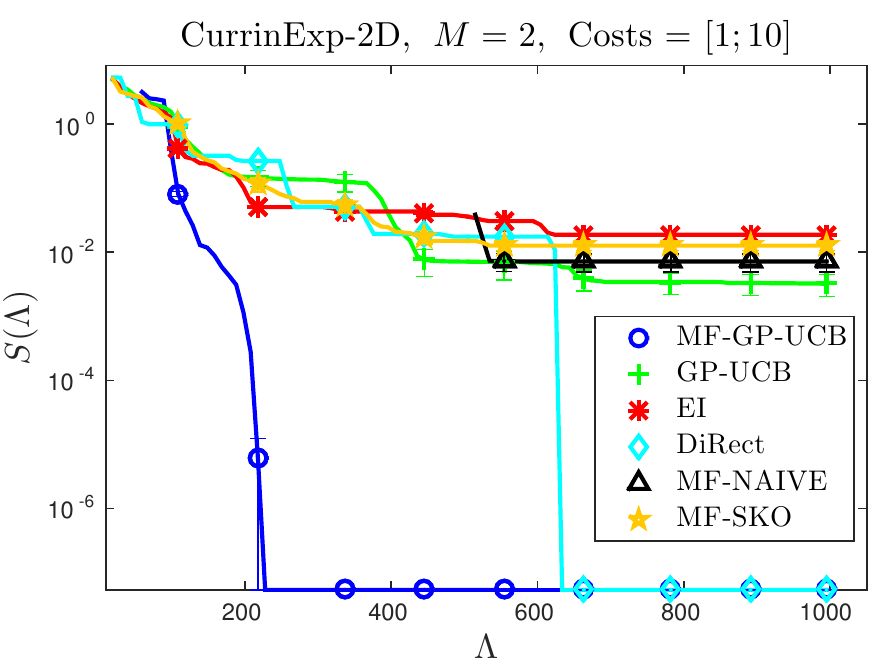} \hspace{\imhspthree}
}
\subfigure{
  \includegraphics[width=\imarrwthree]{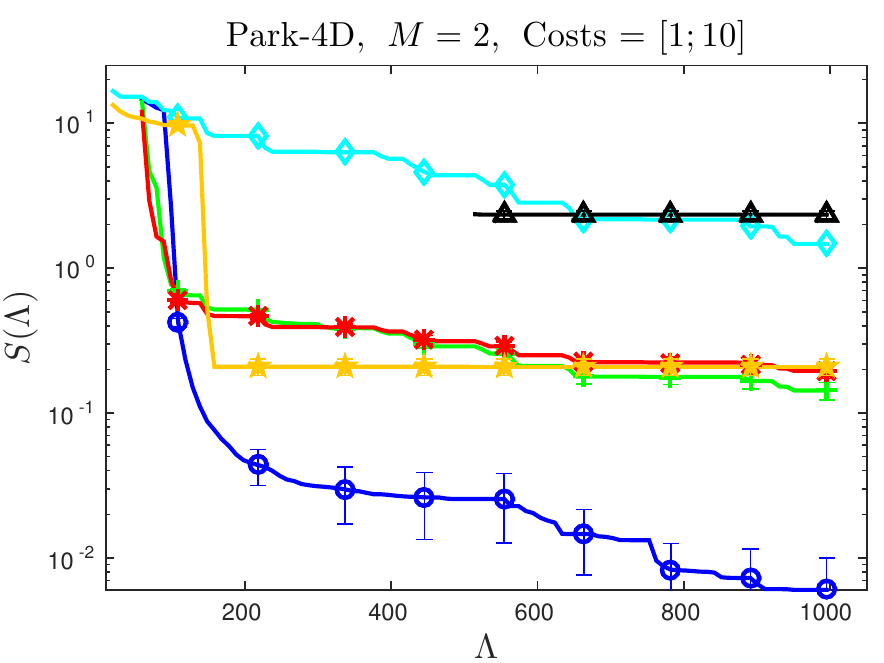} \hspace{\imhspthree}
}
\subfigure{
  \includegraphics[width=\imarrwthree]{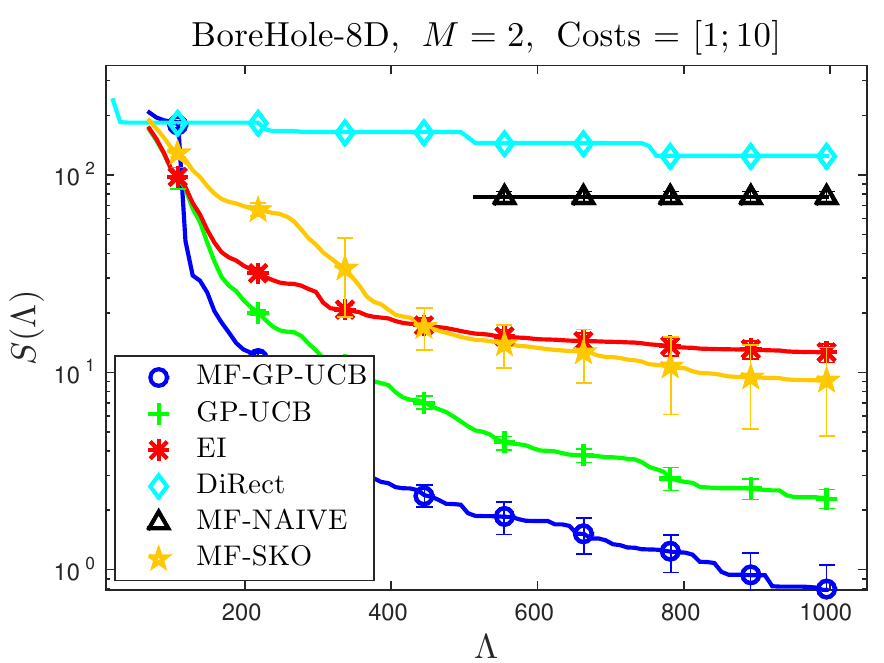} \hspace{\imhspthree}
} \\[\imcaptionspace]
\hspace{\imhspthree}
\subfigure{
  \includegraphics[width=\imarrwthree]{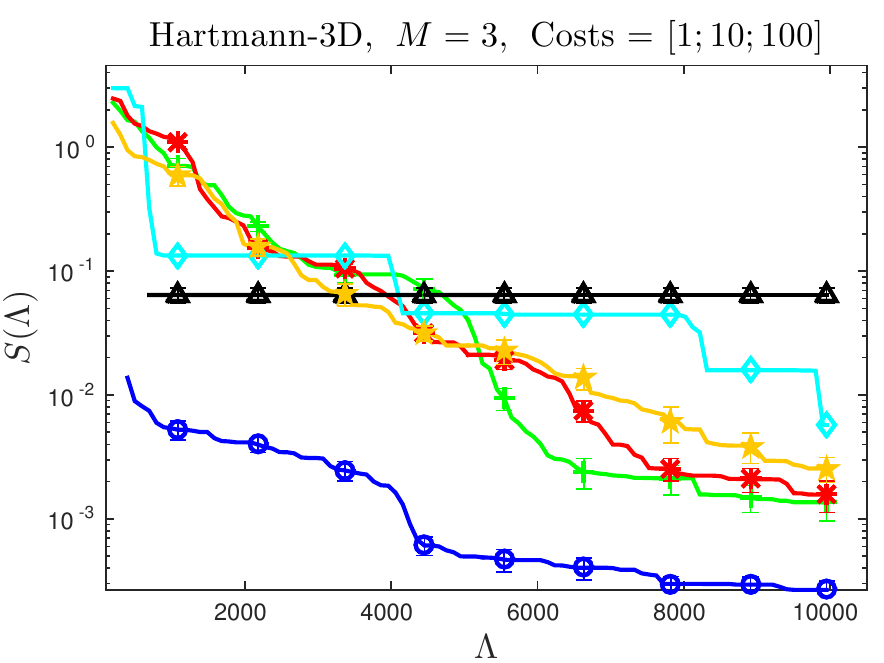} \hspace{\imhspthree}
}
\subfigure{
  \includegraphics[width=\imarrwthree]{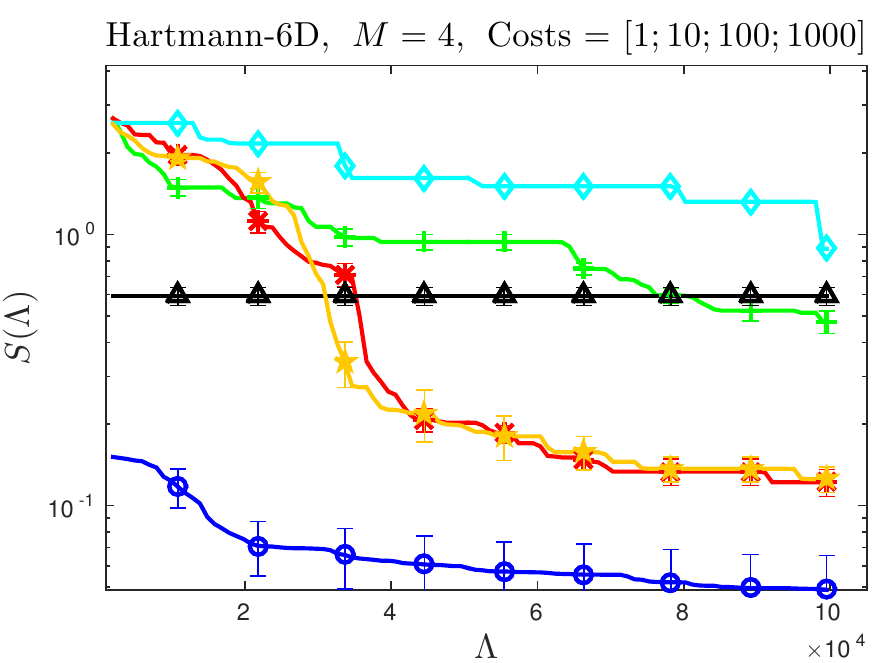} \hspace{\imhspthree}
}
\subfigure{
  \includegraphics[width=\imarrwthree]{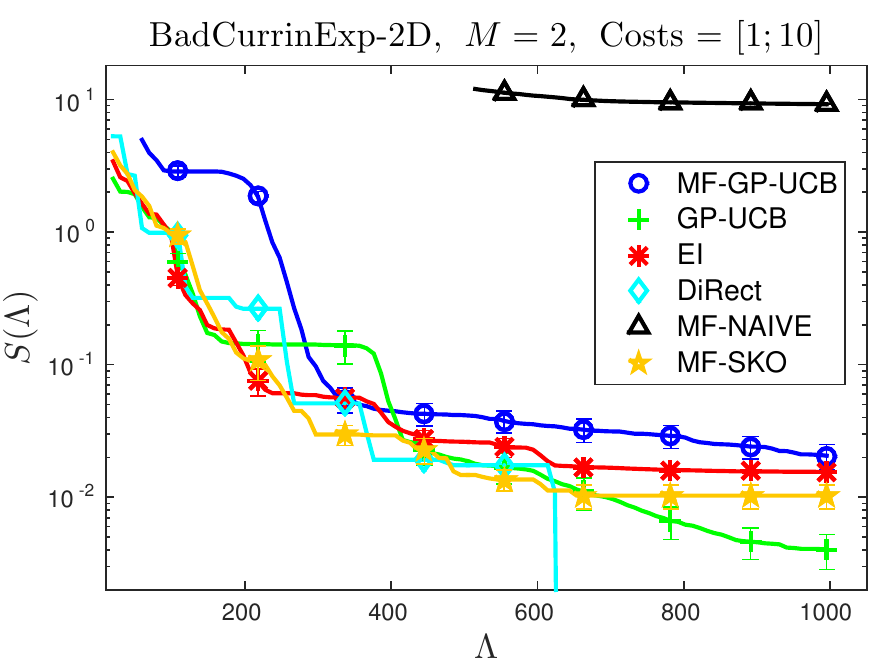} \hspace{\imhspthree}
} \\[-0.2in]
\caption[]{\small
The simple regret $S(\COST)$ against the spent capitcal $\COST$ on
the synthetic functions. 
\syntheticResultsCommon
\vspace{\imtextspace}
\label{fig:toySR}
}
\end{figure*}
}

\newcommand{\insertToyResultsCRFigure}{
\begin{figure*}
\centering
\hspace{\imhspthree}
\subfigure{
  \includegraphics[width=\imarrwthree]{figs/CurrinExp-CR} \hspace{\imhspthree}
}
\subfigure{
  \includegraphics[width=\imarrwthree]{figs/Park1-CR} \hspace{\imhspthree}
}
\subfigure{
  \includegraphics[width=\imarrwthree]{figs/BoreHole-CR} \hspace{\imhspthree}
} \\[\imcaptionspace]
\hspace{\imhspthree}
\subfigure{
  \includegraphics[width=\imarrwthree]{figs/Hartmann3-CR} \hspace{\imhspthree}
}
\subfigure{
  \includegraphics[width=\imarrwthree]{figs/Hartmann-CR} \hspace{\imhspthree}
}
\subfigure{
  \includegraphics[width=\imarrwthree]{figs/BadCurrinExp-CR} \hspace{\imhspthree}
} \\[\imcaptionspace]
\caption[]{\small
The cumulative regret $R(\COST)$ against the spent capitcal $\COST$ on
the synthetic functions. 
\syntheticResultsCommon
\vspace{\imtextspace}
\label{fig:toyCR}
}
\end{figure*}
}

\newcommand{\insertRealResultsFigure}{
\begin{figure*}
\centering
\hspace{\imhspthree}
\subfigure{
  \includegraphics[width=\imarrwthree]{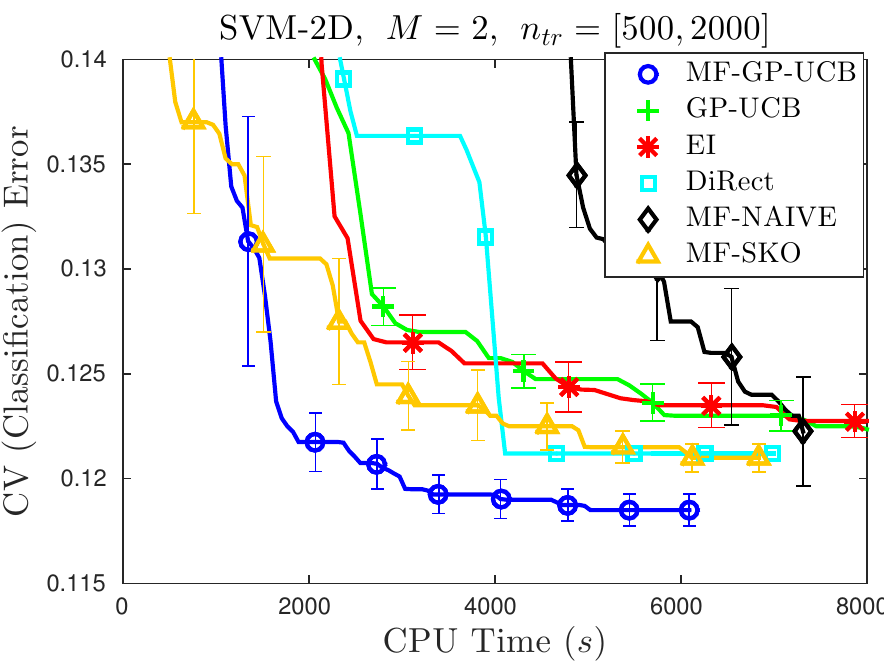} \hspace{\imhspthree}
}
\subfigure{
  \includegraphics[width=\imarrwthree]{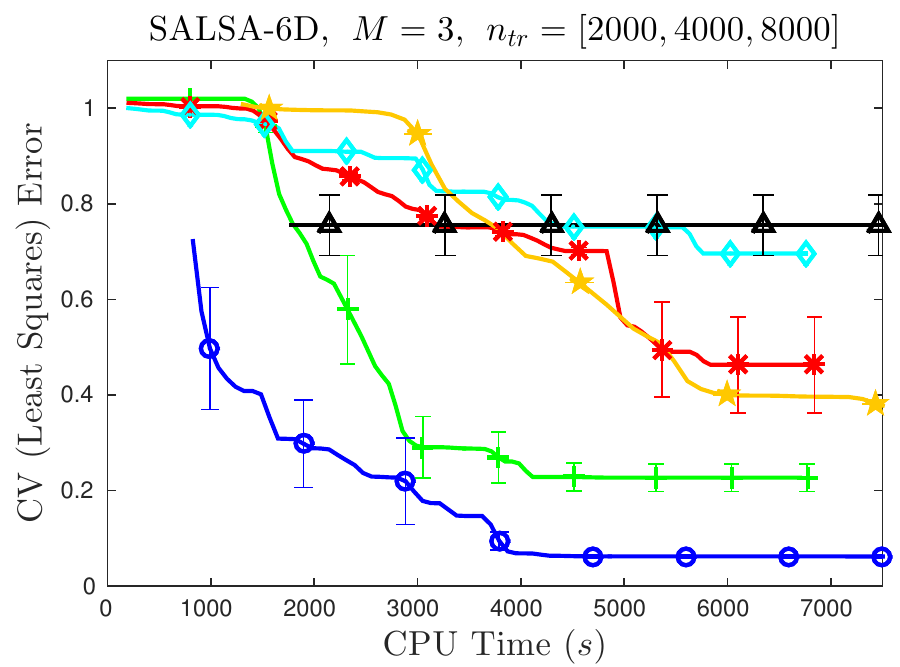} \hspace{\imhspthree}
}
\subfigure{
  \includegraphics[width=\imarrwthree]{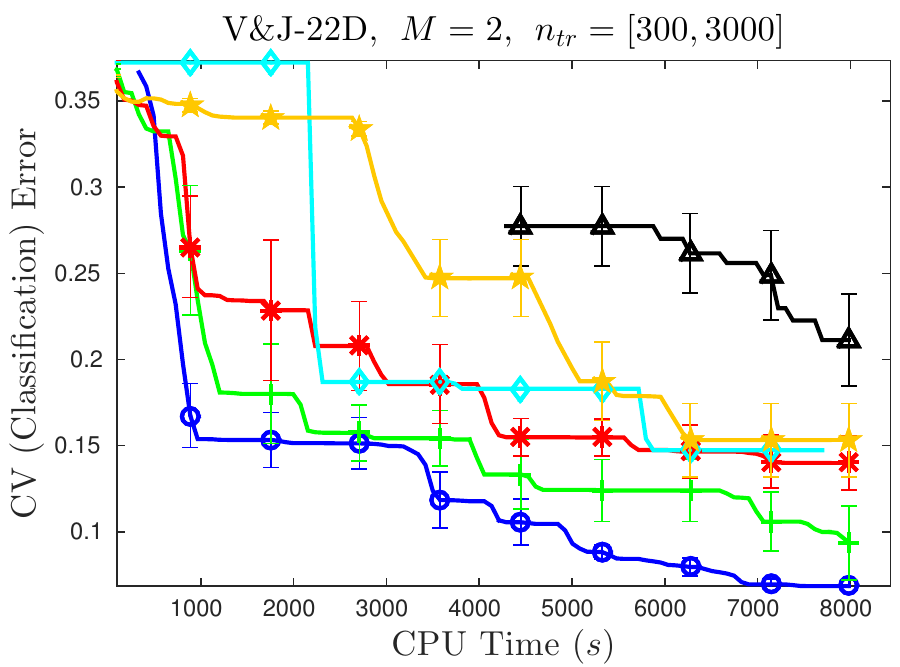} \hspace{\imhspthree}
} \vspace{-0.15in}
\caption[]{\small
\label{fig:hptuning}
% Results on hyper-parameter tuning experiments. The title states the experiment,
% dimensionality (number of hyperparameters) and training set size
% at each fidelity. 
% All curves were produced by averaging over 10 experiments.
% The curve lengths are different in time as we ran each method for a
% pre-specified number of iterations and they concluded at different times.
Results on the hyper-parameter tuning experiments. The title states the experiment,
dimensionality (number of hyperparameters) and training set size
at each fidelity. 
All curves were produced by averaging over 10 experiments.
The error bars indicate one standard error.
The lengths of the curves are different in time as we ran each method for a
pre-specified number of iterations and they concluded at different times.
\vspace{\imtextspace}
\vspace{-0.20in}
}
\end{figure*}
}

\newcommand{\insertRealAll}{
\begin{figure*}
\centering
\subfigure{
  \includegraphics[width=\imarrwtwo]{figs/svm} \hspace{\imhsptwo}
}
\subfigure{
  \includegraphics[width=\imarrwtwo]{figs/salsa}
}
\\[0.2in]
\subfigure{
  \includegraphics[width=\imarrwtwo]{figs/vj} \hspace{\imhsptwo}
} 
\subfigure{
  \includegraphics[width=\imarrwtwo]{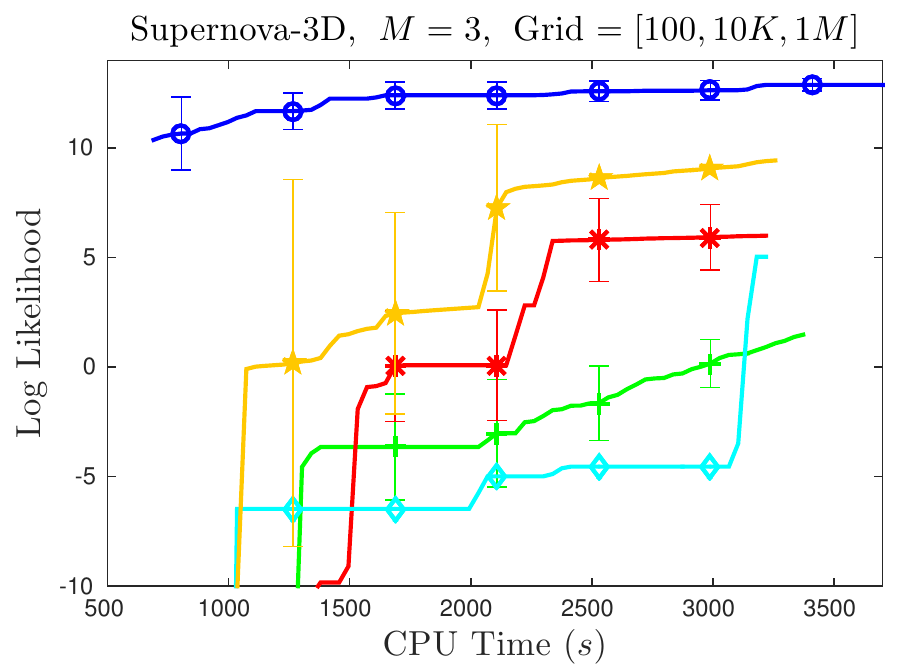}
}
\vspace{-0.1in}
\caption[]{\small
\label{fig:realAll}
Results on the real experiments. The first three figures are hyper-parameter tuning
tasks while the last is an astrophysical maximum likelihood problem.
The title states the experiment, dimensionality (number of hyper-parameters or
cosmological parameters) and the number of fidelities.
For the three hyper-parameter tuning tasks we plot the best cross validation error (lower is
better) and for the astrophysics task we plot the highest log likelihood (higher is
better).
For the hyper-parameter tuning tasks we obtained the lower fidelities by using smaller
training sets, indicated by $n_{tr}$ in the figures and for the astrophysical problem
we used a coarser grid for numerical integration, indicated by ``Grid''.
\mfnaives is not visible in the last experiment because it performed very poorly.
All curves were produced by averaging over 10 experiments.
The error bars indicate one standard error.
The lengths of the curves are different in time as we ran each method for a
pre-specified number of iterations and they concluded at different times.
\vspace{\imtextspace}
\vspace{-0.20in}
}
\end{figure*}
}

\newcommand{\insertSNFigure}{
  \begin{wrapfigure}{L}{2.1in}
  \centering
  \vspace{-0.2in}
  \includegraphics[width=\imarrwthree]{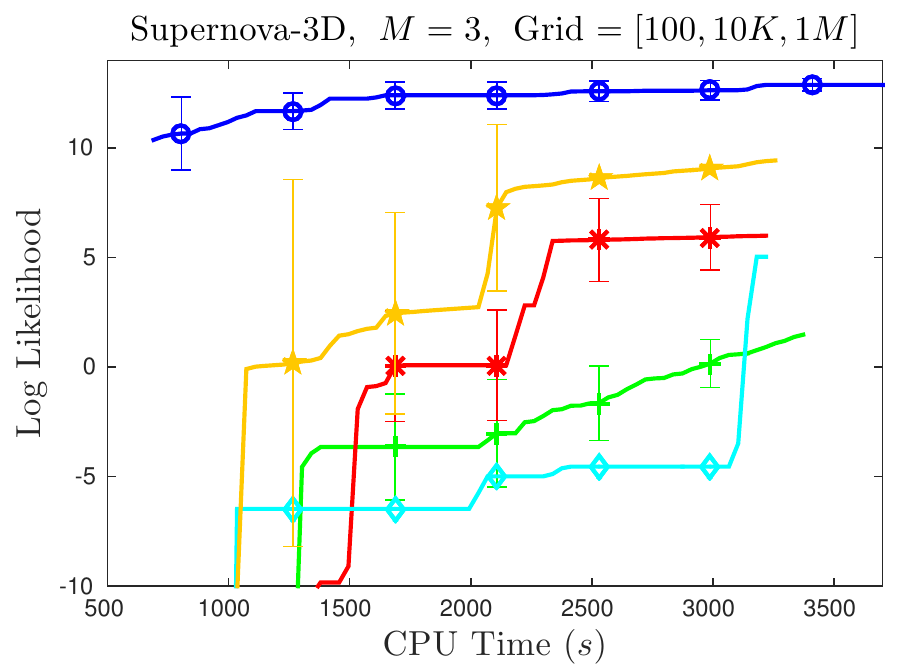}
  \vspace{\imcaptionspace}
  \caption{\small
  \label{fig:davisSN}
  Results on the supernova inference problem. The $y$-axis is the log likelihood
so higher is better.
\mfnaives is not visible as it performed very poorly.
%  See caption in Fig.~\ref{fig:hptuning} for more details. \hspace{-0.1in}
  \vspace{-0.2in}
  }
  \end{wrapfigure}
}

\newcommand{\insertSNAppFigure}{
% \begin{figure}
% \centering
% \vspace{\imcaptionspace}
% \caption{\small
% Results on the Type Ia Supernova dataset.
% \vspace{-0.07in}
% \label{fig:gpucb}
% }
% \end{figure}
\begin{figure}
\centering
  \begin{minipage}[c]{2.26in}
    \includegraphics[width=\imarrwthree]{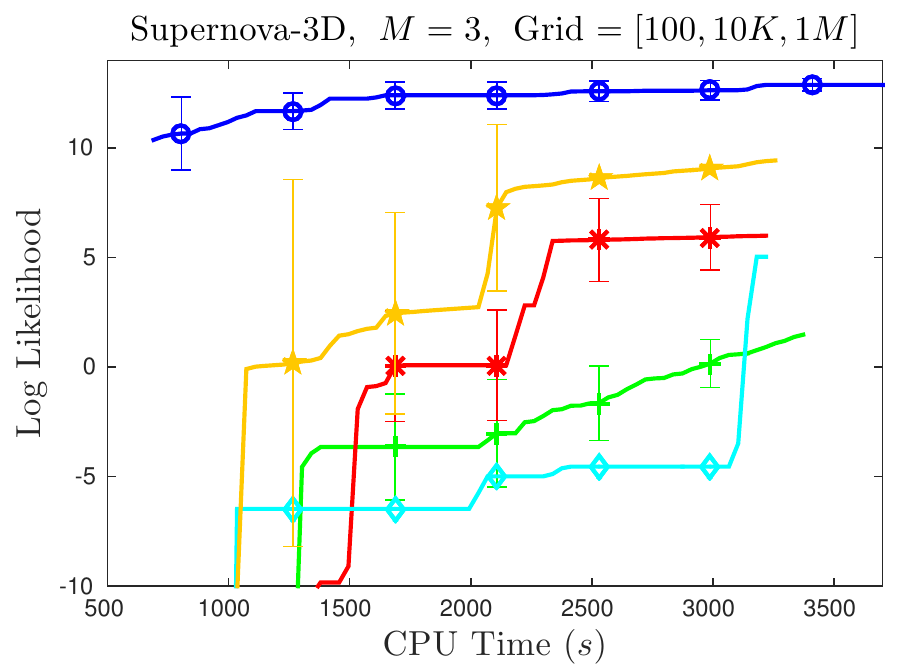}
  \end{minipage} \hspace{0.1in}
  \begin{minipage}[l]{4.0in}
  \vspace{-0.2in}
    \caption{
Results on the Cosmological maximum likelihood inference problem using Type Ia
supernovae data.
The curves show the maximum value of the log likelihood achieved by each method
(higher is better). 
The figure was produced by averaging over $10$ experiments and the error bars
indicate one standard error.
The curves for most methods are quite noisy since the range of the function was 
quite large ($\approx-7000$ to $\approx 13.5$) and hence there is a lot of variance. 
On the other hand, \mfgpucbs seems to be less
noisy since it has already gauged the landscape of $\func$ well from the lower
fidelity queries and used the high fidelity at a small region.
\rands is not visible in the graph as it performed poorly.
    } 
\label{fig:davisSNApp}
  \end{minipage}
\end{figure}
}

\newcommand{\insertGPUCBFigure}{
\begin{figure}
\centering
\includegraphics[width=2.5in]{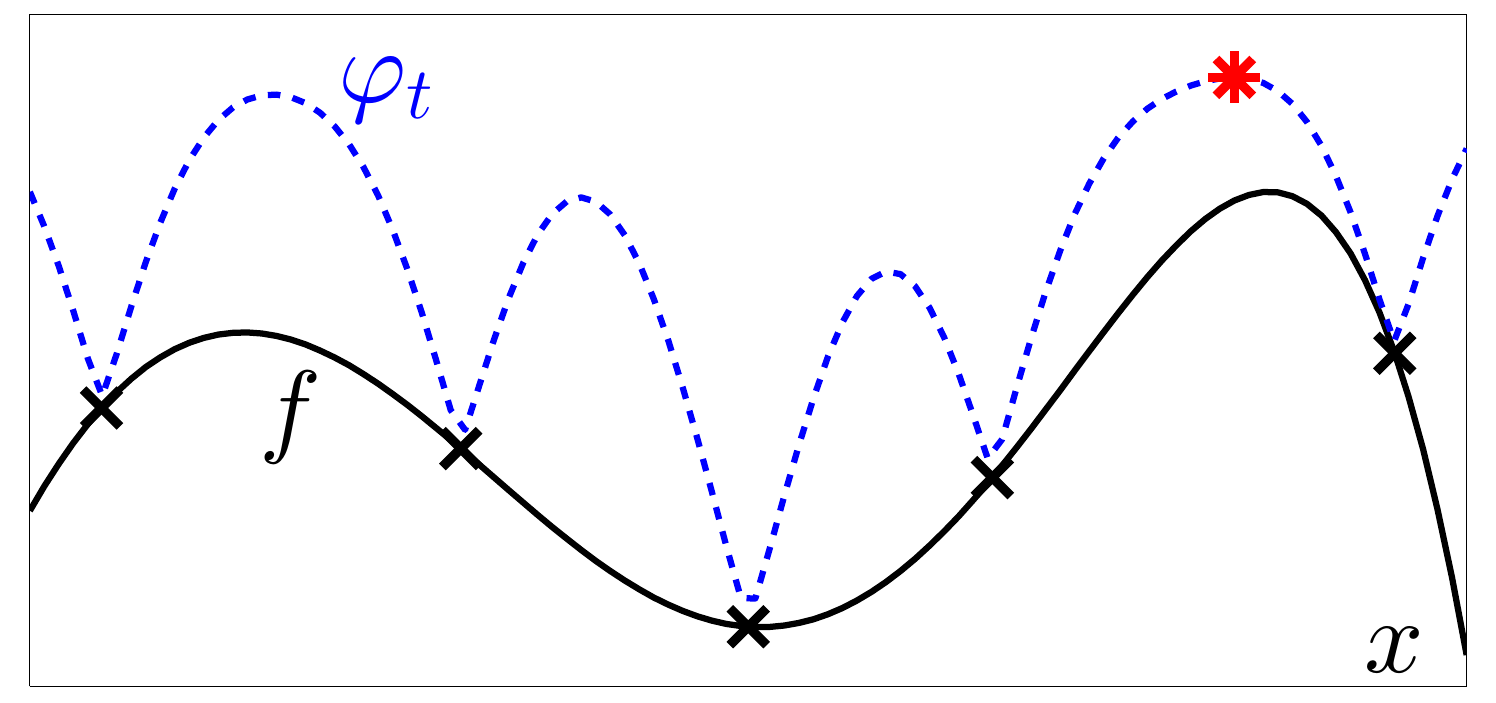}
\vspace{\imcaptionspace}
\caption{\small
Illustration of \gpucb. 
The solid black line is $\func(x)$ and the dashed blue line is $\ucbtx$. 
The observations until $t-1$ are shown as black crosses.
At time $t$,
we query at the maximiser $\xt = \argmax_{x\in\Xcal}\ucbtx$ shown via the red star.
\vspace{-0.07in}
\label{fig:gpucb}
}
\end{figure}
}

\newcommand{\insertKDEUCBFigure}{
\begin{figure}
\centering
\subfigure{
  \includegraphics[width=2.1in]{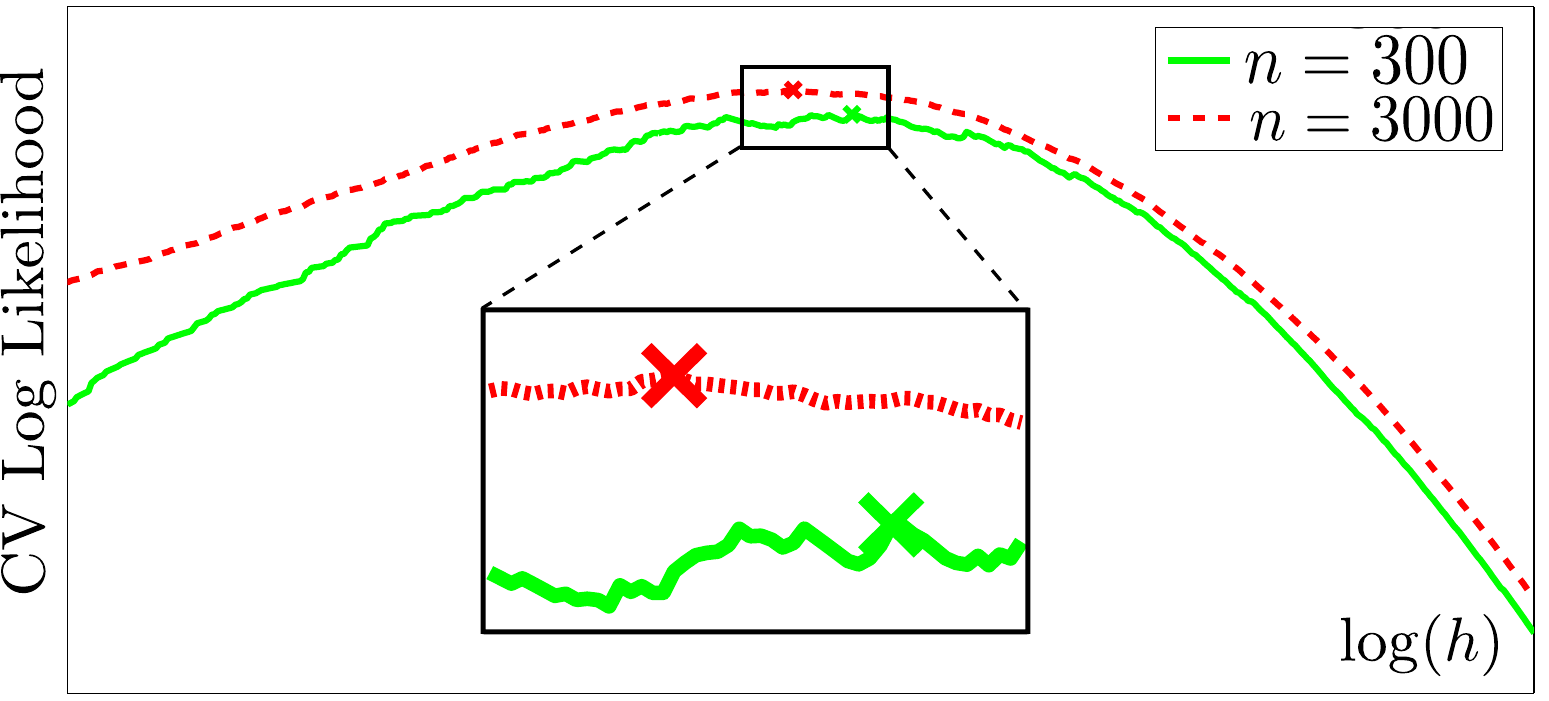} \hspace{0.2in}
}
\subfigure{
  \includegraphics[width=2.0in]{figs/gpucbSim}
}
\vspace{\imcaptionspace}
\caption{\small
\emph{Left:} Average CV log likelihood on datasets of size $300, 3000$
on a synthetic KDE task.
The crosses are the maxima.
\emph{Right:} 
Illustration of \gpucbs at time $t$. 
The figure shows $\func(x)$ (solid black line), the UCB $\ucbtx$ (dashed blue line)
and queries until $t-1$ (black crosses).
% The solid black line is $\func(x)$ and the dashed blue line is $\ucbtx$. 
% The observations until $t-1$ are shown as black crosses.
% At time $t$,
We query at $\xt = \argmax_{x\in\Xcal}\ucbtx$ (red star).\hspace{-0.2in}
\vspace{-0.2in}
\label{fig:gpucb}
\label{fig:kdeEg}
}
\end{figure}
}

\newcommand{\insertKDEFigure}{
\begin{figure}
\centering
\includegraphics[width=2.5in]{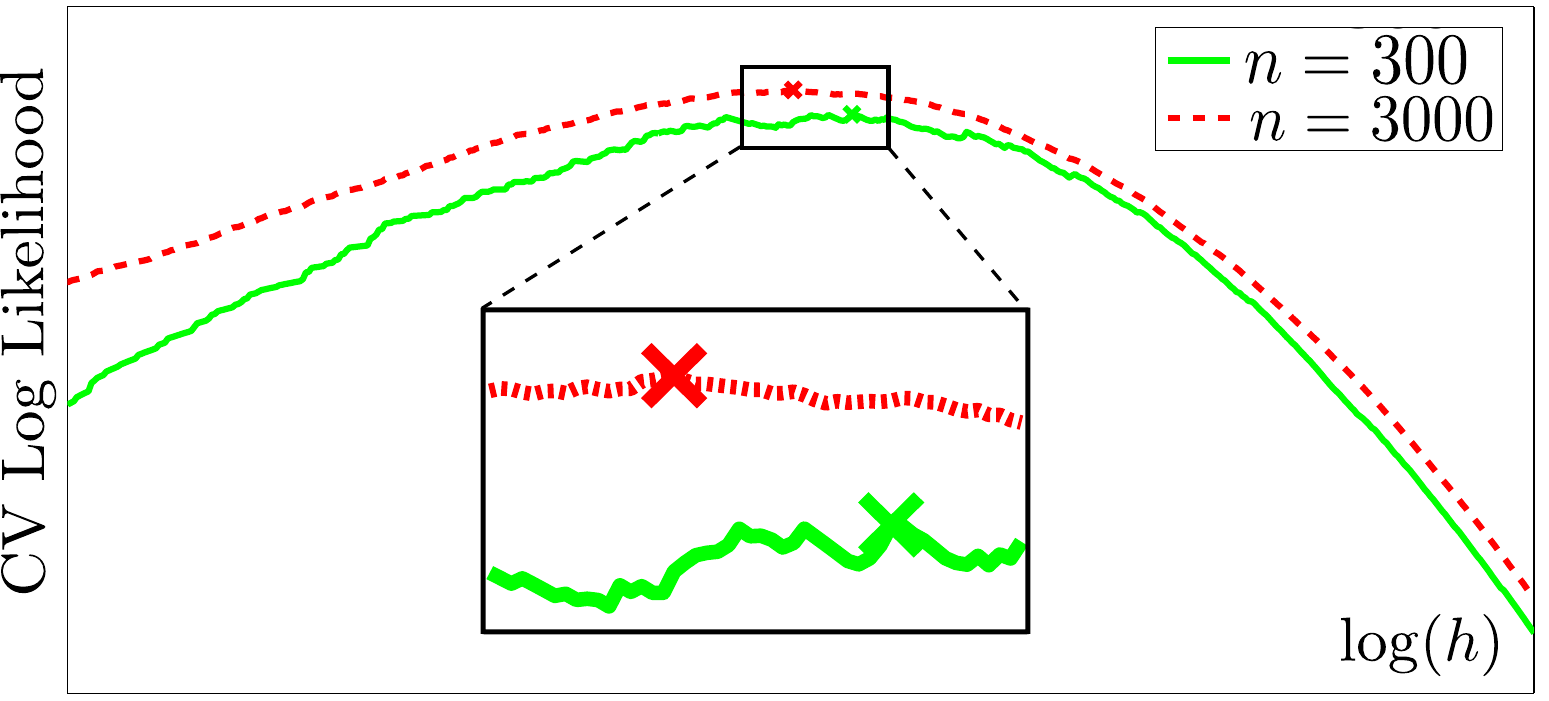}
\vspace{\imcaptionspace}
\caption{\small
\label{fig:kdeEg}
Average $5$-fold CV log likelihood on datasets of size $300, 3000$
on a synthetic KDE task.
The crosses are the maxima.
\vspace{-0.2in}
}
\end{figure}
}

\newcommand{\insertKDEHorFigure}{
\begin{figure}
\centering
  \begin{minipage}[c]{2.3in}
    \includegraphics[width=2.2in]{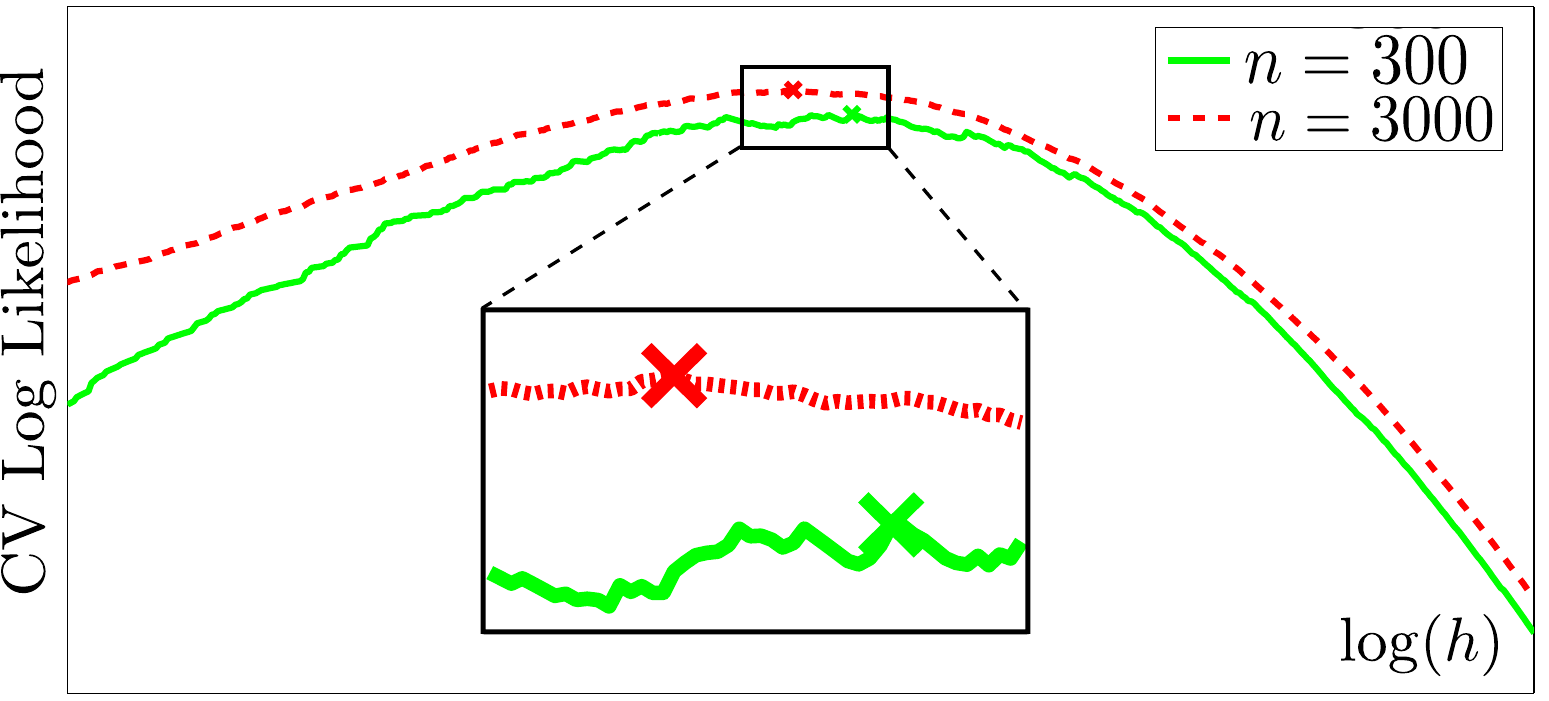}
  \end{minipage} \hspace{0.3in}
  \begin{minipage}[l]{3.0in}
  \vspace{-0.0in}
    \caption{
Average $5$-fold cross validation log likelihood on datasets of size $300$
and $3000$ on a synthetic kernel density estimation task.
The crosses are the maxima.
    }
  \label{fig:kdeEg}
  \end{minipage}
  \vspace{\imtextspace}
\end{figure}
}

\newcommand{\insertGPUCBHorFigure}{
\begin{figure}
\centering
  \begin{minipage}[c]{2.3in}
    \includegraphics[width=2.2in]{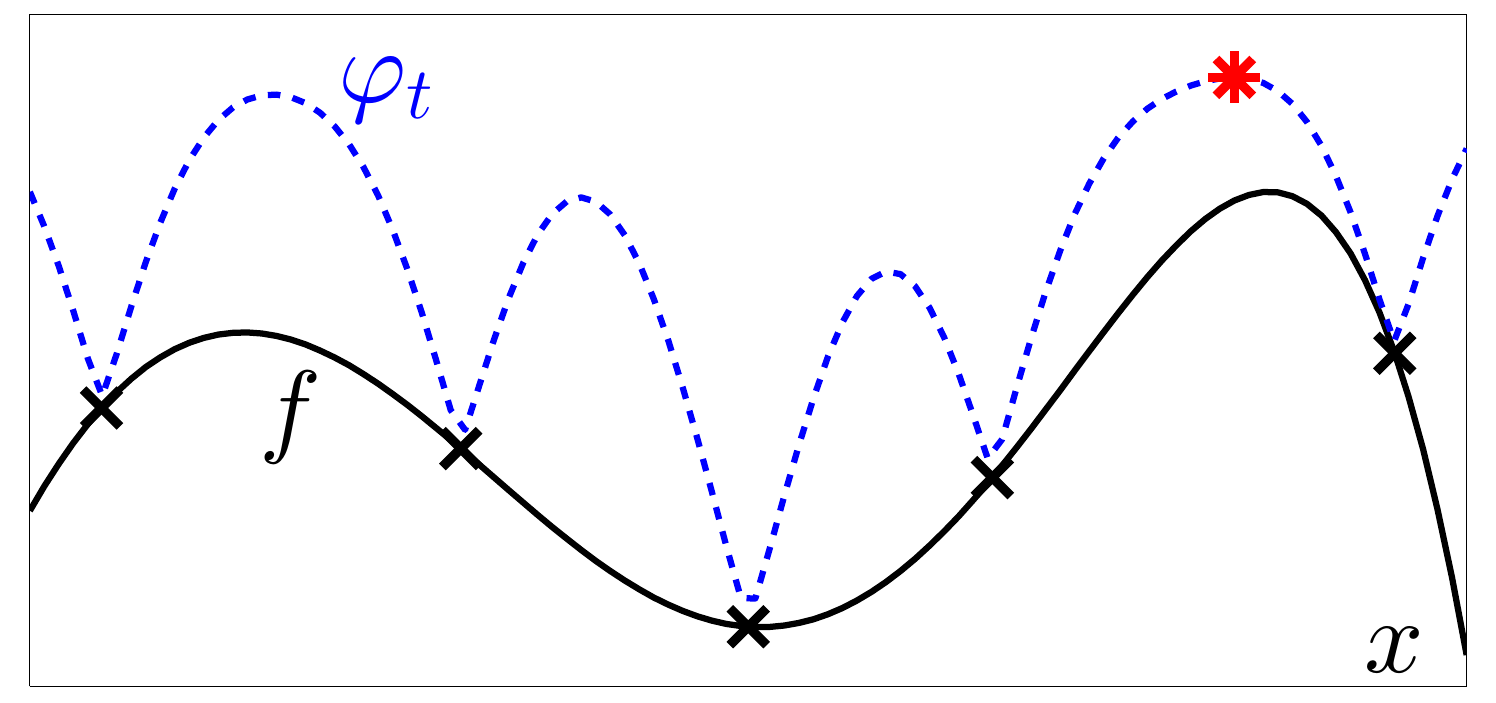}
  \end{minipage} \hspace{-0.05in}
  \begin{minipage}[l]{3.65in}
  \vspace{-0.0in}
    \caption{
Illustration of \gpucb. 
The solid black line is $\func$.
The dashed blue line is $\ucbt = \munontmo + \betath\sigmanontmo$ which
upper bounds $\func$ with high probability.
The observations until $t-1$ are shown as black crosses.
At time $t$,
we query at the maximiser $\xt = \argmax_{x\in\Xcal}\ucbtx$ (red star).
    } 
  \label{fig:gpucb}
  \end{minipage}
  \vspace{-0.2in}
\end{figure}
}

\newcommand{\insertKDEWrapFigure}{
  \begin{wrapfigure}{L}{2.1in}
  \centering
  \vspace{-0.2in}
  \includegraphics[width=2.0in]{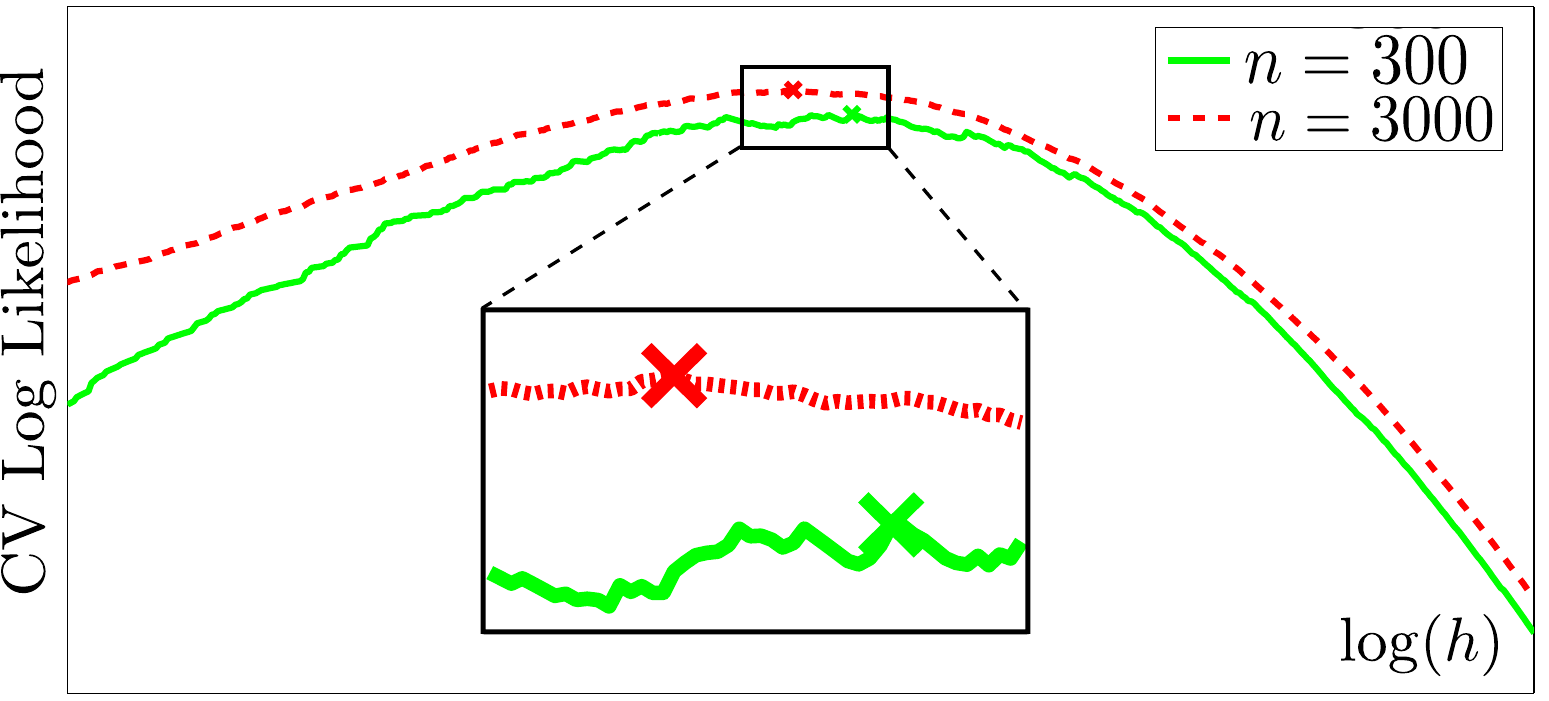}
  \vspace{\imcaptionspace}
  \caption{\small
  \label{fig:kdeEg}
Average $5$-fold CV log likelihood on datasets of size $300, 3000$
on a synthetic KDE task.
The crosses are the maxima.
  \vspace{-0.2in}
  }
  \end{wrapfigure}
}

\newcommand{\insertFigSets}{
\begin{figure}
\centering
  \begin{minipage}[c]{3.20in}
    \hspace{0.05in}
    \includegraphics[width=3.20in]{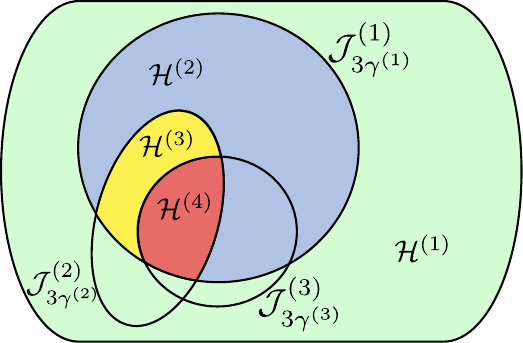}
  \end{minipage} \hspace{0.10in}
  \begin{minipage}[l]{2.60in}
  \vspace{-0.2in}
    \caption{
  Illustration of the partition $\Hcalm$'s for a $M=4$ fidelity problem. The sets $\Jcalmz$
are indicated next to their boundaries.
% $\Xcalone,\Xcaltwo,\Xcalmm{3},\Xcalmm{4}$
The sets $\Hcalone,\Hcaltwo,\Hcalmm{3},\Hcalmm{4}$
are shown in green, blue, yellow and red respectively. 
Most of the capital invested at points in $\Hcalm$ will be due to
queries to the $m$\ssth fidelity function $\funcm$.
%   An illustration for the sets $\Hcalm$'s would look similar.
    } 
\label{fig:Xcalms}
  \end{minipage}
%   \vspace{-0.1in}
\end{figure}
}

\newcommand{\insertFigFcalHcal}{
\begin{figure}
\centering
  \begin{minipage}[c]{2.3in}
    \includegraphics[width=2.6in]{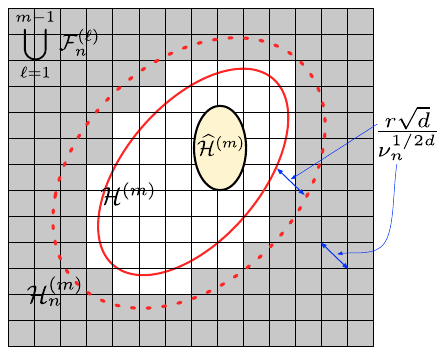}
  \end{minipage} \hspace{0.10in}
  \begin{minipage}[l]{3.5in}
%   \vspace{-0.2in}
    \caption{ \small
Illustration of the sets $\{\Fcalln\}_{\ell=1}^{m-1}$ with respect to $\Htcalm$.
The grid represents a $r\sqrt{d}/n^{1/(2d)}$ covering of $\Xcal$.
The yellow region is $\Htcalupm$. The area enclosed by the solid red line (excluding
$\Htcalupm$) is $\Htcalm$. $\Htncalm$, shown by a dashed red line,
is obtained by dilating $\Htcalm$ by $r\sqrt{d}/n^{\alpha/2d}$.
The grey shaded region represents $\bigcup_{\ell=1}^{m-1}\Fcalln$.
By our definition, $\bigcup_{\ell=1}^{m-1}\Fcalln$ contains the cells which are
entirely outside $\Htcalm$. However, the inflation $\Htncalm$ is such that
$\Htcalupm \,\cup\, \Htncalm \,\cup\, \bigcup_{\ell=1}^{m-1}\Fcalln = \Xcal$.
We further note that as $n\rightarrow\infty$, $\Htncalm\rightarrow\Htcalm$.
    } 
\label{fig:FcalHcal}
  \end{minipage}
  \vspace{-0.05in}
\end{figure}
}

\newcommand{\insertChallengesFigure}{
\newcommand{\chalimwidth}{2.5in}
\newcommand{\chalimspace}{0.2in}
\begin{figure}
\centering
\subfigure[]{
\includegraphics[width=\chalimwidth]{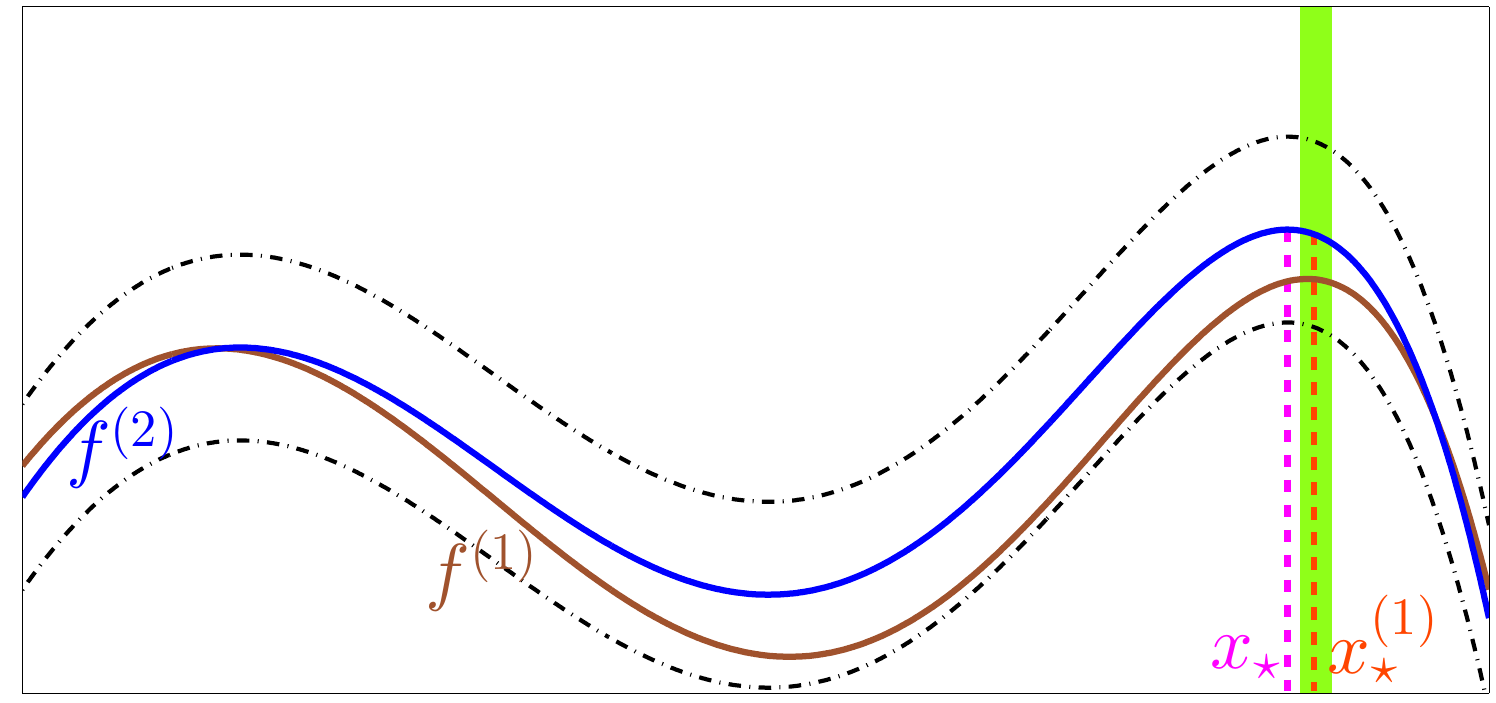}\hspace{\chalimspace}
\label{fig:f1f2regSmall}
}
\subfigure[]{
\includegraphics[width=\chalimwidth]{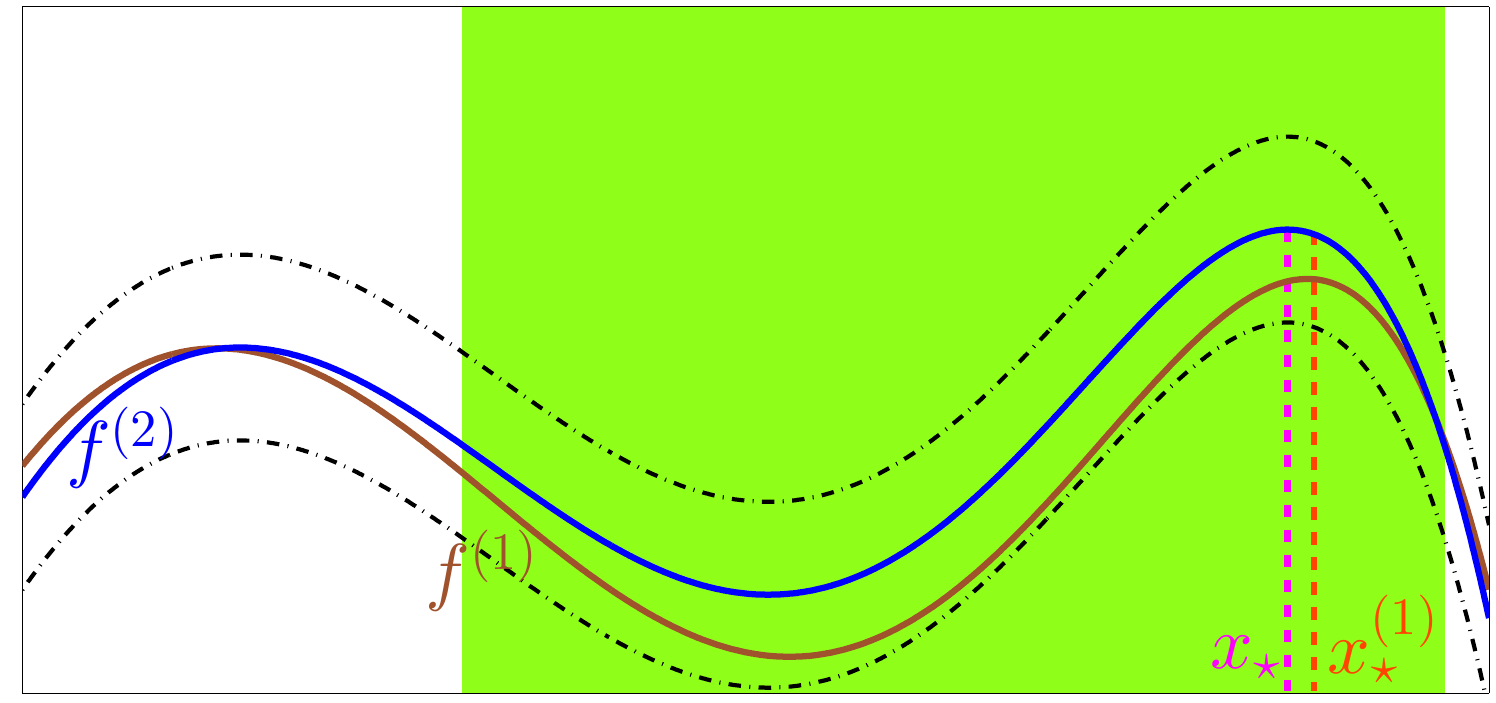}
\label{fig:f1f2regLarge}
}
\\[-0.1in]
\subfigure[]{
\includegraphics[width=\chalimwidth]{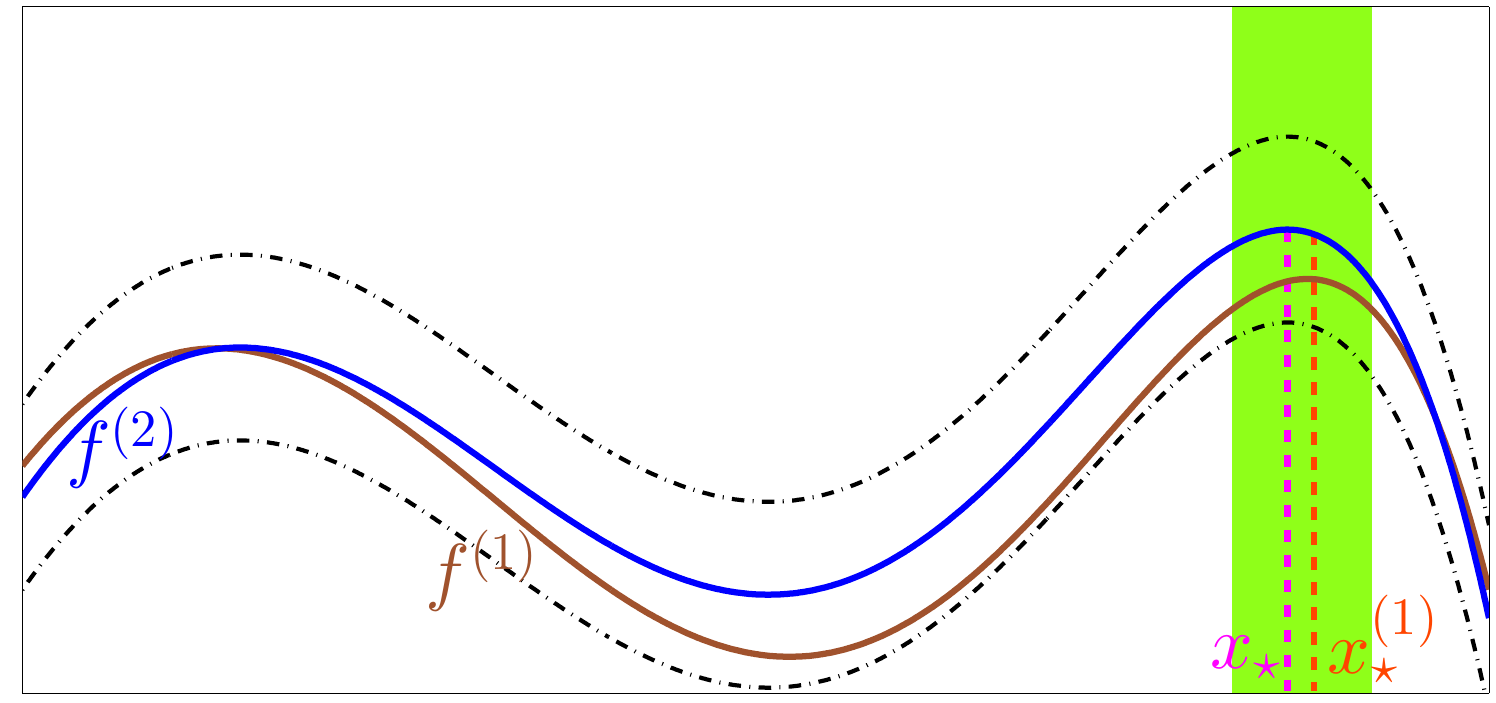}\hspace{\chalimspace}
\label{fig:f1f2regCorrect}
}
\subfigure[]{
\includegraphics[width=\chalimwidth]{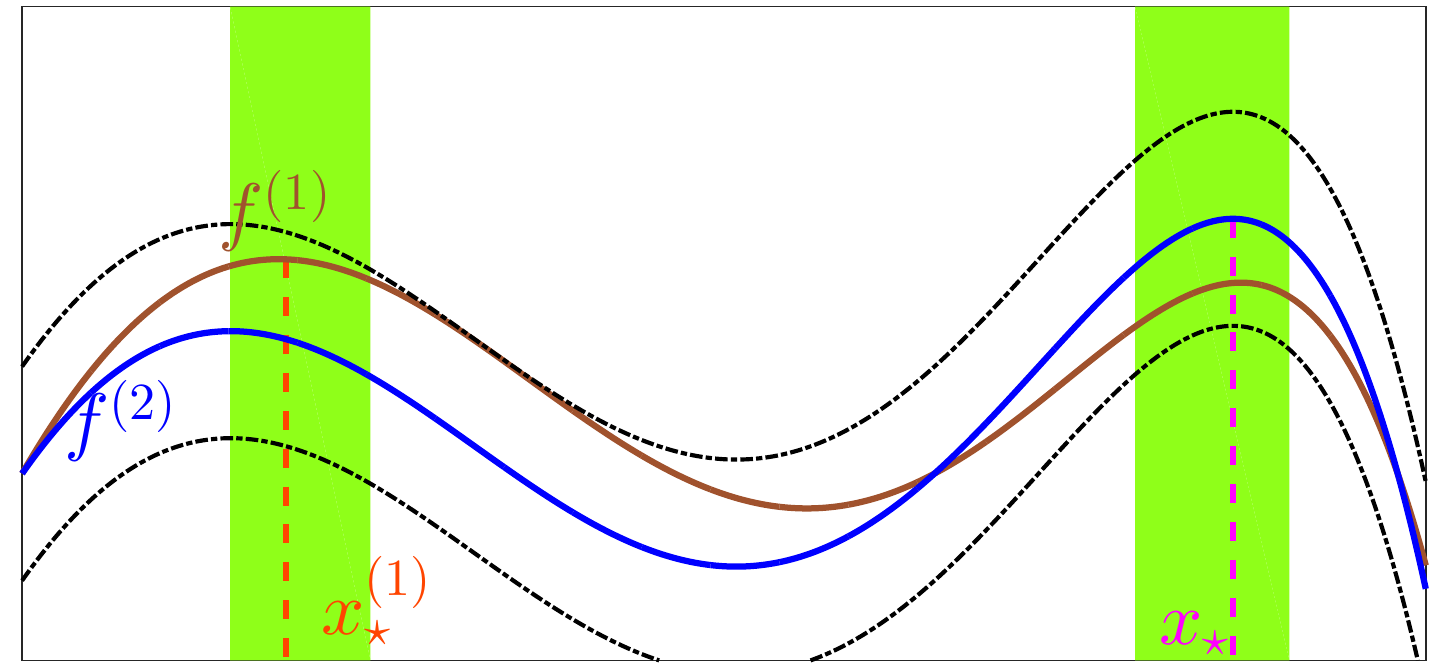} 
\label{fig:f1f2regDouble}
}
% \hspace{\chalimspace}
\vspace{-0.1in}
\caption{\small
\label{fig:challenges}
An illustration of the challenges in multi-fidelty optimisation.
See main text.
}
\vspace{-0.1in}
\end{figure}
}

\newcommand{\insertFigBoreHoleLambdaComp}{
\begin{figure}
\centering
  \begin{minipage}[c]{2.8in}
    \includegraphics[width=2.9in]{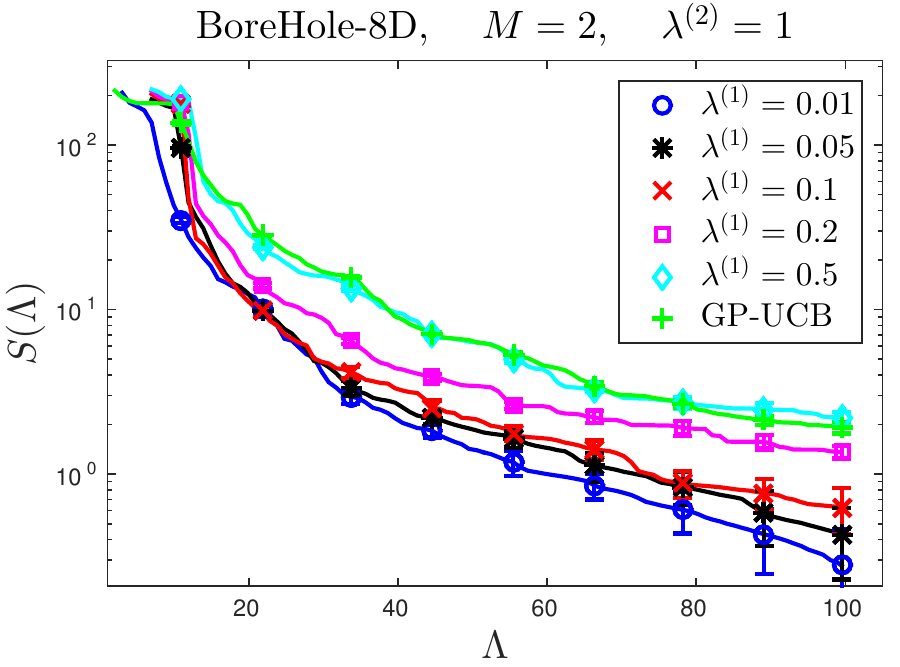}
  \end{minipage} \hspace{0.20in}
  \begin{minipage}[l]{2.8in}
%   \vspace{-0.2in}
    \caption{ \small
    The performance of our implementation of \mfgpucbs for different values of
    $\costone$ in the $2$ fidelity Borehole experiment.
    Our implementation uses the techniques and heuristics described in
    Section~\ref{sec:implementation}.
    In all experiments we used $\costtwo=1$.
    We have also shown the curve for \gpucbs for reference.
    } 
\label{fig:bhlambda}
  \end{minipage}
  \vspace{-0.05in}
\end{figure}
}

\newcommand{\insertFigGammaHartmannComp}{
\begin{figure}
\centering
  \begin{minipage}[c]{2.8in}
    \includegraphics[width=2.9in]{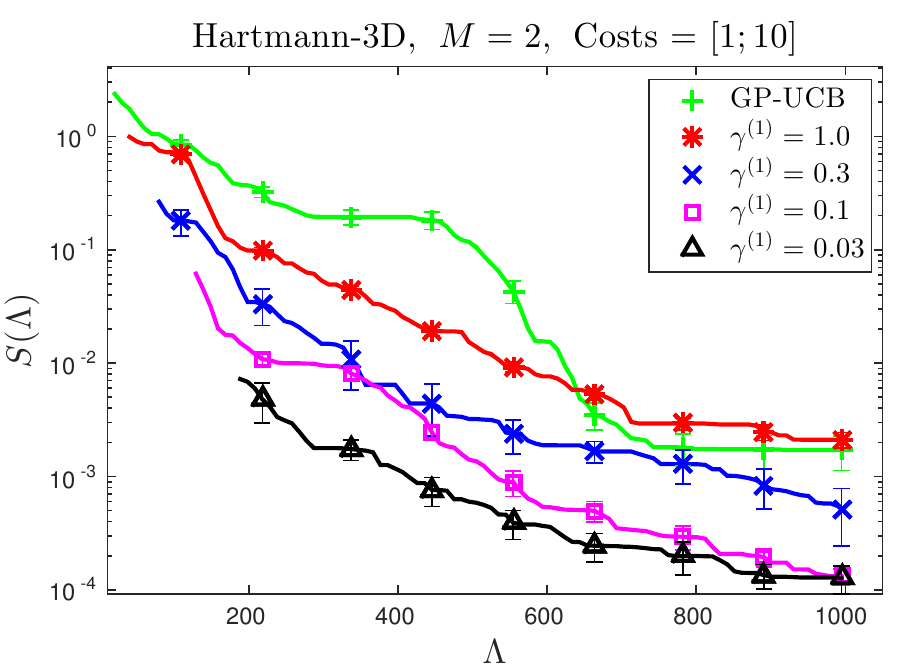}
  \end{minipage} \hspace{0.20in}
  \begin{minipage}[l]{2.8in}
%   \vspace{-0.2in}
    \caption{ \small
    The performance of \mfgpucbs for different choices of \emph{fixed} threshold 
    values $\gammaone$.
    The curves were averaged over $20$ independent runs and
    in this figure, they start when at least $10$ of the $20$ runs have queried
    at least once at the top (second) fidelity.
    This experiment was run on the $3-$dimensional Hartmann function in the two fidelity
    set up where $\zetaone\approx0.112$.
    The true $\zetaone$ value was made known to \mfgpucb.
%     As the threshold value decreases, the curves start later in the figure indicating
%     that \mfgpucbs spends more time at the approximation before proceeding to
    }
\label{fig:hartmanngamma}
  \end{minipage}
  \vspace{-0.05in}
\end{figure}
}

\newcommand{\insertzetaHmFigure}{
\newcommand{\zetaHmimwidth}{2.7in}
\newcommand{\zetaHmimspace}{0.2in}
\begin{figure}
\centering
\subfigure{
\includegraphics[width=\zetaHmimwidth]{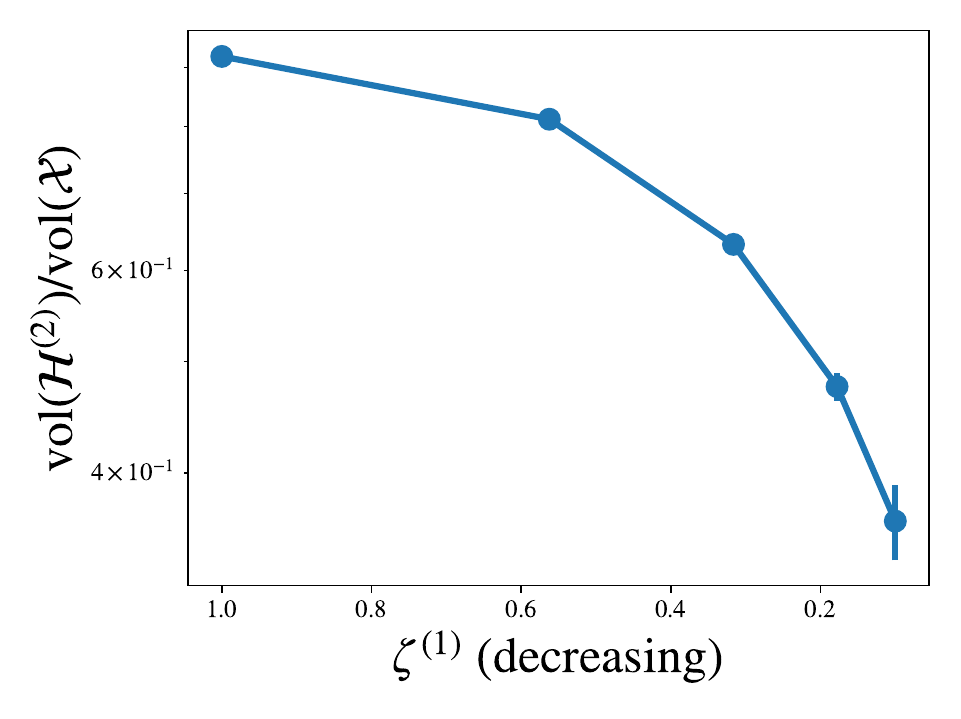}\hspace{\zetaHmimspace}
\label{fig:f1f2regSmall}
}
\subfigure{
\includegraphics[width=\zetaHmimwidth]{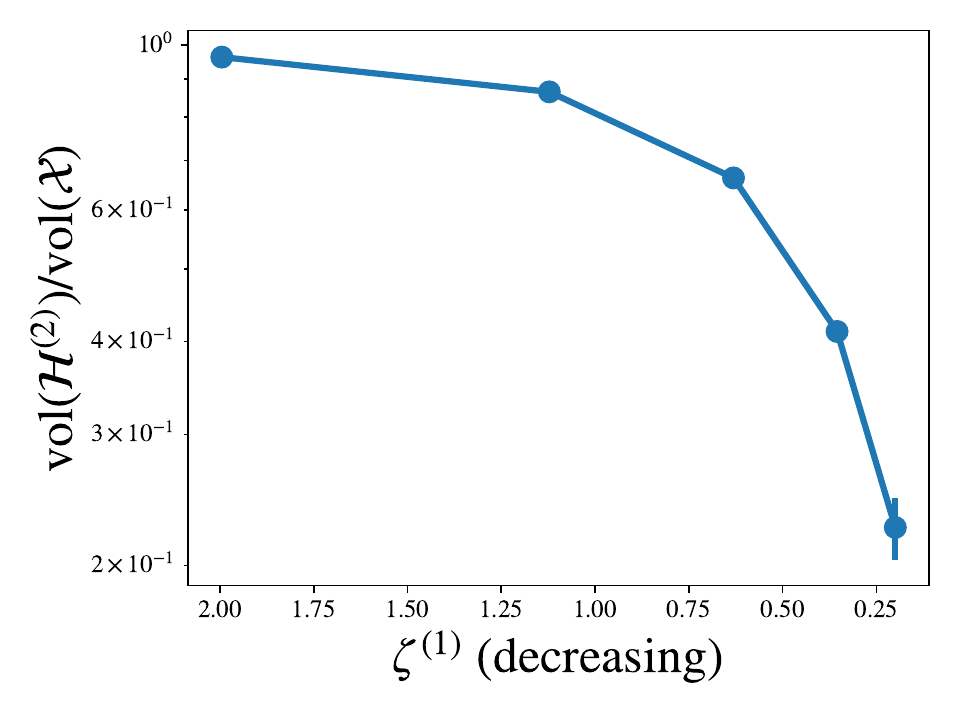}
\label{fig:f1f2regLarge}
}
% \hspace{\zetaHmimspace}
\vspace{-0.2in}
\caption{\small
\label{fig:zetaHmIllus}
Empirically computed values for the ratio $\vol(\Hcaltwo)/\vol(\Xcal)$ for a
one dimensional (left) and two dimensional (right) 2-fidelity problem.
For this, the samples $\funcone,\functwo$ were generated using the generative mechanism
of Section~\ref{sec:mfogenprocess}, under the stipulated value for $\zetaone$.
In both cases, we used an SE kernel with bandwidth $1$ and scale parameter $1$.
The $y$-axis is the mean value for the ratio over several samples and
the $x$-axis is $\zetaone$. In both cases, we used $\gammaone=\zetaone/3$,
and approximated the continuous domain with a uniform grid of size $10^4$.
The figure indicates that as the approximation improves, i.e. $\zetaone$ decreases,
the ratio decreases and consequently, we get better bounds.
}
\vspace{-0.1in}
\end{figure}
}

\begin{abstract}%   <- trailing '%' for backward compatibility of .sty file
In many scientific and engineering applications, we are tasked with the maximisation
of an expensive to evaluate black box function $\func$.
Traditional settings for this
problem assume just the availability of this single function.
However, in many cases, cheap approximations to $\func$ may be obtainable. 
For example, the expensive real world behaviour of a robot can be approximated 
by a cheap computer simulation.
We can use these approximations to eliminate low function value 
regions cheaply and use the expensive evaluations of $\func$
in a small but promising region and speedily identify the optimum.
We formalise this task as a \emph{multi-fidelity} bandit problem where the target
function and its approximations are sampled from a Gaussian process.
We develop \mfgpucb, a novel method based on upper confidence bound techniques.
In our theoretical analysis we demonstrate that it
exhibits precisely the above behaviour
and achieves better bounds on the regret than strategies which ignore multi-fidelity
information.
Empirically, \mfgpucbs outperforms such naive strategies and other multi-fidelity methods
 on several synthetic and real experiments.
\end{abstract}

% \begin{keywords}
%   Multi-fidelity optimisation, Bandits, Bandit Optimisation,
%   Bayesian optimisation, Gaussian processes.
% \end{keywords}

\section{Introduction}
\label{sec:intro}

In stochastic bandit optimisation, we wish to optimise a function
$\func:\Xcal\rightarrow \RR$ by sequentially querying it and obtaining 
\emph{bandit feedback}, i.e. when we query at any $x\in\Xcal$, we observe a possibly noisy 
evaluation of $\func(x)$. $\func$ is typically expensive
and the goal is to identify its maximum while keeping the number of
queries as low as possible. 
Some applications are hyper-parameter tuning in expensive machine learning
algorithms, optimal policy search in complex systems, online advertising, scientific
% experiments~\cite{parkinson06wmap3,hutter11smac,martinez07robotplanning,snoek12practicalBO}.
experiments, and statistical tasks such as collaborative filtering and
clustering~\citep{parkinson06wmap3,snoek12practicalBO,martinez07robotplanning,%
li2016collaborative,li2016art,korda2016distributed,kar2016online,%
gentile2016context}.
Historically, bandit problems were studied in settings where the goal is to
maximise the cumulative reward of all queries to the payoff instead of just finding 
the maximum. 
Applications in this setting include clinical trials and
online advertising.

Conventional methods in these settings assume access to only this single
expensive function of interest $\func$. We will collectively refer to them as
\emph{single fidelity} methods.
In many practical problems however, cheap
approximations to $\func$ might be available.
For instance, when tuning hyper-parameters of learning algorithms, the goal
is to maximise a cross validation score on a training set, which can be expensive
if the training set is large.
However validation curves tend to vary smoothly
with training set size; therefore, we can train and cross validate on small
subsets to approximate the validation accuracies of the entire dataset.
For a concrete example, consider kernel density estimation (KDE), 
where we need to tune the bandwidth $h$ of a kernel when using  a dataset of 
size $3000$.
Figure~\ref{fig:kdeEg} shows the average cross validation likelihood 
against $h$ for a dataset of size $n=3000$ and a smaller subset of size $n=300$.
Since the cross validation performance of a hyper-parameter depends on the training
set size~\citep{vapnik1998statistical},
we can obtain only a biased estimate of the cross validation performance with $3000$
points using a subset of size $300$.
% the $n=300$ curve can only provide a biased estimate for $n=3000$.
Consequently, the two maximisers are also different.
That said, the curve for
$n=300$ approximates the $n=3000$ curve quite well.
Since training and cross validation on small $n$ is cheap, we can use it
to eliminate bad values of the hyper-parameters
and reserve the expensive experiments with the entire dataset for the promising
hyper-parameter values (for example, boxed region in Figure~\ref{fig:kdeEg}). 

\insertKDEHorFigure

In the conventional treatment for online advertising, each query to $\func$ is, say,
the public display of an ad on the internet for a certain time period.
However, we could also choose smaller experiments by, say, confining the display to a
small geographic region and/or for shorter periods.
The estimate is biased, since users in different geographies are likely to have
different preferences, but will nonetheless be
useful in gauging the all round performance of an ad.
In optimal policy search in robotics and autonomous driving, 
vastly cheaper computer simulations are
used to approximate the expensive real world performance of the
system~\citep{urmson2008autonomous,cutler14mfsim}.
Scientific experiments can be approximated to varying degrees using less expensive
data collection, analysis, and computational techniques%
~\citep{parkinson06wmap3}.
% Cheap approximations can be similarly obtained in many applications in finance, on-line 
% advertising and industrial design problems.

In this paper, we cast these tasks as \emph{multi-fidelity bandit optimisation}
problems assuming
the availability of cheap approximate functions (fidelities) to the payoff $\func$.
\textbf{Our contributions} are:
% \vspace{-0.10in}
\newcommand{\insertContrVspace}{\vspace{-0.05in}}
\begin{enumerate} [leftmargin=0.2in]
\item We present a formalism for multi-fidelity bandit optimisation using
Gaussian process (GP) assumptions on $\func$ and its approximations.
 We develop a novel algorithm, Multi-Fidelity Gaussian Process Upper 
Confidence Bound (\mfgpucb) for this setting. %\hspace{-0.3in}
% \mfgpucbs is based on UCB ideas popularly used in bandit literature.
\insertContrVspace
\item Our theoretical analysis proves that \mfgpucbs explores the space $\Xcal$ at lower
fidelities and  uses the high fidelities in successively
smaller regions to converge on the optimum.
As lower fidelity queries are cheaper, \mfgpucbs has better upper
bounds on the regret than single
fidelity strategies which have to rely on the expensive function to explore
the entire space.
\insertContrVspace
% \item Empirically, we demonstrate that \mfgpucbs outperforms single fidelity methods 
%  on a series of synthetic examples, three
% hyper-parameter tuning tasks and one inference problem in Astrophysics.
\item We demonstrate that \mfgpucbs outperforms single fidelity methods and other
alternatives empirically, via a series of synthetic examples, three
hyper-parameter tuning tasks and one inference problem in astrophysics.
Our matlab implementation and experiments are available at
\incmtt{\small{\href{https://github.com/kirthevasank/mf-gp-ucb}{github.com/kirthevasank/mf-gp-ucb}}}.
\end{enumerate}

\subsection*{Related Work}

Since the seminal work by~\citet{robbins52seqDesign}, the multi-armed bandit problem
has been studied extensively in the $K$-armed
setting. Recently, there has been a
surge of interest in the optimism under uncertainty principle for $K$-armed bandits,
typified by upper confidence bound (UCB) 
methods~\citep{bubeck12regret,auer03ucb}.
UCB strategies have also been used in  bandit tasks with 
linear~\citep{dani08linearBandits} and 
GP~\citep{srinivas10gpbandits} payoffs.
There is a plethora of work on single fidelity methods for global optimisation both
with noisy and noiseless evaluations.
Some examples are branch and bound techniques such as dividing rectangles
(\direct), simulated annealing,
genetic algorithms  and
more~\citep{jones93direct,munos11soo,kawaguchi15bo,kirkpatrick83simAnnealing}.
A suite of single fidelity methods in the GP framework
closely related to our work is Bayesian Optimisation (BO).
While there are several techniques for
BO~\citep{mockus94bo,jones98expensive,thompson33sampling,lobato14pes},
of particular interest to us is the
Gaussian process upper confidence bound (\gpucb) algorithm
of~\citet{srinivas10gpbandits}. 

Many applied domains of research such as aerodynamics, industrial design and
hyper-parameter tuning have studied multi-fidelity
methods~\citep{huang06mfKriging,swersky2013multi,klein2015towards,forrester07cokriging,%
swersky2014freeze,li2017hyperband};
a plurality of them use BO techniques.
However these treatments neither formalise nor analyse any notion of
\emph{regret} in the multi-fidelity setting.
In contrast, \mfgpucbs is an intuitive UCB idea with good theoretical properties.
\citet{bogunovic2016truncated} study a version of BO where an algorithm might use
cheap, noisy, yet unbiased approximations to a function $\func$;
but as we will explain in Section~\ref{sec:prelims}, this is different to the
multi-fidelity problem.
\citet{agarwal2011oracle} derive oracle inequalities for hyper-parameter tuning with
ERM under computational budgets.
Our setting is more general as it applies to any bandit optimisation task.
\citet{sabharwal2015selecting} present a UCB based idea for tuning hyper-parameters with
incremental data allocation. However, their theoretical results are for an idealised
non-realisable algorithm.
\citet{cutler14mfsim} study reinforcement learning with multi-fidelity simulators by
treating each fidelity as a Markov Decision Process.
Finally,~\citet{zhang15weakAndStrong} study active learning when there is access to a
cheap weak labeler and an expensive strong labeler.
These works study  problems different to optimisation.

% Our work builds on \gpucbs, but 
% The multi-fidelity framework poses substantially new theoretical and algorithmic challenges.
Recently, in~\citet{kandasamy2016mfbandit} we studied the classical $K$-armed bandit
in multi-fidelity settings.
Here, we build on this work to study multi-fidelity Bayesian optimisation;
as such, we share similarities in the assumptions, algorithm, and some analysis techniques.
A preliminary version of this paper appeared in~\citet{kandasamy16mfbo} where we
provided theoretical results in continuous domains and
with two fidelities (one approximation).
In this paper, we expand on the above and provide results both in discrete domains and for a general number of fidelities.
Furthermore, we eliminate some technical assumptions from our previous work and present
cleaner and more interpretable versions of our theorems.
In follow up work~\citep{kandasamy2017boca}, we extend multi-fidelity
optimisation to settings with continuous approximations.
While the assumptions there are considerably different, it builds on the main intuitions
from this work.
To the best of our knowledge, this is the first line of work to formalise a notion of regret
and provide a theoretical analysis for multi-fidelity optimisation.

The remainder of this manuscript is organised as follows.
Section~\ref{sec:prelims} presents our formalism including a notion of simple regret
for multi-fidelity GP optimisation.
Section~\ref{sec:mfbo} presents our algorithm.
We present our theoretical results in Section~\ref{sec:analysis} beginning with
an informal discussion of results for $M=2$ fidelities in
Section~\ref{sec:analysisPreview} to elucidate the main ideas.
The proofs are given in Section~\ref{sec:proofs}.
Section~\ref{sec:experiments} presents our experiments with some details deferred to
Appendix~\ref{app:experiments}.
Appendix~\ref{app:ancillary} collects some ancillary material including a table
of notations and abbreviations in Appendix~\ref{app:notation}.

\section{Problem Set Up}
\label{sec:prelims}
% \vspace{-0.05in}

We wish to maximise a function $\func:\Xcal\rightarrow\RR$
% where $\Xcal$ is either a finite discrete or compact subset of $[0,r]^d$, where $r>0$
where $\Xcal$ is a finite discrete or compact subset of $[0,r]^d$,
where $r>0$ and $d$ is the dimension of $\Xcal$.
We can interact with $\func$ only by querying it at some $x\in\Xcal$ and obtaining a
noisy evaluation $y = f(x) + \epsilon$ of $\func$, where the noise satisfies
$\EE[\epsilon] = 0$.
Let $\xopt \in \argmax_{x\in\Xcal}\func(x)$ be a maximiser of $\func$
 and $\funcopt = \func(\xopt)$ be the maximum value.
Let $\xt\in\Xcal$ be the point queried at time $t$ by a sequential procedure.
The  goal in bandit optimisation is to achieve small \emph{simple regret}
% $S_n = \min_{t=1,\dots,n} \funcopt - \func(\xt)$ after $n$ queries.
$S_n$, defined below, after $n$ queries to $\func$.
\vspace{-0.05in}
\begin{align*}
S_n = \min_{t=1,\dots,n} \;\funcopt - \func(\xt).
\label{eqn:Sn}
\numberthis
\end{align*}
Our primary distinction from the usual setting is that
we have access to $M-1$ successively accurate
approximations $\funcone,\functwo,\dots,\funcmm{M-1}$ to the function of interest
$\func = \funcM$.
We refer to these approximations as fidelities.
The multi-fidelity framework is attractive when the following two conditions are true
about the problem.
\vspace{-0.05in}
% \begin{itemize}[label=-\hspace{-0.02in},leftmargin=0.13in]
\begin{enumerate}[leftmargin=0.19in]
\item
\emph{The approximations $\funcone,\dots,\funcmm{M-1}$ approximate $\funcM$.}
To this end, we will assume a uniform bound for the fidelities,
$\|\funcM - \funcm\|_\infty \leq \zetam$ for $m=1,\dots, M$, where the bounds
$\zetaone>\zetatwo>\dots>\zetaM=0$ are known.
\vspace{-0.05in}
\item
\emph{The approximations are cheaper than evaluating at $\funcM$.}
We will assume that a query at fidelity $m$ expends a cost $\costm$ of a resource,
such as computational effort or money.
The costs are known and satisfy  $0<\costone<\costtwo<\dots<\costM$.
\end{enumerate}
\vspace{-0.05in}
Therefore, as the fidelity $m$ increases, the approximations become better
but are also more costly.
% 
% We encode the fact that an approximation at fidelity $m$ approximates
% the expensive function at fidelity $M$
% via the assumption, $\|\funcM - \funcm\|_\infty \leq \zetam$, where
% $\zetaone>\zetatwo>\dots>\zetaM=0$.
% Each query at fidelity $m$ expends a cost $\costm$ of a resource, e.g.
% computational effort or money, where $\costone < \costtwo < \dots <
% \costM$. 
% As the fidelity $m$ increases, the approximations become better but are also more costly.
An algorithm for multi-fidelity bandits is a sequence of query-fidelity
pairs $\{(\xt,\mt)\}_{t\geq 0}$, where at time $n$, the algorithm chooses $({x}_n,{m}_n)$
using information from previous query-observation-fidelity triples
$\{(\xt,\yt,\mt)\}_{t=1}^{n-1}$.
Here $\yt = \funcmm{\mt}(\xt) + \epsilon_t$ where, the
$\epsilon_t$ values are independent at each time step $t$ and $\EE[\epsilon_t] = 0$.
% Typically, after $n$ steps of such a procedure, we will have queried any of the $M$
% fidelities multiple times.

Some smoothness assumptions on $\funcm$'s are needed to make the problem
tractable. A standard in the Bayesian nonparametric literature is to use a Gaussian 
process (GP) prior~\citep{rasmussen06gps}  with covariance kernel
$\kernel$. 
% In this work we focus on the squared exponential (SE) $\kernel_{\sigma, h}$ and
% the \matern $\kernel_{\nu, h}$ kernels as they are popularly used in practice
% and their theoretical properties are well studied.
Two popular kernels of choice are the
squared exponential (SE) kernel $\kernel_{\sigma, h}$ and the \matern 
kernel $\kernel_{\nu, h}$.
Writing $z = \|x-x'\|_2$, they are defined as
% $\kernel_{\sigma, h}(x, x') = \sigma \exp\left( -z^2/(2h^2) \right)$,
% $\kernel_{\nu, h}(x, x') = \frac{2^{1-\nu}}{\Gamma(\nu)}
%     \left( \frac{\sqrt{2\nu}z}{h} \right)^\nu
%     B_\nu\left( \frac{\sqrt{2\nu}z}{h} \right)$,
\[
% \kernel_{\sigma, h}(x, x') = \sigma \exp\left( -z^2/(2h^2) \right),\hspace{0.2in}
\kernel_{\sigma, h}(x, x') = \sigma \exp\left(-\frac{z^2}{2h^2} \right),\hspace{0.2in}
\kernel_{\nu, \rho}(x, x') = \frac{2^{1-\nu}}{\Gamma(\nu)}
    \left( \frac{\sqrt{2\nu}z}{\rho} \right)^\nu
    B_\nu\left( \frac{\sqrt{2\nu}z}{\rho} \right),
\]
respectively.
Here $\sigma,h,\nu,\rho>0$ are parameters of the kernels and 
$\Gamma,B_\nu$ are the Gamma and modified Bessel functions.
A convenience the GP framework offers is that posterior distributions are
analytically tractable. 
% If $\func\sim\GP(0,\kernel)$ is a sample from a GP,
% and we have observations $\Dcaln=\{(x_i,y_i)\}_{i=1}^n$,
%  where $y_i = f(x_i) + \epsilon$ and $\epsilon\sim\Ncal(0,\eta^2)$ is Gaussian noise,
% then the posterior distribution for $\func(x)|\Dcaln$ is also Gaussian 
% $\Ncal(\mu_n(x),\sigma^2_n(x))$ with~\citep{rasmussen06gps},
If $\func\sim\GP(0,\kernel)$ is a sample from a GP,
and we have observations $\Dcalt=\{(x_i,y_i)\}_{i=1}^t$,
 where $y_i = f(x_i) + \epsilon$ and $\epsilon\sim\Ncal(0,\eta^2)$ is Gaussian noise,
then the posterior distribution for $\func(x)|\Dcalt$ is also Gaussian 
$\Ncal(\mu_t(x),\sigma^2_t(x))$ with~\citep{rasmussen06gps},
\begin{align*}
\hspace{-0.05in}
% \mu_n(x) = \kb^\top\Delta^{-1}Y, \hspace{0.35in}
% \sigma^2_n(x) = \kernel(x,x) - \kb^\top\Delta^{-1}\kb.
\mu_t(x) = k^\top(K + \eta^2 I_t)^{-1}Y, \hspace{0.35in}
\sigma^2_t(x) = \kernel(x,x) - k^\top(K + \eta^2 I_t)^{-1}k.
\hspace{0.05in}
\numberthis \label{eqn:gpPostEqns}
\end{align*}
Here, $Y\in\RR^t$ is a vector with $Y_i=y_i$, $k\in\RR^t$ is a vector with $k_i =
\kernel(x,x_i)$. The matrix $K \in \RR^{t\times t}$ is given by
$K_{i,j} = \kernel(x_i,x_j)$.
$I_t\in\RR^{t\times t}$ is the $t\times t$ identity matrix.
%
% \textbf{Generative Model:}
% \insertprespacing

\subsection{The Generative Process for Multi-fidelity Optimisation}
\label{sec:mfogenprocess}

In keeping with the above framework,
we assume the following generative model for the functions $\funcone,\dots,\funcM$.
% we make the following assumptions on our problem.
% 
A generative mechanism is given constants $\zetaone,\dots,\zetamm{M-1}$.
It then generates the functions as follows.
\vspace{-0.05in}
\begin{itemize}[leftmargin=0.7in]
\item[\textbf{Step 1.\,}] Sample $\funcm\sim\GP(0,\kernel)$ for $m=1,\dots,M$.
\hfill  \textbf{(\Gone)}
\vspace{-0.05in}
\item[\textbf{Step 2.\,}] Check if
$\;\|\funcM-\funcm\|_\infty \leq \zetam$ for all $m=1,\dots,M-1$.
If true, then deliver $\funcone,\dots,\funcM$.
If false, go back to Step 1.
\hfill  \textbf{(\Gtwo)}
\end{itemize}
In addition to this, we will also assume that
upon querying $\funcm$ at $\xt$ we observe $\funcm(\xt) + \epsilon$
where $\epsilon\sim\Ncal(0,\eta^2)$ is Gaussian noise with variance $\eta^2$.

% 
% \vspace{0.05in}
% % \newcommand{\insertAssVSpace}{\vspace{-0.05in}}
% \newcommand{\insertAssVSpace}{\vspace{-0.00in}}
% % \vspace{-0.14in}
% \begin{itemize}%[leftmargin=0.3in]
% \item[\Gone.] \textbf{(GP Assumption) }\hspace{0.1in}%
% $\funcm\sim\GP(\zero,\kernel)$ for all $m=1,\dots,M$.
% Upon querying $\funcm$ at $\xt$ we observe $\funcm(\xt) + \epsilon$
% % \hphantom{T}\hspace{0.23in}
% where $\epsilon\sim\Ncal(0,\eta^2)$ is Gaussian noise with variance $\eta^2$.
% \insertAssVSpace
% \item[\Gtwo.] \textbf{(Multi-fidelity Assumption) }\hspace{0.1in}%
% For the functions $\funcone,\dots,\funcM$ generated above,
% there exists \emph{known} constants $\zetaone,\zetatwo,\dots\zetaM\geq0$ such that,
% $\zetaone>\zetatwo>\dots>\zetaM=0$ and
% $\;\|\funcM-\funcm\|_\infty \leq \zetam$ for all $m=1,\dots,M$.
% \end{itemize}
% % \vspace{0.05in}
% 

Condition~\Gtwos characterises the approximation conditions for the lower fidelities.
% As we will see in our analysis, smaller values for the $\zetam$ parameters, i.e. better
% approximations, will lead to better bounds for a multi-fidelity algorithm over
% single fidelity algorithms.
% However, since~\Gones is of a Bayesian flavour, it is natural to ask if%
% ~\Gtwos is consistent with~\Gone?
Lemma~\ref{lem:gtwoprob} shows that~\Gtwos is satisfied with positive
probability when  $\funcone,\dots,\funcM$ are sampled from a GP.
Hence this is a valid  generative process 
since~\Gtwos will eventually be satisfied.
% It is natural to ask if condition~\Gtwos is consistent with~\Gone, that is, does it
% occur with non-zero probability when the functions $\funcone,\dots,\funcM$ are sampled
% from a GP so that the generative process is valid.
% In Lemma, Appendix~\ref{app:ancillaryResults} we argue that
% this is in fact the case;
% the generative mechanism can keep
% sampling the functions and deliver them when the conditions hold true.
Moreover,
in Section~\ref{sec:mfbo} we argue that while~\Gtwos renders the 
computation of the true posterior of all GPs
inefficient via closed form equations such as in~\eqref{eqn:gpPostEqns},
it is still possible to derive an efficient algorithm that uses~\eqref{eqn:gpPostEqns} to
determine future points for evaluation.

We note that other natural approximation conditions can be used to characterise
the cheaper fidelities.
We choose a uniform bound condition because it provides a simple way to reason about one
fidelity from the others, hence keeping the analysis tractable while ensuring the model
is interesting enough so as to yield reasonable results in practice.
% A disadvantage of this form however, is that it does not allow for an efficient
% way to share information from one fidelity to another.
That said,
we believe that the intuitions in this work can be used to develop other
upper confidence based multi-fidelity BO algorithms for other approximation conditions.
In fact, the approximation conditions in our follow up work in~\citet{kandasamy2017boca},
are of a Bayesian flavour via a kernel on the fidelities.
The algorithm, \boca, builds on the key insights developed here.
% \toworkon{better way to phrase this para and the one before?}

% \modifjan{
It is worth mentioning that while our theoretical results are valid for arbitrary
$M$ and $\costm$ values, it is instructive to think of $M$ as being a small fixed value
and of $\costone$ as being comparable to $\costM$.
For instance, in many practical applications of multi-fidelity optimisation, while an
approximation may be cheaper than the real experiment, it could itself be quite expensive
and hence require an intelligence procedure, such as Bayesian optimisation,
to choose the next point. This is the regime the current paper focuses on, as opposed to asymptotic regimes where $M\rightarrow\infty$ and/or $\costone\rightarrow 0$.
Moreover, very large values of $M$ are better handled by the formalism
in our follow up work in~\citet{kandasamy2017boca}.
%Concretely, our goal here is to develop a framework that can be applied to multiple
%approximations as opposed to an interest in
% }

% \hphantom{T}\hspace{0.23in}
Finally, we note that
Assumption~\Gones can be relaxed to hold for different kernels and noise variances
for each fidelity, i.e. different $\kernelm, \eta^{(m)}$ for $m=1,\dots,M$, with minimal
modifications to our analysis but we use the above form to simplify the presentation
of the results. In fact, our practical implementation uses different kernels.

\vspace{0.1in}

\subsection{Simple Regret for Multi-fidelity Optimisation}

% \vspace{0.1in}
% \noindent
% \textbf{Simple Regret for Multi-fidelity Optimisation:}
Our goal is to achieve small simple regret
$S(\COST)$ after spending capital $\COST$ of a resource.
We will aim to provide \emph{any-capital} bounds, meaning that we will assume that
the game is played indefinitely and will try to bound the regret for all
(sufficiently large) values of $\COST$.
This is similar in spirit to any-time analyses in single fidelity bandit
methods as opposed to fixed time horizon analyses.
% Suppose we have made $N$ queries at any fidelity within capital $\COST$,
% i.e. $N$ is the \emph{random} quantity such that
% $N = \max\{n\geq 1: \sum_{t=1}^n \costmt \leq \COST\}$ during an execution
% of the method.
Let $\{\mt\}_{t\geq 0}$ be the fidelities queried by a multi-fidelity method at
each time step.
Let $N$ be the \emph{random} quantity such that
$N = \max\{n\geq 1: \sum_{t=1}^n \costmt \leq \COST\}$, i.e. it is the number of queries
the strategy makes across all fidelities until capital $\COST$.
Only the optimum of $\func=\funcM$ is of interest to us. The lower fidelities
are useful to the extent that they help us optimise $\funcM$ with less cost,
but there is no reward for optimising a cheaper approximation.
Accordingly, we set the instantaneous reward $\qt$ at time $t$ to be
$-\infty$ if $\mt\neq M$ and $\funcM(\xt)$ if $\mt=M$.
If we let $\rt=\funcopt-\qt$ denote the instantaneous regret, we have
$\rt = + \infty$ if $\mt\neq M$ and $\funcopt - \funcM(\xt)$ if $\mt=M$.
For optimisation, the simple regret is simply the best instantaneous regret,
$S(\COST) = \min_{t=1,\dots,N} \rt$.
Equivalently,
\begin{align*}
S(\COST) \,=\,
\min_{t=1,\dots,N} \rt \;=\;
\begin{cases}
\displaystyle\min_{\substack{t=\,1,\dots,N\\t\,:\,\mt = M}}\;
  \funcopt - \funcM(\xt) \;\,% \hspace{0.01in}
%     &\substack{\text{if we have queried at the $M$\ssth fidelity}\\
%                      \text{at least once}} \\
     &  \parbox[t]{.37\columnwidth}{if we have queried at the $M$\ssth fidelity at least
                                   once,}
  \\[0.30in]
+\,\infty & \text{otherwise.}
\end{cases}
\numberthis
\label{eqn:defnMFSR}
\end{align*}
Note that the above reduces to $S_n$ in~\eqref{eqn:Sn}
when we only have access to $\funcM$ with $n=N=\floor{\COST/\costM}$.

Before we proceed, we note that it is customary in the bandit literature to analyse
\emph{cumulative regret}.
The definition of cumulative regret depends on the
application at hand~\citep{kandasamy2016mfbandit} and our results can be
extended to to many sensible notions of cumulative regret.
However, both to simplify exposition and since our focus in this paper is optimisation,
we stick to simple regret.

\insertChallengesFigure

\vspace{0.1in}

\noindent
\textbf{Challenges:}
We conclude this subsection with a commentary on some of the challenges in multi-fidelity
optimisation using Figure~\ref{fig:challenges} for illustration.
For simplicity, we will focus on $2$ fidelities when
we have one approximation $\funcone$ to an expensive function $\functwo$. For now
assume that (unrealistically) $\funcone$ and its optimum $\xoneopt$ are known.
Typically $\xoneopt$ is suboptimal for $\functwo$.
A seemingly straightforward solution might be to search for $\xopt$
in an appropriate subset, such as a neighborhood of $\xoneopt$.
However, if this neighborhood is too small, we might miss the
optimum~$\xopt$ (green region in Figure~\ref{fig:f1f2regSmall}).
A crucial challenge for multi-fidelity methods is to not get stuck at the optimum of
a lower fidelity.
While exploiting information from lower fidelities,
it is also important to \emph{explore} sufficiently at higher fidelities. 
In our experiments, we demonstrate that naive strategies which do not do so could get
stuck at the optimum of a lower fidelity.
Alternatively, if we pick a very large subset (Figure~\ref{fig:f1f2regLarge})
we might not miss $\xopt$;
however, it defeats the objectives of the multi-fidelity set up
where the goal is to use the approximation to be prudent about where we query $\functwo$.
 Figure~\ref{fig:f1f2regCorrect} displays a seemingly sensible subset,
but it remains to be seen how it is chosen.
Further, this subset might not even be a neighborhood as illustrated in
Figure~\ref{fig:f1f2regDouble}, where $\funcone,\functwo$ are multi-modal and
the optima are in different modes.
In such cases, an appropriate algorithm should explore all such modes.
On top of the above, an algorithm does not actually know $\funcone$.
A sensible algorithm should explore $\funcone$ and simultaneously identify the above
subset, either implicitly or explicitly, for exploration at the second fidelity $\functwo$.
Finally, it is also important to note that $\funcone$ is not simply a noisy version of
$\functwo$;
this setting is more challenging as an algorithm needs to explicitly
account for the bias in the approximations.

\subsection{Some Useful Properties of GPs}

For what follows, we present some useful properties and concepts related to GPs
with well behaved kernels.
We we will denote probabilities when $\funcone,\dots,\funcM\sim\GP(0,\kernel)$
independently, by $\PPgp$.
$\PP$ will denote probabilities under the prior in the multi-fidelity
setting which includes \Gtwos after
sampling the functions; i.e. for any event $E$, $\PP(E) = \PPgp(E|\Gtwo)$.
First, we will need the following regularity conditions on the kernel.
It is satisfied for four times differentiable
kernels such as the SE kernel and \matern kernel when
$\nu>2$; see \citet{ghosal06gpconsistency}, Theorem 5.

% \insertprespacing
\begin{assumption}(Theorem 5 in \cite{ghosal06gpconsistency})
Let $\func\sim\GP(\zero,\kernel)$, where $\kernel:[0,r]^d\times
[0,r]^d \rightarrow \RR$ is a stationary kernel~\citep{rasmussen06gps}.
%  $\kernel(\cdot, x)$ is $L$-Lipschitz for all $x$. 
The partial derivatives
of $\func$  satisfies the following condition.
There exist constants $a, b >0$ such that,
\[
\text{for all $J>0$, $\;$and for all $i \in \{1,\dots,d\}$},\quad \PPgp\left( \sup_{x} 
\Big|\partialfrac{x_i}{\func(x)}\Big| > J \right)
\leq a e^{-(J/b)^2}.
\]
\label{asm:kernelAssumption}
\end{assumption}

\insertpostspacing
Observe that we have used notation $\PPgp$ to indicate the prior probability
when $\func\sim\GP(0,\kernel)$ for consistency.
Next, the following assumption supposes that
there is a positive probability to the event that the supremum
of a GP in a bounded domain is smaller than any given $\epsilon>0$.

\begin{assumption}
%(Theorem 4 in \cite{ghosal06gpconsistency})
Let $\Xcal=[0,r]^d$ and
$\func\sim\GP(\zero,\kernel)$.
Let $\kernel:\Xcal\times
\Xcal^d \rightarrow \RR$ be such that
for all $\epsilon > 0$, there exists $\gpsupbound(\epsilon)> 0$ such that,
%  $\kernel(\cdot, x)$ is $L$-Lipschitz for all $x$. 
\[
\PPgp\Big(\sup_{x\in\Xcal}|f(x)| < \epsilon\Big) > \gpsupbound(\epsilon).
\]
\label{asm:supgp}
\end{assumption}
As shown by
Theorem 4 in~\citet{ghosal06gpconsistency}, this is satisfied for the
SE and \matern kernels.
% 
% \vspace{0.1in}
% 
Finally, following~\citep{srinivas10gpbandits},
our theoretical results will be given in terms of the 
\emph{Maximum Information Gain} (\mig), defined below.
% Defined below, it characterises the
% statistical difficulty of GP bandits.
% 
\begin{definition}(Maximum Information Gain~\citep{srinivas10gpbandits})
Let $f\sim \GP(\zero, \kernel)$. Consider any $A\subset\RR^d$ and
let $\Atilde = \{x_1, \dots, x_n\} \subset A$ be a finite subset.
Let $f_\Atilde, \epsilon_\Atilde\in\RR^n$ be such that $(f_\Atilde)_i=f(x_i)$,
$(\epsilon_\Atilde)_i\sim\Ncal(0,\eta^2)$ for $i=1,\dots,n$, and
$y_\Atilde = f_\Atilde+\epsilon_\Atilde$.
Let $I$ denote the Shannon mutual information.
The Maximum Information Gain $\IG_n(A)$ of set $A$ after $n$ evaluations is the maximum
mutual information between the function values and observations
among all choices of $n$ points in $A$.
Precisely,
\[
\IG_n(A) = \max_{\Atilde\subset A, |\Atilde| = n} I(y_\Atilde; f_\Atilde).
\]
\label{def:infGain}
\end{definition}
\insertpostspacing
% \insertpostspacing
The MIG, which depends on the kernel and the set $A$,
will be an important quantity in our analysis as it characterises the statistical
difficulty of GP Bandits.
For a given kernel it typically scales with the volume of
$A$~\citep{srinivas10gpbandits}%
\footnote{%
In section C.2 of~\citet{srinivas10gpbandits}, the kernel's eigenspectrum
is defined with respect to
the uniform measure on the domain $\Xcal$.
When we consider any subset $A\subset\Xcal$ with the same measure and eigenspectrum,
a multiplicative $\vol(A)$ term appears.
}.
For example, if $A=[0,r]^d$ then $\IGn(A) \in
\bigO(r^d\IGn([0,1]^d))$.
It is known that
for the SE kernel, $\IGn([0,1]^d) \in \bigO( (\logn)^{d+1})$ and
for the \matern kernel, $\IGn([0,1]^d) \in \bigO( n^{\frac{d(d+1)}{2\nu +
d(d+1)}}\logn)$~\citep{srinivas10gpbandits,seeger08information}.
% Next, we will need the following regularity conditions on the kernel.
% It is satisfied for four times differentiable
% kernels such as the SE kernel and \matern kernel when
% $\nu>2$; see \citet{ghosal06gpconsistency}, Theorem 5.

\section{A Review of \gpucb}
\label{sec:gpucbreview}

\insertGPUCBHorFigure

% We begin with a brief review of \gpucb and its theoretical properties.
Sequential optimisation methods adopting UCB principles maintain a high probability
upper bound $\ucbt:\Xcal\rightarrow\RR$ for $\func(x)$ for all
$x\in\Xcal$~\citep{auer03ucb}.
At time $t$ we query at the maximiser of this upper bound
$\xt = \argmax_{x\in\Xcal}\ucbt(x)$.
Our work builds on \gpucb~\citep{srinivas10gpbandits}, where
$\ucbt$ takes the form 
$\ucbt(x) = \mu_{t-1}(x) + \betath \sigma_{t-1}(x)$.
Here
$\mu_{t-1}, \sigma_{t-1}$ are the posterior mean and standard deviation of the
GP conditioned on the previous $t-1$ queries $\{(x_i,y_i)\}_{i=1}^{t-1}$ and
$\betat>0$.
The key intuition here is that the mean $\mu_{t-1}$ encourages an
exploitative strategy -- in that we
want to query where we know the function is high -- and
the standard deviation  $\sigma_{t-1}$ encourages an explorative strategy
-- in that we want to query at regions we are uncertain about $\func$ lest we
miss out on high valued regions.
$\betat$ will control the trade-off between exploration and exploitation.
We have presented \gpucbs in Algorithm~\ref{alg:gpucb} and illustrated it in
Figure~\ref{fig:gpucb}.
\insertAlgorithmGPBALGO

The following theorem from~\citet{srinivas10gpbandits} bounds the simple regret
$S_n$~\eqref{eqn:Sn} for \gpucb.%
~\citet{srinivas10gpbandits} give their bounds in terms for the cumulative regret, but
converting it to simple regret is straightforward.
% The algorithm is given in Algorithm~\ref{alg:gpucb}.
% \insertprespacing

\begin{theorem}(Theorems 1 and 2 in \cite{srinivas10gpbandits})
\label{thm:gpucb}
Let $\func\sim\GP(\zero,\kernel)$, $\func:\Xcal\rightarrow\RR$ and the
kernel $\kernel$ satisfies
Assumption~\ref{asm:kernelAssumption}). At each query, we have noisy
observations
$y = f(x) + \epsilon$ where $\epsilon\sim\Ncal(0,\eta^2)$.
Denote $C_1 = 8/\log(1+\eta^{-2})$.
Pick a failure probability $\delta\in(0,1)$. 
The following bounds on the simple regret $S_n$ hold with
$\PPgp$-probability $>1-\delta$
for all $n\geq 1$.
% \hspace{-0.2in}
% \vspace{-0.1in}
\begin{itemize}
\item If $\Xcal$ is a finite discrete set, run \emph{\gpucbs} with
% $\betat = 2\log\left(\frac{|\Xcal|t^2\pi^2}{6\delta}\right)$.
$\betat = 2\log\left(|\Xcal|t^2\pi^2/6\delta\right)$.
Then,
% \vspace{-0.10in}
\[
% \PP\left( \forall n\geq 1,\; R_n \leq \sqrt{C_1n\betan\IGn(\Xcal)} \right)
% \geq 1-\delta
\text{for all $n\geq 1$,}
\hspace{0.4in}
S_n \leq \sqrt{\frac{C_1\betan\IGn(\Xcal)}{n}}
\]
% \vspace{-0.35in}
\item If $\Xcal=[0,r]^d$, run \emph{\gpucbs} with
  $\betat = 2\log\left(\frac{2\pi^2t^2}{3\delta}\right) + 
  2d\log\left(t^2bdr\sqrt{\frac{4ad}{\delta}}\right)$.
Then,
% \vspace{-0.10in}
\[
% \PP\left( \forall n\geq 1,\; R_n \leq \sqrt{C_1n\betan\IGn(\Xcal)} \,+\, 2 \right)
% \geq 1-\delta
\text{for all $n\geq 1$,}
\hspace{0.4in}
S_n \leq \sqrt{\frac{C_1\betan\IGn(\Xcal)}{n}} \,+\, \frac{2}{n}
\]
% \vspace{-0.35in}
\end{itemize}
\end{theorem}

% Before we proceed, we make another reasonable
% assumption about GPs with well behaved kernels.
% Assumption~\ref{asm:supgp} supposes that
% there is a positive probability to the event that the supremum
% of a GP in a bounded domain is smaller than any given $\epsilon>0$.
% 
% \begin{assumption}
% %(Theorem 4 in \cite{ghosal06gpconsistency})
% Let $\Xcal=[0,r]^d$ and
% $\func\sim\GP(\zero,\kernel)$.
% Let $\kernel:\Xcal\times
% \Xcal^d \rightarrow \RR$ be such that 
% for all $\epsilon > 0$, there exists $\gpsupbound(\epsilon)> 0$ such that,
% %  $\kernel(\cdot, x)$ is $L$-Lipschitz for all $x$. 
% \[
% \PPgp\Big(\sup_{x\in\Xcal}|f(x)| < \epsilon\Big) > \gpsupbound(\epsilon).
% \]
% \label{asm:supgp}
% \end{assumption}
% 
% 
% As shown by
% Theorem 4 in~\citet{ghosal06gpconsistency}, this is readily satisfied for the
% SE and \matern kernels.

% 
% \vspace{0.05in}

\section{Multi-fidelity Gaussian Process Upper Confidence Bound $\;$(\mfgpucb)}
\label{sec:mfbo}

We now propose \mfgpucb, which extends \gpucbs to the multi-fidelity setting.
Like \gpucb, \mfgpucbs will also maintain a UCB for $\funcM$ obtained via
the previous queries at \emph{all} fidelities.
Denote the posterior GP mean and standard deviation of $\funcm$ conditioned 
\emph{only} on the previous queries at fidelity $m$ by $\mumt,\sigmamt$
respectively (See~\eqref{eqn:gpPostEqns}).
% $\Dcalmt$ to
% be $\mumt(x) = \EE[\funcm(x)|\Dcalmt]$, $\sigmamt(x) = (\VV[\funcm(x)|\Dcalmt])^{1/2}$.
Then define,
\begin{align*}
\hspace{-0.1in}
\ucbmtx = \;\mumtmo(x) + \betath\sigmamtmo(x) + \zetam,\;\;\;\forall\,m,\;
\hspace{0.5in}
\ucbtx = \;\min_{m=1,\dots,M} \,\ucbmtx.
\numberthis \label{eqn:mfgpucbDefn}
% \\[-0.3in]
\end{align*}
For appropriately chosen $\betat$, $\mumtmo(x) + \betath\sigmamtmo(x)$ will upper
bound $\funcm(x)$ with high probability. By~\Gtwo{} and~\eqref{eqn:mfgpucbDefn},
$\ucbmtx$ upper bounds $\funcM(x)$ for all $m$.
We have $M$ such bounds, and their minimum $\ucbtx$ gives the 
best upper bound for $\funcM$.
Following UCB strategies such as \gpucb,
our next query is at the maximiser of this UCB, $\xt =
\argmax_{x\in\Xcal}\ucbtx$.

Next we need to decide which fidelity to query at. 
Consider any $m<M$.
The $\zetam$ constraints on $\funcm$ restrict the value of $\funcM$ --
the confidence band $\betath\sigmamtmo$  for $\funcm$ is lengthened by 
$\zetam$ to obtain confidence on $\funcM$.
If $\betath\sigmamtmo(\xt)$ for $\funcm$ is large,
% relative to $\zetam$,
it means that we have not constrained $\funcm$ sufficiently well at
$\xt$ and should query at the $m$\ssth fidelity.
On the other hand, 
querying indefinitely in the same region to reduce the uncertainty  $\betath\sigmamtmo$
at the $m$\ssth fidelity in that region will not help us much as the $\zetam$
elongation caps off how much we can learn about $\funcM$ from $\funcm$; i.e. even if
we knew $\funcm$ perfectly, we will only have constrained $\funcM$ to within a
$\pm\zetam$ band.
Our algorithm captures this simple intuition. Having selected $\xt$, we begin 
by checking at the first fidelity. If $\betath\sigmammtt{1}{t-1}(\xt)$ is smaller
than a threshold $\gammamm{1}$, we proceed to the second fidelity. 
If at any stage $\betath\sigmammtt{m}{t-1}(\xt)\geq\gammam$ we query at fidelity $\mt=m$.
If we proceed all the way to fidelity $M$, we query at $\mt=M$.
We will discuss choices for $\gammam$ in Sections~\ref{sec:analysisPreview}
and~\ref{sec:implementation}.
We  summarise the resulting procedure in Algorithm~\ref{alg:mfgpucb}.
\insertAlgoMFGPUCB
% \vspace{-0.05in}
% Simulation Figure ================================================================
\insertSimulationLong
% Simulation Figure ================================================================

Before we proceed, we make an essential observation. 
The posterior for any 
$\funcm(x)$ conditioned on previous queries at \emph{all} fidelities 
$\bigcup_{\ell=1}^M\Dcalmmtt{\ell}{t}$ 
is not Gaussian due to the $\zetam$ constraints (\Gtwo). 
However, $|\funcm(x)-\mumtmo(x)|<\betath\sigmamtmo(x)$ holds with high probability,
since, by conditioning only on queries at the $m$\ssth fidelity
we have Gaussianity for $\funcm(x)$.
(See Lemma~\ref{lem:discreteUCB}, Section~\ref{sec:proofsDiscrete}).

% \subsubsection*{A Simulation}
\vspace{0.2in}
\noindent
% \textbf{A Simulation:}
\textbf{An Illustration of \mfgpucb:}
Figure~\ref{fig:mfSimLong} illustrates \mfgpucbs via a simulation 
on a $2$--fidelity problem.
At the initial stages, \mfgpucbs is mostly exploring $\Xcal$ in the first
fidelity. $\betath\sigmaonetmo$ is large and we are yet to constrain $\funcone$ well
to proceed to $m=2$.
At $t=10$, we have constrained $\funcone$ sufficiently well at a region around the
optimum. $\betath\sigmaonetmo(\xt)$ falls below $\gammaone$ and we query at $\mt=2$.
Notice that once we do this (at $t=11$), $\ucbtwot$ dips to change $\ucbt$ in that
region.
At $t=14$, \mfgpucbs has identified the maximum $\xopt$ with just $4$ queries to
$\functwo$.
The region shaded in cyan in the last figure is the ``good set'' $\Xcalg$, which we
alluded to in Section~\ref{sec:prelims}.
We will define it formally and explain its significance in the multi-fidelity set up
shortly.
Our analysis predicts that most second fidelity queries in \mfgpucbs will be
be confined to this set (roughly) and the simulation corroborates this claim.
For example, in the last figure, at $t=50$, the algorithm decides to explore at
a point far away from the optimum.
However, this query occurs in the first fidelity since we have not
sufficiently constrained $\funcone(\xt)$ in this region and $\betath\sigmaonetmo(\xt)$
is large.
The key idea is that it is \emph{not necessary} to query such regions at the second
fidelity as the first fidelity alone is enough to conclude that it is suboptimal.
% Herein lies the crux of our method.
In addition, observe that in a large portion of $\Xcal$, $\ucbt$ is given by
$\ucbonet$ except in a small neighborhood around $\xopt$, where it is given
by $\ucbtwot$.
% \toworkon{State the last sentence somewhere in the main text.}

Next we present our main theoretical results. 
We wish to remind the reader
that a table of notations  is available in Appendix~\ref{app:notation}.

% % Analysis ========================================================================
% \input{analysis}
% % =================================================================================
% 
% % Implementation ==================================================================
% \input{implementation}
% % =================================================================================

\section{Theoretical Results}
\label{sec:analysis}

First and foremost, we will show that condition \Gtwos occurs with positive probability
when we sample the functions from a GP.
% For this, we will denote probabilities when $\funcone,\dots,\funcM\sim\GP(0,\kernel)$ 
% by $\PPgp$.
% Let $\PP$ denote probabilities under the prior which includes \Gtwos after
% sampling the functions; i.e. for any event $E$, $\PP(E) = \PPgp(E|\Gtwo)$.
The following lemma shows that $\PPgp(\Gtwo) =\Gtwoprob > 0$
which establishes that the generative
mechanism is valid. The proof is given in Section~\ref{sec:proofs}.

\begin{lemma}
\label{lem:gtwoprob}
Let $\funcone,\dots,\funcM$ be sampled from $\GP(0,\kernel)$ and
{\normalfont \Gtwo}
denote the event $\{\|\funcM-\funcm\|_\infty \leq \zetam, \forall m \leq M-1\}$.
Then,
\begin{equation}
\PPgp({\normalfont \Gtwo})
=
\Gtwoprob
% = \gpsupbound\big(\zetamm{M-1}/2\big)
\;\geq\;
\gpsupbound\bigg(\frac{\zetamm{M-1}}{2}\bigg)
\cdot
\prod_{m=1}^{M-1}
\gpsupbound\bigg(\frac{\zetamm{m}}{2}\bigg)
\label{eqn:gtwoprob}
\end{equation}
Here $\gpsupbound$ is from Assumption~\ref{asm:supgp}.
$\Gtwoprob>0$ since each of the terms in the product are positive.
\end{lemma}

We are now ready to present our theoretical results.
We begin with an informal yet intuitive introuduction
to our theorems in $M=2$ fidelities.

\subsection{A Preview of our Theorems}
\label{sec:analysisPreview}

% We begin an informal yet intuitive introduction to
% our theorems in $M=2$ fidelities.
In this subsection, we will
ignore constants and $\polylog$ terms when they are dominated by other terms.
$\lesssim,\gtrsim,\asymp$ denote inequality and equality ignoring constants.
When $A\subset\Xcal$, we will denote its complement by $\complement{A}$.

Fundamental to the $2$-fidelity problem is the
good set $\Xcalg = \{x\in\Xcal; \funcopt-\funcone(x) \leq \zetaone\}$.
$\Xcalg$ is a high-valued region for $\functwo(x)$:
for all $x\in\Xcalg$, $\functwo(x)$ is at most $2\zetaone$ away from the optimum.
If a multi-fidelity strategy were to use \emph{all} its second fidelity queries only in
$\Xcalg$, then, by Theorem~\ref{thm:gpucb},
the regret will only have $\IGn(\Xcalg)$ dependence after $n$ high
fidelity queries.
In contrast, a strategy that only operates at the highest fidelity, such as \gpucb,
will have $\IGn(\Xcal)$ dependence.
When $\zetaone$ is small, i.e. when $\funcone$ is a good
approximation to $\functwo$, $\Xcalg$ will be much smaller than $\Xcal$.
Then, $\IGn(\Xcalg)\ll \IGn(\Xcal)$, and the
multi-fidelity strategy will have better bounds on the regret than a single fidelity
strategy.
Alas, achieving this somewhat ideal goal is not possible without perfect knowledge of
the approximation.
However,  with \mfgpucbs we can come quite close.
As we will show shortly, \emph{most} second fidelity queries will be
confined to the slightly inflated good set
$\Xcalrg = \{x\in\Xcal; \funcopt - \funcone(x) \leq \zetaone + 3\gammaone\}$.
% Here $\rho$ is a global constant that is different for the continuous and discrete
% cases.
% Here $\rho>0$ is a parameter which will be explained in our theorems.
The following lemma bounds the number of first and second fidelity evaluations in
$\Xcalrg$ and its complement $\complement{\Xcalrg}$.
We denote the number of queries at the $m$\ssth fidelity in a set
$A\subset\Xcal$ within the first $n$ time steps by $\Tmnxx{A}$.

\begin{lemma}[Informal, Bounding the number of evaluations for $M=2$]
Let $\Xcal\subset [0,r]^d$.
Consider \emph{\mfgpucbs} after $n$ total evaluations at either fidelity.
Let $\Tmnxx{A}$ denote the number of fidelity $m$ queries in some set $A\subset\Xcal$
in $n$ steps. Then,
% Then, for all $\alpha\in(0,1)$, there exists $\rho>0$ depending only on $\alpha$
% such that the following statements hold with high probability,
\begingroup
\allowdisplaybreaks
\begin{align*}
&
T^{(1)}_n\Big({\Xcalrgbar}\Big)
%  \lesssim \;{\rm sublinear}(n)\,\cdot\Pi(\Xcalrg),
 \lesssim \;\polylog(n)\,\cdot\Pi(\Xcalrg),
 \hspace{0.4in}
\Tmmttxx{1}{n}{\Xcalrg}
 \lesssim \;\frac{{{\footnotesize\rm polylog}}(n)}{\gammaonesq}\,\cdot
  \Pi(\Xcalrg),
 \\[0.1in]
&T^{(2)}_n\Big({\Xcalrgbar}\Big)
 \lesssim \;\taun\cdot\,\Pi\Big({\Xcalrgbar}\Big),
 \hspace{0.86in}
\Tmmttxx{2}{n}{\Xcalrg} \asymp\; n.
\end{align*}
\endgroup
Here $\Pi(A) = |A|$ for discrete $A$
and $\Pi(A) = \vol(A)$ for continuous $A$.
The bound for $T^{(2)}_n\Big(\Xcalrgbar\Big)$ holds for any sublinear
increasing sequence $\{\taun\}_{n\geq 1}$ 
% The ${\rm sublinear}(n)$ term for $T^{(1)}_n$ is $\polylog(n)$ for discrete $\Xcal$ and
% $\poly(n)$ for continuous compact $\Xcal$.
\label{lem:TmnxBoundsInformal}
\end{lemma}

The above lemma will be useful for two reasons.
First, the bounds on $\Tmmttxx{2}{n}{\cdot}$ show that most second fidelity
queries are inside $\Xcalrg$;
the number of such expensive queries outside $\Xcalrg$ is small.
% precisely $\asymp n^\alpha$ for all $\alpha>0$ for an appropriate $\rho$.
This \emph{strong} result is only possible in the multi-fidelity setting.
From the results of~\citet{srinivas10gpbandits}, 
we can infer that
the best achievable bound on the number of plays for \gpucbs inside a
suboptimal set is $\asymp n^{1/2}$ for the SE kernel and even worse for the \matern
kernel.
For example, in the simulation of Figure~\ref{fig:mfSimLong}, all queries to $\functwo$
are in fact confined to $\Xcalg$ which is a subset of $\Xcalrg$.
This allows us to obtain regret that scales with $\IGn(\Xcalrg)$ as explained above.
Second, we will use Lemma~\ref{lem:TmnxBoundsInformal} to control $N$,
the (random) number of queries by \mfgpucbs  within capital $\COST$.
Let $\nCOST= \floor{\COST/\costtwo}$ be the (non-random) number of queries by a
single fidelity method operating only at the second fidelity.
As $\costone<\costtwo$, $N$ could be large for an arbitrary
multi-fidelity method.
However, using the bounds on $\Tmmttxx{1}{n}{\cdot}$
we can show that $N$ is $\asymp\nCOST$ when $\COST$ is larger than some value
$\COSTnought$. 
Below, we detail the main ingredients in the proof of Lemma~\ref{lem:TmnxBoundsInformal}.
% 
% \noindent
% \textbf{Main Proof Ingredients:} 
% We outline the main intuitions as to why ...
% \\[-0.25in]
\begin{itemize}[leftmargin=0.3in]
\item
$ T^{(1)}_n(\Xcalrg)$:
By the design of our algorithm, \mfgpucbs will begin querying $\funcone$.
To achieve finite regret we need to show that we will eventually query $\functwo$.
% In fact, we show that after sufficient exploration at $\funcone$, \emph{most} queries are
% at $\functwo$.
For any region in $\Xcalrg$ the switching condition of step 2 in Algorithm~\ref{alg:mfgpucb}
ensures that we do not query that region indefinitely.
That is, if we keep querying a certain region, the first fidelity GP uncertainty
$\betath\sigmamtmo$ will reduce below $\gammaone$ in that region.
% This allows us to bound the number of first fidelity queries inside $\Xcalrg$ this way.
% As expected, the bound has critical dependence on the threshold parameter $\gammaone$.
We will discuss the implications of the choice of
$\gammaone$ at the end of this subsection and in
Section~\ref{sec:implementation}.

\vspace{-0.05in}

\item 
$ T^{(1)}_n\big({\Xcalrgbar}\big)$:
For  queries to $\funcone$ outside $\Xcalrg$, we use the following reasoning:
as $\funcone$ is small outside $\Xcalrg$, it is unlikely to contain the UCB maximiser
and be selected in step 1 of Algorithm~\ref{alg:mfgpucb} several times.

\vspace{-0.05in}

\item
$ T^{(2)}_n\big({\Xcalrgbar}\big)$:
We appeal to previous first fidelity queries. If we are querying
at the second fidelity at a certain region, it can only be because the first fidelity
confidence band is small.
This implies that there must be several first fidelity queries in that
region which in turn implies that we can learn about $\funcone$ with high confidence.
As $\funcone$ alone would tell us that any point in $\Xcalrg$ is suboptimal for
$\functwo$, the maximiser of the UCB is unlikely to lie in this region frequently.
Hence, we will not query outside $\Xcalrg$ often.
% The threshold parameter $\gammaone$ does appear in the bound, but its dependence is very
% mild.

\end{itemize}

It follows from the above that the number of second fidelity queries in
$\Xcalrg$ scales $\Tmmttxx{2}{n}{\Xcalrg}\asymp n$.
Finally, we invoke techniques from~\citet{srinivas10gpbandits} 
to control the regret using the MIG.
However, unlike them, we can use the MIG of $\Xcalrg$ since an overwhelming amount
of evaluations at the second fidelity are in $\Xcalrg$.
This allows us to obtain a tighter bound on $R(\COST)$ of the following form.
% Consequently, we have the following bound on the regret of \mfgpucbs for $M=2$ fidelities.

\begin{theorem}[Informal, Regret of \mfgpucbs for $M=2$]
\label{thm:compactThm2}
Let $\Xcal\subset[0,r]^d$.
% and $\funcone, \functwo\sim \GP(\zero,\kernel)$. 
% There exists $\rho$ depending on $\alpha$ and $\COSTnought$ such that
% for all $\COST>\COSTnought$ the following holds with high probability.
% Then, for all $\alpha\in(0,1)$, there exists $\rho>0,\COSTnought$ such that
% for all $\COST>\COSTnought$ the following holds
Then there exists 
$\COSTnought$ depending only on $\gammaone$, $\costone$ and the approximation $\funcone$
such that,
for all $\COST>\COSTnought$ the following holds
% w.h.p.
with high probability.
{ %\small
\begingroup
\allowdisplaybreaks
\begin{align*}
\vspace{-0.05in}
S(\COST)\;\lesssim\;\;\;
   \sqrt{\frac{\beta_{\nCOST}\IGnn{\nCOST}(\Xcalrg)}{\nCOST}}
%   \;\;+\;\;
%   \sqrt{\frac{\beta_{\nCOST}\IGnn{\nCOST^\alpha}(\Xcal)}{\nCOST^{2-\alpha}}}
% \\[-0.25in]
\end{align*}
\endgroup
}
% Here $\rho$ depends only on $\alpha$ and
% $\COSTnought$ depends on $\alpha$, $\gammaone$, $\costone$ and the approximation $\funcone$.
\end{theorem}

It is instructive to compare the above rates against that for \gpucbs
in Theorem~\ref{thm:gpucb}.
By dropping the common and sub-dominant terms,
the rate for \gpucbs is $\IGnch(\Xcal)$ whereas for \mfgpucbs it is
$\IGnch(\Xcalrg)$.
% the rate for \mfgpucbs is 
% $\IGnch(\Xcalrg) + \nCOST^{\alpha-1}\IGnn{\nCOST^\alpha}^{1/2}(\Xcal)$
% whereas for \gpucbs it is $\IGnch(\Xcal)$.
% The latter term for \mfgpucbs is small due to the decaying $\nCOST^{\alpha-1}$ term 
% and since the MIG's dependence scales as $\nCOST^\alpha$.
Therefore, whenever the approximation is very good
($\vol(\Xcalrg)\ll\vol(\Xcal)$) the rates for \mfgpucbs are very appealing.
When the approximation worsens and $\Xcalg,\Xcalrg$ become larger, the bound decays
gracefully.
In the worst case, \mfgpucbs is never worse than \gpucbs up to constant terms for
$\COST \geq \COSTnought$.
% Intuitively, the above result states that \mfgpucbs uses ``most'' of its queries to
% $\functwo$ inside $\Xcalrg$ having eliminated the remaining regions using $\funcone$.
The $\COSTnought$ term is required since at the initial stages, \mfgpucbs will be
exploring $\funcone$ before proceeding to $\functwo$, at which stage
its regret will still be $+\infty$.
The costs $\costone,\costtwo$ get factored into the result via the $\COST > \COSTnought$
condition. If $\costone$ is large, for fixed $\gammaone$, a larger amount of capital is
spent at the first fidelity, so $\COSTnought$ will be large.

% rehash ... 

Now let us analyse the effect of the parameter $\gammaone$ on the result.
At first sight, large $\gammaone$ seems to increase the size of $\Xcalrg$ which would
suggest that we should keep it as small as possible. However, smaller $\gammaone$ also
increases $\COSTnought$; intuitively, if $\gammaone$ is too small, then one will wait for
a long time in step 2 of Algorithm~\ref{alg:mfgpucb} for $\betath\sigmammtt{1}{t-1}$ to
decrease without proceeding to $\functwo$.
% As we incur infinite regret for lower fidelity
% queries, it causes $\COSTnought$ to be large.
% break new para here?
As one might expect, an ``optimal'' choice of $\gammaone$ depends on how large a
$\COSTnought$ we are willing to tolerate; i.e. how long we are
willing to wait investigating the cheap approximation.
Moreover, 
if the approximation is extremely cheap, it makes sense to use very small  $\gammaone$
and learn as much as possible about $\functwo$ from $\funcone$.
However, it also depends on other problem dependent quantities such as $\Xcalg$.
In Section~\ref{sec:analysisDisc} we
describe a choice for $\gammaone$ based on $\costone,\costtwo$ and
$\zetaone$ that aims to balance the cost spent at each fidelity.
% In~\citet{kandasamy16mfbo} we were able to exactly specify a practical choice for
% $\gammaone$ in the $K$-armed setting, which, unfortunately is not as straightforward in the
% GP setting.
% However, this trade-off is also determined by unknown problem dependent quantities.
In our experiments however,
we found that more aggressive choices for these threshold values $\gammam$ perform better
in practice.
We describe one such technique
Section~\ref{sec:implementation}.

\vspace{0.1in}

For general $M$, we will define a hierarchy of good sets, the complement
of which will be eliminated when we proceed from one fidelity to the next.
At the highest fidelity, we will be querying mostly inside a small subset of $\Xcal$
informed by the approximations $\funcone,\dots,\funcmm{M-1}$.
% The proofs follow a similar strategy to above.
We will formalise these intuitions in the next two subsections.

\insertFigSets

\subsection{Discrete $\Xcal$}
\label{sec:analysisDisc}

We first analyse the case when $\Xcal$ is a discrete subset of $[0,r]^d$.
Denote $\Deltamx = \funcopt - \funcm(x) - \zetam$ and
$\Jcalme = \{x\in\Xcal; \Deltamx\leq \eta\}$.
Note that $\Deltam > 0$ for all $m$ by our assumptions.
% Let $\rho > 1$ be given.
Central to our analysis will be the partitioning $(\Hcalm)_{m=1}^M$ of $\Xcal$.
First define
$\Hcalone = \Jcalcmmee{1}{3\gamma}=\{x: \funcone(x) < \funcopt - \zetaone -
3\gammaone\}$ to be the arms whose $\funcone$ value
is at least $\zetaone + 3\gammaone$ below the optimum $\funcopt$.
Then recursively define,
\begin{align*}
&\Hcalm = \Jcalcmthg \cap \left(\bigcap_{\ell=1}^{m-1}\Jcallthg \right) 
\hspace{0.1in} \textrm{for } 2\leq m\leq M-1,
\hspace{0.4in}
\HcalM = \bigcap_{\ell=1}^{M-1}\Jcallthg.
\numberthis \label{eqn:HcalDefn}
\end{align*}
In addition to the above, we will also find it useful to define the sets
``above" $\Hcalm$ as $\Hcalupm = \bigcup_{\ell=m+1}^{M}\Hcall$ and the sets
``below" $\Hcalm$ as $\Hcaldownm = \bigcup_{\ell=1}^{m-1}\Hcall$.
% We will see that $\Hcalm$ is the set of points in $\Xcal$ that will be points that
% can be safely excluded from queries beyond the $m$\superscript{th} fidelity  using
% information at the $m$\ssth fidelity.
Our analysis reveals that most of the capital invested at points in $\Hcalm$ will be due to
queries to the $m$\ssth fidelity function $\funcm$.
$\Hcaldownm$ is the set of points that can be excluded from queries at fidelities
$m$ and beyond due to information from lower fidelities.
$\Hcalupm$ are points that will be queried  at fidelities higher than $m$
several times.
In the $2$ fidelity setting described in Section~\ref{sec:analysis},
$\Xcalrg = \Hcaltwo$ and $\complement{\Xcalrg} = \Hcalone = \Hcaldownmm{2}$.
We have illustrated these sets in Figure~\ref{fig:Xcalms}.

Recall that $\nCOST = \floor{\COST/\costM}$ is the number of queries by a single-fidelity
method; it is a lower bound on $N$, the number of queries by a multi-fidelity method.
Similarly, $\nCOSTtop = \floor{\COST/\costone}$ will be an upper bound on $N$.
We will now define two quantities $\COSTone,\COSTtwo$ where $\COSTone<\COSTtwo$.
We will show improved simple regret over \gpucbs when the capital $\COST$ is larger
than these quantities, with the
$\COST>\COSTtwo$ regime being better by an additive
$\log(\costM/\costone)$ factor over the $\COST>\COSTone$ case.
Formally, we define $\COSTone$ to be the smallest $\COST$ satisfying the following
condition,
\begin{align*}
% \costM|\Xcal| \;+\;
\sum_{m=2}^M \costmm{m} |\Hcalmm{m-1}|\,+\,
 \sum_{m=1}^{M-1} \costm |\Hcalm \cup \Hcalupm|
\left\lceil\frac{\eta^2}{{\gammam}^2}\betancosttop \right\rceil
\;\leq\; \frac{\COST}{2},
\label{eqn:costonecondn} \numberthis
\end{align*}
and $\COSTtwo$ to be the smallest $\COST$ satisfying the following condition,
\begin{align*}
\costM|\Xcal| \;+\;
  \costM\summMmo
|\Hcalm \cup \Hcalupm|\left\lceil\frac{\eta^2}{\gammamsq}\betancosttop\right\rceil
\;\leq\; \frac{\COST}{2}.
\label{eqn:costtwocondn} \numberthis
\end{align*}
We can find such $\COSTone, \COSTtwo$, since for fixed $\gammam$'s, in both cases,
the right side is linear in $\COST$ and the left is logarithmic
since $\betan\asymp\bigO(\logn)$
and $\nCOSTtop \asymp \COST$.
Since $\{\Hcalm\}_{m=1}^M$ form a partition of $\Xcal$ and
$\costone<\dots<\costM$, we see that $\COSTone < \COSTtwo$.
Recall that at the initial stages, \mfgpucbs has infinite simple regret since the
evaluations are at lower fidelities.
$\COST>\COSTone$ indicates the phase where $\bigTheta(\nCOST)$ evaluations have been
made inside $\HcalM$, but the total number of evaluations $N$ could be much larger.
When $\COST>\COSTtwo$, we have reached a phase where $N$ is also in $\bigTheta(\nCOST)$.

Moreover, note that
when the approximations are good, i.e. the sets $\Hcalm$ are small,
both $\COSTone$ and $\COSTtwo$ are small.
$\COSTone$ is also small when the approximations are cheap, i.e. $\costm$'s are small.
Therefore, the cheaper and better the approximations,
we have to wait less time (for fixed $\gammam$)
before \mfgpucbs starts querying at the $M$\ssth fidelity and achieves
good regret.

% 
% Finally we note that the fraction $1/2$ in the right sides of both~\eqref{eqn:costonecondn}
% and~\eqref{eqn:costtwocondn} can be replaced by any constant $<1$.

We now state our main theorem for discrete $\Xcal$.
To simplify the analysis, we will introduce an additional condition in the fidelity
selection criterion in step 2 of Algorithm~\ref{alg:mfgpucb}.
We will always evaluate $\funcm$ at $\xt$ only if $\xt$ has been evaluated at
all lower fidelities, $1,\dots,m-1$;
precisely, that $\mt = \min_m\{\, m\, | \betath\sigmamtmo(\xt) \geq \gammam
      \;\;\textrm{or}\;\; m = M\;\;\textrm{or}\;\;
      \Tmmttxx{m}{n}{\xt} = 0 \}$.
Both this condition,
and the dependence of $\COSTtwo$ on $|\Xcal|$ in~\eqref{eqn:costtwocondn}
are an artefact of our analysis.
They arise only because we do not account for the correlations between
the arms in our discrete analysis; doing so requires us to make assumptions about the
locations of the arms in $[0,r]^d$.
We will not need this condition or have
$\COSTtwo$ depend on $|\Xcal|$ for the continuous case.

% \vspace{\thmprespacing}

\begin{theorem}
\label{thm:discreteThm}
Let $\Xcal$ be a discrete subset of $[0,r]^d$. Let $\funcm\sim\GP(\zero,\kernel)$
for all $m$. Assume that $\funcm$'s
satisfy assumptions \emph{\Gone}, \emph{\Gtwo} and $\kernel$ satisfies
Assumption~\ref{asm:kernelAssumption}.
Pick $\delta\in(0,1)$ and run \emph{\mfgpucbs} (Algorithm~\ref{alg:mfgpucb}) with
% $\betat = 2\log\left(\frac{M|\Xcal|\pi^2t^2}{3\delta}\right)$.
$\betat = 2\log\big(M|\Xcal|\pi^2t^2/(3\Gtwoprob\delta) \big)$.
% Then, for all $\alpha\in(0,1)$, there exists $\rho,\COSTnought$ such that,
% with probability $>1-\delta$ we have for all $\COST\geq \COSTnought$,
Then, we have the following bounds on $S(\COST)$ with
$\PP$-probability greater than $1-\delta$.
\begin{align*}
\text{for all $\COST>\COSTone$},\quad &
S(\COST)\;\leq\; \sqrt{\frac{2C_1\beta_{\nCOSTtop}\IGnn{\nCOST}(\HcalM)}{\nCOST}} \\
% \;+\; \sqrt{\frac{2C_1|\Xcal|\beta_{2\nCOST}\IGnn{2\nCOST^\alpha}(\Xcal)}
%       {\nCOST^{2-\alpha}}}.
\text{for all $\COST>\COSTtwo$},\quad &
S(\COST)\;\leq\; \sqrt{\frac{2C_1\beta_{2\nCOSTbot}\IGnn{\nCOST}(\HcalM)}{\nCOST}}
\end{align*}
Here $C_1 = 8/\log(1+\eta^2)$ is a constant, $\nCOST=\floor{\COST/\costM}$,
$\nCOSTtop = \floor{\COST/\costone}$,
and $\Gtwoprob$ is from~\eqref{eqn:gtwoprob}.
\end{theorem}

The difference between the two results is the $\betancosttop$ dependence in the former
setting and $\betancost$ in the latter;
the latter bound is better by an additive $\log(\costM/\costone)$ term, but we have to
wait for longer.
% First note that $\IGn(\cdot) \asymp \IGnn{2n}(\cdot)$
% and that $\beta_{n}\asymp\beta_{2n}\asymp\logn$.
Dropping constant and $\polylog$ terms and comparing to the result in
Theorem~\ref{thm:gpucb} reveals that we outperform
\gpucbs by a factor of $\sqrt{\IGnc(\HcalM)/\IGnc(\Xcal)}
\asymp\sqrt{\vol(\HcalM)/\vol(\Xcal)}$ asymptotically.
The set $\HcalM$ from~\eqref{eqn:HcalDefn} is determined by the
$\zetaone,\dots.\zetamm{M-1}$
values, the
approximations $\funcone,\dots,\funcmm{M-1}$ and the parameters
$\gammaone,\dots,\gammamm{M-1}$.
The better the approximations, the smaller the set $\HcalM$ and there is more advantage
over single fidelity strategies.
In Figure~\ref{fig:zetaHmIllus}, we have shown the ratio
$\vol(\Hcaltwo)/\vol(\Xcal)$ for a two fidelity problem as $\zetaone$ decreases---the
figure corroborates our claim that the rates improve as the $\zetam$ values decrease.
As the approximations worsen, the advantage to multi-fidelity optimisation diminishes as
expected, but we are never worse than \gpucbs up to constant factors.

% \modifjan{
A few remarks are in order.
First, note that the dependence on $\nCOST$ (or equivalently $\COST$) is the same
for both \gpucbs and \mfgpucb.
In fact, one should not expect multi-fidelity optimisation to yield ``rate'' improvements
since such $\sqrt{1/n}$ dependencies are typical in the bandit
literature~\citep{bubeck2011x,shang2017adaptive}.
The multi-fidelity framework allows us to find a good region, i.e. $\HcalM$, where the
optimum exisits, and as such, we should expect the improvements to be in terms of the
size of this set, relative to $\Xcal$.
Second, the bound is given in terms of $\HcalM$ which, as illustrated by
Figure~\ref{fig:zetaHmIllus},
gives us insight into the types of gains we can expect from
multi-fidelity optimisation.
However, $\HcalM$ is a random quantity and obtaining high probability bounds
on its volume could shed more light on the gains of our multi-fidelity
optimisation framework; this is an interesting avenue for future work.
% }

\insertzetaHmFigure

\vspace{0.05in}

\noindent {\bf Choice of $\gammam$.} It should be noted that an ``optimal''
choice of $\gammam$ depends on the available budget,
i.e. how long we are willing to wait before achieving non-trivial regret.
If we are willing to wait long, we can afford to choose small $\gammam$
and consequently have better guarantees on the regret.
This optimal choice also depends on several unknown problem dependent factors --
such as the sizes of the sets $\Hcalm$. In~\citet{kandasamy2016mfbandit}, the choice 
$\gammam = \zetam \sqrt{\costm/\costmm{m+1}}$ was used which ensures that for an arm
$x\in\Hcalm$, the cost spent at lower fidelities $1,\dots, m-1$ is not more than
the cost spent at fidelity $m$.
Beyond this intuitive property, this choice further achieves a lower bound on the
$K$-armed multi-fidelity problem. 
The same choice for $\gammam$ here ensures that the cost spent at the lower
fidelities is not more than an upper bound on the cost spent at fidelity $m$ --
we have elaborated more in Remark~\ref{rmk:discgammamchoice} after our proofs. 
% Another option would be to choose $\gammam$ to balance the
% numerator and denominator in equations~\eqref{eqn:costonecondn}
% and~\eqref{eqn:costtwocondn}.
% This however, requires an understanding of how the sizes of the sets
% $\Hcalm$ varies with $\gammam$ values, 
% and, such a chracterisation seems difficult, even in simplified settings such as
% uncorrelated Gaussians.
% 
We have empirically demonstrated the effect of different choice of $\gammam$ values
via an experiment in Figure~\ref{fig:hartmanngamma}.
Building on these ideas, an explicit prescription for the choice of $\gammam$ is bound
to be a fruitful avenue of research, and we leave this to future work.
In the meanwhile, in Section~\ref{sec:implementation}, we describe a heuristic
for adaptively choosing $\gammam$ adaptively which worked well in our experiments.

% \begin{corollary}
% \label{cor:discreteThmGamma}
% Assume the same conditions as Theorem~\ref{thm:discreteThm}.
% Suppose we choose $\gammam =$
% \end{corollary}

\subsection{Continuous and Compact $\Xcal$}
\label{sec:analysisCts}

% As for discrete $\Xcal$ we will begin with a partitioning of $\Xcal$.
% The partitioning will depend on the $\gammam$ values as before, but in addition will
% also depend on a parameter $\tau>0$. 
% Let $\Jcalme = \{x\in\Xcal; \Deltamx\leq \eta\}$ be as defined in the discrete case. 
% We make the dependence on $\tau$ explicit and define $\Htcalone =
% \Jcalcmmee{1}{\maxtrg} = \{x: \funcone(x) < \funcopt - \zetaone -
% \max(\tau,\rho\gammaone)\}$ to be the subset whose $\funcone$ value
% is at least $\zetaone + \max(\tau,\rho\gammaone)$ below the optimum $\funcopt$.
% Then define,
% \begin{align*}
% \hspace{-0.1in}
% \Htcalm = \Jcalcmtrg \cap \bigg(\bigcap_{\ell=1}^{m-1}\Jcalltrg \bigg) 
% % \hspace{0.1in}
% \;
% \textrm{for } 2\leq m\leq M-1,
% \hspace{0.3in}
% \HtcalM = \bigcap_{\ell=1}^{M-1}\Jcalltrg.
% \label{eqn:HtcalmDefn}
% \numberthis
% \end{align*}
% The purpose of $\tau$ is to create a ``separation'' between the function
% values in the sets.
% In the discrete case, this separation is guaranteed as the function values $\funcm(x)$
% is a discrete set.
% This however, is not the case in continuous settings. 
% The requirement $\tau>0$ is mostly a technical condition and is otherwise orthogonal
% to the gist of our theorems.
% Indeed, our theorems hold for all $\tau >0$.

We define the sets $\Hcalm, \Hcalupm$ for $m=1,\dots,M$ as in the discrete case.
Let $\{\nun\}_{n\geq 0}$ be any sublinear sequence such that $\nun\rightarrow\infty$.
Let
\begin{align*}
\Htncalm = \left\{x\in\Xcal:\; {\rm B}_2\big(x, r\sqrt{d}/\nun^{\frac{1}{2d}}\big)
 \,\cap \,
\Htcalm \neq \emptyset \indAnds x \notin \Hcalupm \right\}
% \numberthis \label{eqn:Htncalmdefn}
\end{align*}
to be a $\nun$-dependent $L_2$ dilation of $\Htncalm$ by $r\sqrt{d}/%
\nun^{\frac{1}{2d}}$.
Here, ${\rm B}_2(x, \epsilon)$ is an $L_2$ ball of radius $\epsilon$ centred at $x$.
% The sets $\{\Htncalm\}_{m=1}^M$ depend on $\rho, \gamma, \tau, n$ and $\alpha$.
Notice that as $n\rightarrow\infty$, $\Htncalm \rightarrow \Htcalm$.
% In addition to the above,
% denote the $\epsl$ covering number of a set $A\subset \Xcal$ in the $\|\cdot\|_2$ 
% metric by $\Omegaii{\epsl}(A)$.
Similar to the discrete case,
we define $\COSTone$ to be the smallest $\COST$ satisfying the following the condition,
\begin{align*}
% \costM|\Xcal| \;+\;
\costM\nuncosttop
\,+\,
C_\kernel\eta^2\beta_{\nCOSTtop}^{p+1}
 \sum_{m=1}^{M-1} \costm \frac{\vol(\Hcal_{\nCOST}^{(m)} \cup \Hcalupm)}{{\gammam}^{2p}}
% \left\lceil\frac{\eta^2}{{\gammam}^2}\betancosttop \right\rceil
\;\leq\; \frac{\COST}{2},
\label{eqn:costonects} \numberthis
\end{align*}
and $\COSTtwo$ to be the smallest $\COST$ satisfying the following condition,
\begin{align*}
\costM\nuncosttop
\,+\,
C_\kernel\eta^2\beta_{\nCOSTtop}^{p+1} \costM
 \sum_{m=1}^{M-1} \frac{\vol(\Hcal_{\nCOST}^{(m)} \cup \Hcalupm)}{{\gammam}^{2p}}
\;\leq\; \frac{\COST}{2}.
\label{eqn:costtwocts} \numberthis
\end{align*}
Here $p=1/2$ for the SE kernel and $p=1$ for the \matern{} kernel.
$C_\kernel$ is a kernel dependent constant elucidated in our proofs;
for the SE kernel, $C_\kernel = 2^{2+d/2}(d\kernel_0/h^2)^{d/2}$ where
$\kernel_0, h$ are parameters of the kernel.
Via a reasoning similar to the discrete case we see that $\COSTone<\COSTtwo$.
Our main theorem is as follows.
% \insertprespacing

\begin{theorem}
\label{thm:compactThm}
Let $\Xcal\subset[0,r]^d$ be compact and convex. Let $\funcm\sim\GP(\zero,\kernel)
\;\forall\,m$,
and satisfy assumptions \emph{\Gone}, \emph{\Gtwo}.
Let $\kernel$ satisfy
Assumption~\ref{asm:kernelAssumption} with some constants $a,b$.
Pick $\delta\in(0,1)$ and run \emph{\mfgpucbs} (Algorithm~\ref{alg:mfgpucb}) with
\[
\betat =
2\log\left(\frac{M\pi^2t^2}{2\Gtwoprob\delta}\right) + 4d\log(t) + 
\max\left\{\,0\,,\, 
2d\log\left(brd\sqrt{\log\bigg(\frac{6Mad}{\Gtwoprob\delta}\bigg)} \right)\right\}.
\]
Then, we have the following bounds on $S(\COST)$ with $\PP$-probability greater than
$1-\delta$.
% Then, for all $\alpha\in(0,1)$ and $\tau>0$, 
% there exists $\rho,\COSTnought$ such that with probability at least $1-\delta$ we have
% for all $\COST \geq \COSTnought$,
\begin{align*}
\text{for all $\COST>\COSTone$},\quad &
S(\COST)\;\leq\; \sqrt{\frac{2C_1\beta_{\nCOSTtop}
\IGnn{\nCOST}(\Hcal_{\nCOST}^{(M)})}{\nCOST}} + \frac{\pi^2}{3\nCOST}\\
\text{for all $\COST>\COSTtwo$},\quad &
S(\COST)\;\leq\; \sqrt{\frac{2C_1\beta_{2\nCOSTbot}\IGnn{\nCOST}(\Hcal_{\nCOST}^{(M)})}%
                            {\nCOST}} + \frac{\pi^2}{3\nCOST}
\end{align*}
Here $C_1 = 8/\log(1+\eta^2)$ is a constant, $\nCOST=\floor{\COST/\costM}$,
and $\nCOSTtop = \floor{\COST/\costone}$.
% number of queries by an algorithm which ignores the lower fidelities.
% $\rho$ depends only on $\alpha$ and satisfies
% $\rho > \rho_0 = \max\{2, 1 + \sqrt{(1+2/\alpha)/(1+d)}\}$.
% $\COSTnought$ depends on several problem dependent quantities including
% the volumes of the sets $\Htcalm$ and consequently on $\rho$.
\end{theorem}
Note that the sets $\HtnccalM$ depend on the sublinear
increasing sequence $\{\nun\}_{n\geq 0}$ -- the theorem is valid for any such choice of
$\nun$.
% When the $\nun$ values are large, as per~\eqref{eqn:Htncalmdefn},
% there is less inflation in the $\Htncalm$ sets from $\Hcalm$.
% Hence, the second term of the LHS in equations~\eqref{eqn:} is large
The comparison of the above bound against \gpucbs is similar to the discrete case.
The main difference is that we have an additional dilation of $\HtcalM$ to $\HtnccalM$
which occurs due to a covering argument in our analysis.
Recall that $\Htnccalm\rightarrow\Htcalm$ as $\COST\rightarrow\infty$.
The bound is determined by the MIG of the set $\HtnccalM$, which is small when the
approximations are good.

\section{Some Implementation Details of \mfgpucbs and other Baselines}
\label{sec:implementation}

Our implementation uses some standard techniques in the Bayesian optimisation literature
given below.
In addition, we describe the heuristics used to set the $\gammam,\zetam$ parameters of
our method.

\begin{itemize}[leftmargin=0.0in]
\vspace{-0.05in}
\item[]
\textbf{Initialisation:}
Following recommendations in~\citet{brochu12bo}
all GP methods were initialised with uniform random
queries using an initialisation capital $\COST_0$.
For single fidelity methods, we used it at the $M$\ssth fidelity, whereas for
multi-fidelity methods we used $\COST_0/2$ at the first fidelity and $\COST_0/2$ at
the second fidelity.

\textbf{Kernel:}
In all our experiments, we used the SE kernel.
We initialise the kernel by maximising the GP marginal
likelihood~\citep{rasmussen06gps} on the initial sample and then update the kernel
every $25$ iterations using marginal likelihood.

\vspace{-0.05in}
\item[]
\textbf{Choice of $\betat$:}
$\betat$, as specified in Theorems~\ref{thm:gpucb},~\ref{thm:compactThm} has
unknown constants and tends to be too conservative in practice.
Following~\citet{kandasamy15addBO} we use $\betat = 0.2d\log(2t)$ which captures
the dominant dependencies on $d$ and $t$.

\vspace{-0.05in}
\item[]
\textbf{Maximising $\ucbt$:}
We used the \directs algorithm~\citep{jones93direct}.

\vspace{-0.05in}
\item[]
% \noindent
\textbf{Choice of $\zetam$'s:}
Algorithm~\ref{alg:mfgpucb} assumes that the $\zetam$'s are given with the problem
description, which is hardly the case in practice.
In our implementation, instead of having to deal with $M-1$, $\zetam$ values we will
assume  $\|\funcm - \funcmm{m-1}\|_\infty \leq \zeta$. This satisfies assumption \Gtwos
with $(\zetaone,\zetatwo,\dots,\zetamm{M-1}) = ((M-1)\zeta, (M-2)\zeta,\dots, \zeta)$.
This allows us to work with only one value of $\zeta$.
We initialise $\zeta$ to a small value, $1\%$ of the range of initial queries.
Whenever we query at any fidelity
$m>1$ we also check the posterior mean of the $(m-1)$\ssth fidelity.
If $|\funcm(\xt) - \mummtt{m-1}{t-1}(\xt)| > \zeta$, we query again at $\xt$, but at
the $(m-1)$\ssth fidelity. If $|\funcm(\xt) - \funcmm{m-1}(\xt)| > \zeta$, we update
$\zeta$ to twice the violation.

\vspace{-0.05in}
\item[]
\textbf{Choice of $\gammam$'s:}
The role of the $\gammam$ values at each fidelity is to ensure that we do not spend too
much effort at the lower fidelities, where if $\gammam$ is too small, \mfgpucbs spends
a large number of queries at fidelity $m$ to reduce the variance below $\gammam$.
This might cause \mfgpucbs to spend an unnecessarily large number of evaluations
at fidelity $m$.
% Hence, we use the following intuition: if the algorithm, is stuck at 
% fidelity  $m$ for too long, then $\gammam$ is probably too small.
Hence, we start with small values for all $\gammam$.
However, if the algorithm does not
query above the $m$\ssth fidelity for more than $\costmm{m+1}/\costmm{m}$ 
iterations, we double $\gammam$.
All $\gammam$ values were initialised to $1\%$ of the range of initial queries.

\end{itemize}

Whilst the first four choices are standard in the BO
literature~\citep{brochu12bo,snoek12practicalBO}, our methods for selecting
the $\zetam$ and $\gammam$ parameters are heuristic in nature.
We obtained robust implementations of \mfgpucbs with
little effort in tweaking these choices.
In fact, we found our implementation was able to recover even from fairly bad
approximations at the lower fidelities (see experiment in Figure~\ref{fig:badCurrin}).
We believe that other reasonable heuristics can also be used in place of our choices here, and a systematic investigation into protocols for the same will be a fruitful avenue for future research.

\section{Experiments}
\label{sec:experiments}

We present experiments for compact and continuous $\Xcal$ since it is the more
practically relevant setting.
We compare \mfgpucbs to the following baselines.
\textbf{Single fidelity methods:}
\gpucb; 
\gpei: the expected improvement criterion for BO~\citep{jones98expensive};
\direct: the dividing rectangles method~\citep{jones93direct}.
\textbf{Multi-fidelity methods:}
\mfnaive: a naive baseline where we use \gpucbs to query at the 
\emph{first} fidelity a large number of times and then query at the last fidelity 
at the points queried at $\funcone$ in decreasing order of $\funcone$-value;
\mfsko: the multi-fidelity sequential kriging method from~\citep{huang06mfKriging}.
Previous works on multi-fidelity methods (including \mfsko) had not made their code
available and were not straightforward to implement.
% Hence, we could not compare 
%  to all of them.
We discuss this more in Appendix~\ref{app:baselines} 
along with some other single and multi-fidelity baselines we tried but excluded in the
comparison to avoid clutter in the figures.
We also detail some design choices and hyper-parameters for the
baselines in Appendix~\ref{app:baselines}.

\subsection{Synthetic Examples}

\insertSRToyAll

We begin with a series of synthetic experiments, designed to demonstrate the
applicability and limitations of \mfgpucb.
We use the Currin exponential ($d=2$), Park ($d=4$) and Borehole ($d=8$) functions
in $M=2$ fidelity experiments and the Hartmann functions in $d = 3$ and $6$ with
$M=3$ and $4$ fidelities respectively. The first three functions are taken from
previous multi-fidelity literature~\citep{xiong13highAccuracy} while we tweaked the
Hartmann functions to obtain the lower fidelities for the latter two cases.
In Appendix~\ref{app:experiments} we give the formulae for these functions and the
approximations used for the lower fidelities.
We show the simple regret $S(\COST)$ against capital $\COST$ in
Figure~\ref{fig:toySRAll}.
The number of fidelities and the costs used for each fidelity are also
given in Figure~\ref{fig:toySRAll}.
\mfgpucbs outperforms other baselines on all problems.

The last panel of Figure~\ref{fig:toySRAll} shows a histogram of the number of queries
at each fidelity after $184$ queries of \mfgpucb,
for different ranges of $\funcmm{3}(x)$ for the Hartmann-3D function.
Many of the queries at the low $\funcmm{3}$ values are at fidelity $1$,
but as we progress they decrease and the second fidelity queries increase.
The third fidelity dominates very close to the optimum but is used sparingly
elsewhere.
This corroborates the prediction in our analysis that \mfgpucbs uses low
fidelities to explore and successively higher fidelities at 
promising regions to zero in on $\xopt$.
(Also see Figure~\ref{fig:mfSimLong}.)

A common occurrence with \mfnaives was that once we started querying at fidelity $M$,
the regret barely decreased.
The diagnosis in all cases was the same: it was stuck around the
maximum of $\funcone$ which is suboptimal for $\funcM$.
This suggests that while we have cheap approximations,
the problem is  by no means trivial. As explained previously, it is also important to
\emph{explore} at higher fidelities to achieve good regret.
The efficacy of \mfgpucbs when compared to single fidelity methods
is that it confines this exploration to a small set containing the optimum.
In our experiments we found that \mfskos did not consistently beat other single
fidelity methods.
Despite our best efforts to reproduce \mfsko, we found it to be quite brittle.
In fact, we also tried another multi-fidelity method and found that it did not perform as
desired (See Appendix~\ref{app:baselines} for details).
% 
% In Appendix~\ref{app:additionalExperiments} we present additional experiments to test
% our implementation of \mfgpucb.
% We study how it performs under bad approximations and
% when the cost of the approximation varies.

\insertFigBoreHoleLambdaComp

\vspace{0.05in}

\noindent
\textbf{Effect of the cost of the approximations:}
We now test the effect the cost of the approximation on performance.
Figure~\ref{fig:bhlambda} shows the results when \mfgpucbs was run on the 
$2$-fidelity Borehole experiment for different costs for the approximation $\funcone$.
We fixed $\costtwo = 1$ and varied $\costone$ between $0.01$ to $0.5$.
As $\costone$ increases, the performance worsens as expected.
At $\costone=0.5$ it is indistinguishable from \gpucbs as the overhead of managing $2$
fidelities becomes significant when compared to the improvements of using the
approximation.

\vspace{0.05in}

\noindent
\textbf{Bad Approximations:}
% \subsection{Bad Approximations:}
% \label{app:badApproximations}
It is natural to ask how \mfgpucbs performs with bad approximations at lower
fidelities. 
We found that our implementation with the heuristics suggested in
Section~\ref{sec:implementation} to be quite robust.
We demonstrate this using the Currin exponential function, but using the negative of
$\functwo$ as the first fidelity approximation, i.e. $\funcone(x) = - \functwo(x)$. 
Figure~\ref{fig:badCurrin} illustrates $\funcone,\functwo$ and gives the simple
regret $S(\COST)$.
% (See Fig.~\ref{fig:badCurrinFunctions} for an illustration). 
% The simple regret is in Fig.~\ref{fig:badCurrin}. 
Understandably, it loses to the single fidelity methods since the first fidelity
queries are wasted and it spends some time at the second fidelity recovering from the
bad approximation. However, it eventually is able to achieve low regret.

\vspace{0.05in}

\noindent
\textbf{Effect of threshold values on \mfgpucb:}
We now demonstrate the effect of different choices for $\gammaone$ on \mfgpucbs as
described in Algorithm~\ref{alg:mfgpucb}.
We use the $3$ dimensional Hartmann function in a $2$ fidelity set up where
$\zetaone\approx0.112$, $\costone=1$ and $\costtwo=10$.
The implementation follows the description in Section~\ref{sec:implementation},
except that the true $\zetaone$ value is made known to \mfgpucbs and the threshold
value $\gammaone$ is kept fixed at values $0.03, 0.1, 0.3, 1.0$.
The result is shown in Figure~\ref{fig:hartmanngamma}.
We see that as $\gammaone$ decreases the curves start later in the figure indicating that
\mfgpucbs spends more time at the approximation $\funcone$
before proceeding to $\functwo$; however, the simple regret is also generally better
for smaller $\gammaone$.
Therefore, if we have a large computational budget and are willing to wait longer,
we can choose small $\gammam$ values and achieve better simple regret.

\subsection{Model Selection and Astrophysics Experiments}
% \vspace{-0.1in}

We now present results on three hyper-parameter tuning tasks  and a maximum likelihood 
inference task in Astrophysics.
We compare methods on computation time since that is the ``cost" in all experiments.
We include the processing time for each method in the comparison 
(i.e. the cost of determining the next query). 
The results are given in Figure~\ref{fig:realAll}, where, as
we see \mfgpucbs outperforms other baselines on all tasks.
The experimental set up for each optimisation problem is described below.

\insertBadCurrin
\insertFigGammaHartmannComp

\begin{itemize}[leftmargin=0.0in]

\item[]
\textbf{Classification using SVMs (SVM):}
We trained a Support vector classifier on the magic gamma dataset using the sequential
minimal optimisation algorithm to an accuracy of $10^{-12}$. 
The goal is to tune the kernel bandwidth and the soft margin coefficient
in the ranges $(10^{-3}, 10^1)$ and $(10^{-1}, 10^5)$ respectively on a dataset of
size $2000$. 
We set this up as a $M=2$ fidelity experiment with the entire training set 
at the second fidelity and $500$ points at the first.
Each query to $\funcm$ required $5$-fold cross validation on the respective training sets.

\item[]
\textbf{Regression using additive kernels (SALSA):}
We used the SALSA method for additive kernel ridge regression~\citep{kandasamy15salsa}
on the  $4$-dimensional coal power plant dataset. We tuned
the $6$ hyper-parameters --the regularisation penalty, the kernel scale and
the kernel bandwidth for each dimension-- each in the range $(10^{-3},10^{4})$ using
$5$-fold cross validation.
This experiment used $M=3$ and $2000, 4000, 8000$ points at each fidelity
respectively.

\item[]
\textbf{Viola \& Jones face detection (V\&J):}
The Viola \& Jones cascade face classifier~\citep{viola01cascade}, which uses a
cascade of weak classifiers, is a popular method for face detection.
To classify an 
image, we pass it through each classifier. If at any point the classifier score falls
below a threshold, the image is classified as negative. If it passes through the
cascade, then it is classified as positive.
One of the more popular implementations comes
with OpenCV and uses a cascade of $22$ weak classifiers. 
The threshold values in the OpenCV implementation are pre-set based on some heuristics 
and there is no reason to think they are optimal for a given face detection problem. 
The goal is to tune these
$22$ thresholds by optimising them over a training set. 
We modified the OpenCV implementation to take in the thresholds as parameters.
As our domain $\Xcal$ we chose a neighbourhood around the configuration used in
OpenCV.
We set this up as an $M=2$ fidelity experiment where
the second fidelity used $3000$
images from the Viola and Jones face database and the first used just $300$.
Interestingly, on an independent test set, the configurations found by \mfgpucbs
consistently achieved over $90\%$ accuracy while the OpenCV configuration
achieved only $87.4\%$ accuracy.

\insertRealAll

\item[]
\textbf{Type Ia Supernovae (Supernova):} 
We use Type Ia supernovae data from~\citet{davis07supernovae} for maximum
likelihood inference on $3$ cosmological parameters, the Hubble constant
$H_0\in(60,80)$, the dark matter fraction $\Omega_{M}\in(0,1)$ and the dark energy
fraction $\Omega_{\Lambda}\in (0,1)$. 
Unlike typical parametric maximum likelihood problems we see in machine learning,
the likelihood is only available as a black-box.
It is computed using the Robertson--Walker
metric~\citet{davis07supernovae} which requires a (one dimensional) numerical integration 
for each sample in the dataset.
We set this up as a $M=3$ fidelity task. At the third fidelity, the
integration was performed using the trapezoidal rule on a grid of size $10^6$.
For the first and second fidelities, we used grids of size $10^2, 10^4$ respectively.
The goal is to maximise the likelihood at the third fidelity.
% The results are given in Fig.~\ref{fig:davisSN}.

\end{itemize}

% We conclude this section with another cautionary note 
% The multi-fidelity framework is not suitable in all cases. To explain this, we
% revert to the KDE example in Figure~\ref{fig:kdeEg}.
% Had we known beforehand that the optimum was to lie in the boxed region, a
% multi-fidelity method would be inappropriate as the difference between the
% fidelities is large when compared to the range of the functions in that region.
% Such instances where strong domain knowledge is available weakens the 
% case for mutli-fidelity methods for hyper-parameter tuning. 

\section{Proofs}
\label{sec:proofs}

In this section we present the proofs of our main theorems. 
While it is self contained,
the reader will benefit from first reading the more intuitive discussion in
Section~\ref{sec:analysis}.
The goal in this section is to bound the simple regret $S(\COST)$
given in~\eqref{eqn:defnMFSR}.
% Recall that $S(\COST) = \min_{\substack{t:\mt = M}}\funcopt - \funcM$ 
% % \begin{align*}
% %   \funcopt - \funcM(\xt)
% % \end{align*}
Recall that $N$ is the random number of plays within capital $\COST$.
While $N\leq\floor{\COST/\costone}$ is a trivial upper bound for $N$, this will be too
loose for our purposes.
In fact, we will show that after a sufficiently large number of queries at any
fidelity, the number of queries at fidelities smaller than $M$ will be sublinear in $N$.
Hence $N\in\bigO(\nCOST)$ where
$\nCOST=\floor{\COST/\costM}$ is the number of plays by any algorithm that 
operates only  at the highest fidelity.

We introduce some notation to keep track of the evaluations at each fidelity in \mfgpucb.
After $n$ steps, we will have queried multiple times at any of the $M$ fidelities.
$\Tmnx$ denotes the number of queries at $x\in\Xcal$ at
fidelity $m$ after $n$ steps.
$\Tmnxx{A}$ denotes the same for a subset $A\subset\Xcal$.
$\Dcalmn = \{(\xt,\yt)\}_{t:\mt=m}$ is the set of query-value pairs
at the $m$\superscript{th} fidelity until time $n$.

\vspace{0.1in}
\noindent
\textbf{Roadmap:}
To bound $S(\COST)$ in both the discrete and continuous settings,
we will begin by studying the algorithm after $n$ evaluations at any
fidelity and analyse the following quantity,
\begin{align*}
% \Rtilden = \sum_{\substack{t=1\\t\,:\,\mt=M}}^n \big(\funcopt - \funcM(\xt) \big).
\numberthis\label{eqn:Rtilden}
\Rtilden &=\;
    {\sum_{\substack{t:\mt=M\\ \xt\in\goodset}}
      \left(\funcopt - \funcM(\xt) \right)}
\end{align*}
Readers familiar with the bandit literature will see that this is similar to the notion
of \emph{cumulative regret}, except we only consider queries at the $M$\ssth fidelity
and inside a set $\goodset\subset\Xcal$.
$\goodset$ contains the optimum and generally 
%  which contains $\xopt$ and
has high value for the payoff function $\funcM(x)$;
it will be determined by the approximations provided
via the lower fidelity evaluations.
% Then, we decompose $\Rtilden$ as follows,
% \begin{align*}
% \Rtilden &=\;
%     \underbrace{\sum_{\substack{t:\mt=M\\ \xt\in\goodset}}
%       \left(\funcopt - \funcM(\xt) \right)}_{\Rtildenone}
%   \;\,+\,\;
%   \underbrace{\sum_{\substack{t:\mt=M\\ \xt\in\badset}}
%     \left(\funcopt - \funcM(\xt)\right)}_{\Rtildentwo}.
% \numberthis\label{eqn:RnDecomp}
% \end{align*}
% $\Rtildenone$ is the cumulative regret at the $M$\ssth fidelity due to queries in
% $\goodset$ and $\Rtildentwo$ is due queries outside $\goodset$.
% We will bound $\Rtildenone,\Rtildentwo$ separately using the MIGs
% (Definition~\ref{def:infGain}) of $\goodset$ and $\badset$;
% for this, we will use the techniques in~\citet{srinivas10gpbandits}.
We will show that most of the $M$\ssth fidelity
evaluations will be inside $\goodset$ in the multi-fidelity setting, and
hence, the regret for \mfgpucbs will scale with $\IGn(\goodset)$ instead of
$\IGn(\Xcal)$ as is the case for \gpucb.
Finally, to convert this bound in terms of $n$ to one that depends on $\COST$,
we show that both the total
number of evaluations $N$ and the number of highest fidelity evaluations $\TMNXcal$
are on the order of $\nCOST$ when $\COST$ is sufficiently large.
For this, we bound the number of plays at the lower fidelities
(see Lemma~\ref{lem:TmnxBoundsInformal}).
Then $S(\COST)$ can be bounded by,
\begin{align*}
S(\COST) \;\leq\; \frac{1}{\TMNXcal}\RtildeN \;\lesssim\;
\frac{1}{\nCOST}\Rtildenc.
\numberthis \label{eqn:SimCumRelation}
\end{align*}
% These intuitions will be more lucid in the discrete proofs than in the continuous case.
% In the latter, the set $\goodset$ will be dilated slightly to $\goodset_\COST$
% where the dilation shrinks with $\COST$;
% i.e. $\goodset_\COST \rightarrow \goodset$ as $\COST\rightarrow\infty$.
% This dilation appears due to a covering argument
% when extending some discrete proofs to continuous $\Xcal$.

\vspace{0.1in}

Before we proceed, we will prove a series of results that will be necessary in our proofs
of Theorems~\ref{thm:discreteThm} and~\ref{thm:compactThm}.
We first prove Lemma~\ref{lem:gtwoprob}.

\vspace{0.05in}

\noindent
\textbf{\textit{Proof of Lemma~\ref{lem:gtwoprob}.}}
Let $\Gtwo' = \Big\{\|\funcM\|_\infty \leq \zetamm{M-1}/2 \;\cap\;
                \bigcap_{m=1}^{M-1} \|\funcm\|_\infty \leq \zetam/2 \Big\}$.
It is straightforward to see that $\Gtwo'\subset\Gtwo$ since for any
$m\leq M-1$,
\[
\|\funcM-\funcm\|_\infty \leq \|\funcM\|_\infty + \|\funcm\|_\infty 
\leq \zetamm{M-1}/2 + \zetam/2 \leq \zetam.
\]
Hence,
$\PPgp(\Gtwo) \geq \PPgp(\Gtwo')$.
We can now bound,
\begin{align*}
\PPgp(\Gtwo')
 \,=\, \PPgp\big(\|\funcM\|_\infty \leq \zetamm{M-1}/2\big)
    \cdot \prod_{m=1}^{M-1} 
    \PPgp\big(\|\funcm\|_\infty \leq \zetam/2\big)
  \,\geq\, \Gtwoprob.
\end{align*}
Here the equality in the first step comes from the observation that the
$\funcm$'s are independent under the $\PPgp$ probability.
 The last inequality comes from
Assumption~\ref{asm:supgp}.
\hfill\BlackBox

\vspace{0.05in}
\begin{remark}
\emph{
% \toworkon{better way to phrase this?}
It is worth noting that the above bound
is a fairly conservative lower bound on {\normalfont $\Gtwoprob$}
since {\normalfont$\Gtwo'$} essentially requires that all samples $\funcm$ be small
so as to make the differences $\funcM-\funcm$  small.
% As a simple thought experiment, if we sample $Z_1,Z_2\sim\Ncal(0,1)$ and wish to satisfy
% a condition $|Z_1-Z_2| < 0.1$, the above bound is akin to requiring that both $Z_1$ and
% $Z_2$ are small, i.e. 
% $\PP(|Z_1-Z_2| < 0.1) > \PP(|Z_1|<0.05 \cap |Z_2|<0.05) = 0.0004$.
% However, $|Z_1-Z_2| < 0.1$ can occur even when $Z_1,Z_2$ are large.
% For example,
% by noting that $Z_1-Z_2\sim\Ncal(0,2)$ we have $\PP(|Z_1-Z_2|<\epsilon) =
% \Phi(\epsilon/\sqrt{2}) - \Phi(-\epsilon/\sqrt{2}) = 0.0564$.
% }
% 
% \emph{
We can obtain a more refined
bound on $\Gtwoprob$ by noting that
$\funcM-\funcm\sim\GP(\zero,2\kernel)$ and following proofs for bounding the supremum
of a GP (e.g. Theorem 5.4 in~\citet{adler1990gps},
Theorem 4 in~\citet{ghosal06gpconsistency}).
This leads to smaller values for $\betat$ in Theorems~\ref{thm:discreteThm}
and~\ref{thm:compactThm} and consequently better constants in our bounds.
However, this analysis
will require accounting for correlations when analysing multiple GPs which
is beyond the scope and tangential to the goals of this paper.
Moreover, from a practical perspective it would not result in anything actionable since
many quantities in the expression for $\betat$ are already unknown in practice,
even for \gpucb.
% However, since $\{\funcM-\funcm\}_{m=1}^{M-1}$ are not independent GPs anymore, it
% will require accounting for the dependencies, which are tangential to the main
% ideas in this paper.
It is also worth noting that the dependence of $\Gtwoprob$ on our regret bounds
is mild since it appears as a $\sqrt{\log(1/\Gtwoprob)}$ term.
}
\end{remark}

Next, Lemma~\ref{lem:GoGtbound} provides a way to bound the probability of an event
under our prior (\Gones and \Gtwo)
using the probability of the event when the functions are sampled from a GP (\Gones only).

\begin{lemma}
\label{lem:GoGtbound}
Let $E$ be a $\PPgp$-measurable event. Then,
$\PP(E) \leq \Gtwoprob^{-1}\PPgp(E)$.
\end{lemma}
\noindent
\textbf{\emph{Proof}}
This follows via a straightforward application of Bayes' rule, shown below.
The last step uses Lemma~\ref{lem:gtwoprob} and that the intersection of two sets is
at most as large as either set.
\[
\PP(E) = \PPgp(E|\Gtwo) = \frac{\PPgp(E\cap\Gtwo)}{\PPgp(\Gtwo)} \leq
\frac{1}{\Gtwoprob}\PPgp(E).
% \\[-0.4in]
\]
\hfill\BlackBox

\vspace{0.10in}

For our analysis, we will also need to control the 
sum of conditional standard deviations for queries in a
subset $A\subset \Xcal$.
We provide the lemma below, whose proof is based 
of a similar result in~\citet{srinivas10gpbandits}.
% \insertprespacing
\begin{lemma}
\label{lem:IGBoundLemma}
Let $\func\sim\GP(0,\kernel)$, $\func:\Xcal\rightarrow\RR$ and each time we query at
any $x\in\Xcal$ we observe $y = \func(x)+\epsilon$, where
$\epsilon\sim\Ncal(0,\eta^2)$. 
Let $A\subset\Xcal$. Assume that
we have queried $f$ at $n$ points, $(x_t)_{t=1}^n$ of which $s$ points are in $A$. 
Let $\sigmanontmo$ denote the posterior variance at time $t$, i.e. after $t-1$ queries.
Then, $\sum_{x_t\in A}\sigmanontmosq(x_t)\leq \frac{2}{\log(1+\eta^{-2})}\IGnn{s}(A)$.
\end{lemma}

% \insertprespacing

\noindent
\textbf{\emph{Proof}}
\newcommand{\xAss}[1]{z_{#1}}
\newcommand{\xAt}{\xAss{t}}
Let $A_s = \{\xAss{1},\xAss{2},\dots,\xAss{s}\}$ be the queries inside $A$ in the 
order they were queried.
Now, assuming that we have only queried inside $A$ at $A_s$, denote by 
$\sigmatnontmo(\cdot)$, the posterior standard deviation after $t-1$ such queries. Then,
\begin{align*}
\sum_{t:x_t\in A} \sigmanontmosq(x_t) \;&\leq\;
  \sum_{t=1}^s\sigmatnontmosq(\xAt) 
  \;\leq\; \sum_{t=1}^s \eta^2\frac{\sigmatnontmosq(\xAt)}{\eta^2}
  \;\leq\; \sum_{t=1}^s
    \frac{\log(1+\eta^{-2}\sigmatnontmosq(\xAt))}{\log(1+\eta^{-2})} \\
  \;&\leq\; \frac{2}{\log(1+\eta^{-2})} I(y_{A_s}; f_{A_s})
\end{align*}
Queries outside $A$ will only decrease the variance of the GP so we can upper bound
the first sum by the posterior variances of the GP with only the queries in $A$.
The third step uses the inequality $u^2/v^2 \leq \log(1+u^2)/\log(1+v^2)$ with 
$u = \sigmatnontmo(\xAt)/\eta$ and $v=1/\eta$ and the last step uses 
Lemma~\ref{lem:IGformula} in Appendix~\ref{app:ancillaryResults}. 
The result follows from the fact that $\IGnn{s}(A)$
maximises the mutual information among all subsets of size $s$.
\hfill\BlackBox
% \end{proof}

% \subsection{Discrete $\Xcal$}
\subsection{Discrete $\Xcal$
% Proof of Theorem~\ref{thm:discreteThm}
% (Discrete $\Xcal$)
}
\label{sec:proofsDiscrete}

\vspace{0.10in}
\noindent
\textbf{\emph{Proof of Theorem~\ref{thm:discreteThm}. }}
Without loss of generality,
we can assume that \mfgpucbs is run indefinitely.
% but will analyse the queries until capital $\COST$ is spent.
Let
$N$ denote the (random) number of queries within $\COST$, i.e. the quantity
satisfying $N = \max \{n\geq 1; \sum_{t=1}^n\costmt \leq \COST$\}.
Note that $\supp(N) \subset \{n\in \NN: \nCOSTbot \leq n \leq \nCOSTtop\}$.
In our analysis,
we will first analyse \mfgpucbs after $n$ steps and control the regret
and the number of lower fidelity evaluations. 

\vspace{0.1in}

\noindent
\textbf{Bounding the regret after $n$ evaluations:}
We will need the following lemma  to establish that $\ucbtx$ upper bounds $\funcM(x)$.
The proof is given in Section~\ref{sec:discreteUCBProof}.

\vspace{\thmprespacing}
\begin{lemma}
\label{lem:discreteUCB}
Pick $\delta\in(0,1)$ and choose $\betat \geq
2\log\left(\frac{M|\Xcal|\pi^2t^2}{3\Gtwoprob\delta}\right)$. 
Then, with probability at least $1-\delta/2$, for all $t\geq 1$, for all $x\in\Xcal$ and
for all $m\in\{1,\dots,M\}$, we have
% $|\funcm(x)-\mumtmo(x)| \leq\betath \sigmamtmo(x)$.
\[\big|\funcm(x)-\mumtmo(x)\big| \;\leq\; \betath \sigmamtmo(x).\]
% \[
% \PP\left(\forall t, \,\forall m =1,\dots,M, \,\forall x\in\Xcal:\;\;\;
%   |\funcm(x)-\mumtmo(x)| >\betath \sigmamtmo(x)\right) \leq \frac{\delta}{2}
% \]
\end{lemma}
\vspace{\thmprespacing}
% 
% We now have all the ingredients to bound the simple regret.
First note the following bound on the instantaneous regret when $\mt=M$,
\begin{align*}
\funcopt - \funcM(\xt) \;&\leq\; \ucbttxx{t}{\xopt} - 
  (\muMtmo(\xt) - \betath\sigmaMtmo(\xt))
\numberthis \label{eqn:discreteInstRegretBound}
\\
\;&\leq\; \ucbtxt - (\muMtmo(\xt) - \betath\sigmaMtmo(\xt)) 
  \;\leq\;\; 2\betath\sigmaMtmo(\xt).
\end{align*}
The first step uses that
$\ucbmmttxx{m}{t}{x}$ is an upper bound for $\funcM(x)$ by Lemma~\ref{lem:discreteUCB} and
the assumption \Gtwo, and hence so is the minimum $\ucbtx$.
The second step uses that $\xt$ was the maximiser of $\ucbtx$ and the third step that
$\ucbMtx \geq \ucbtx$.
To control $\Rtilden$, we will use $\goodset = \HcalM$ in~\eqref{eqn:Rtilden} and invoke
Lemma~\ref{lem:IGBoundLemma}.
Applying the Cauchy Schwarz inequality yields,
% $\big(\sum_{i=1}^s a_i\big)^2 \leq s \sum_{i=1}^s a_i^2$ yields,
\begingroup
\begin{align*}
\Rtilden^2 \;&\leq\; 
  \Tmmttxx{m}{n}{\HcalM}
  \sum_{\substack{t:\mt=M\\ \xt\in\HcalM}} \left(\funcopt-\funcM(\xt)\right)^2
  \;\leq\; 
  \Tmmttxx{m}{n}{\HcalM}
\sum_{\substack{t:\mt=M\\ \xt\in\HcalM}} 4\betat (\sigmamtmo(\xt))^2 \\
  \;&\leq\; 
  C_1\Tmmttxx{m}{n}{\HcalM}
\betan\IGnn{\Tmmttxx{m}{n}{\HcalM}}(\HcalM).
\numberthis \label{eqn:RtildenBound}
\end{align*}
\endgroup
Here $C_1 = 8/\log(1+\eta^{-2})$.
% This result will be the premise to establish both results in
% theorem~\ref{thm:discreteThm}.

\vspace{0.1in}

\noindent
\textbf{Bounding the number of evaluations:}
Lemma~\ref{lem:TmnxBounds}, given below, bounds the number of evaluations at
different fidelities in
different regions of $\Xcal$.
This will allow us to bound, among other things,
the total number of plays $N$ and the number of $M$\ssth 
fidelity evaluations outside $\goodset$.
% We will use the following lemma which bounds the number of queries to different
% points at different fidelities.
The proof of Lemma~\ref{lem:TmnxBounds} is given in Section~\ref{sec:TmnxBoundsProof}.
Recall that $\Tmnx$ denotes the number of queries at point $x\in\Xcal$ at fidelity 
$m$.
Similarly, we will denote $\Tgmnx$ to denote the number of queries at point $x$
at fidelities larger than $m$.

\begin{lemma}
\label{lem:TmnxBounds}
Pick $\delta\in(0,1)$ and set $\betat =
2\log\left(\frac{M|\Xcal|\pi^2t^2}{3\Gtwoprob\delta}\right)$. 
% Let $\rho\geq 2$.
Further assume $\ucbttxx{t}{\xopt} \geq \funcopt$.
Consider any $x\in\Hcalm\backslash\xoptset$ for $m<M$. 
We then have the following bounds on the number of queries at any given time step $n$,
\begingroup
\allowdisplaybreaks
\begin{align*}
\Tlnx \;\;&\leq\;\; \frac{\eta^2}{\gammamsq}\betan + 1, \hspace{0.2in} 
  \textrm{for }\; \ell<m, \\
\PP\left(\Tmnx \,>\, \bigceil{5\Big(\frac{\eta}{\Deltamx}\Big)^2\betan} \right)
  \;\;&\leq\;\; \frac{3\delta}{2\pi^2}\frac{1}{|\Xcal|n^2}, \\
\PP\left(\Tgmnx>u\right) \;\;&\leq\;\; \frac{3\delta}{2M\pi^2} 
  \frac{1}{|\Xcal|u}.
\end{align*}
\end{lemma}
\endgroup
% \vspace{0.1in}
% \vspace{\thmpostspacing}

First whenever $\ucbtxx{\xopt}\geq\funcopt$,  by using the union bound on the second
result of Lemma~\ref{lem:TmnxBounds},
\[
\PP\left(\; \exists\, n\geq 1,\; \exists\,m\in\{1,\dots,M\},\;
\exists\,x\in\Hcalm\backslash\xoptset,\;\;
  \Tmnx \,>\, \bigceil{5\Big(\frac{\eta}{\Deltamx}\Big)^2\betan} \right)
  \leq \frac{\delta}{4}.
\]
Here we have used $\sum n^{-2} = \pi^2/6$. The last two quantifiers just
enumerates over all $x\in\Xcal\backslash\xoptset$.
Similarly, applying the union bound for $u=1$
on the third result, we have, for any given $n$,
\begin{align*}
\PP\left(\; \exists\,m\in\{1,\dots,M\},\;
  \exists\,x\in\Hcalm,\;\;
  \Tgmnx \,>\, 1 \right)
  \leq \frac{3\delta}{2\pi^2}
  < \frac{\delta}{4}.
\end{align*}
We will apply the above result for $n = \lfloor \COST/\costone \rfloor$
% and $u=1$ in the third result of Lemma~\ref{lem:TmnxBounds}.
% Applying the union bound over all $x\in\Xcal$ and
and observe that $\Tgmnx$ is
non-decreasing in $n$.
Hence,
% $u = n^{\alpha}$ and $\rho = 1 + \sqrt{1/2+1/\alpha}$ as given in the theorem.
% Then, $u^{2(\rho-1)^2-1} = n^2$.
% Applying the union bound over $n\geq 1$ we have,
\begin{align*}
\PP\left(\; \forall\, n\leq \COST/\costone,\; \forall\,m\in\{1,\dots,M\},\;
  \forall\,x\in\Hcalm,\;\;
  \Tgmnx \,\leq\, 1 \right)
  > 1 - \frac{\delta}{4}.
\label{eqn:tgmnxbound} \numberthis
\end{align*}
The condition for Lemma~\ref{lem:TmnxBounds} holds with probability
at least
$1-\delta/2$ (by Lemma~\ref{lem:discreteUCB}), and therefore
the above bounds hold together with probability $>1-\delta$.
We have tabulated these bounds in Table~\ref{tb:TmnxBounds}.
We therefore have the following bound on the number of fidelity $m\;(<M)$
plays $\TmnXcal$,
\insertTmnxTable
\begingroup
\allowdisplaybreaks
\begin{align*}
\TmnXcal
\;&\leq\;
\Tmmttxx{m}{n}{\Hcaldownm}
\;+\; \sum_{x\in\Hcalm}\left\lceil\frac{5\eta^2}{\Deltamxsq}\betan\right\rceil
\;+\; |\Hcalupm|\left\lceil\frac{\eta^2}{\gammamsq}\betan \right\rceil
\\
\;&\leq\;
\Tmmttxx{m}{n}{\Hcaldownm}
\;+\; |\Hcalm \cup \Hcalupm|\left\lceil\frac{\eta^2}{\gammamsq}\betan\right\rceil
\numberthis \label{eqn:tmnxcalboundone}
\\
\;&\leq\;
|\Hcalmm{m-1}|
\;+\; |\Hcalm \cup \Hcalupm|\left\lceil\frac{\eta^2}{\gammamsq}\betan\right\rceil
\numberthis \label{eqn:tmnxcalboundtwo}
\end{align*}
\endgroup
The second step uses that $\Deltamx \geq 3\gammam$ for $x\in\Hcalm$ and
the last step uses the modification to the discrete algorithm which ensures that we
will always play an arm at a lower fidelity before we play it at a higher fidelity.
Hence, for an arm in $\Hcalm$, the $1$ play at fidelities larger than $m$ will be
played at fidelity $m+1$.
% In addition to this, note that we also have
% $\Tmmttxx{M}{n}{\Hcaldownmm} \leq 1
% Now we can control the number of evaluations at the approximations.
% Precisely, we will show that for sufficiently large $\COST$, $N\leq 2\nCOST$.
% For this we first observe that the number of plays at lower fi

\vspace{0.1in}

\vspace{0.10in}
\noindent
\textbf{Proof of first result:}
% We will study \mfgpucbs after $n$ evaluations, and invoke the high probability conditions
% from the proof above.
First consider the 
total cost $\COST'(n)$ expended at fidelities $1,\dots,M-1$ and at the
$M$\ssth fidelity outside of $\HcalM$ after $n$ evaluations.
Using~\eqref{eqn:tmnxcalboundtwo}, we have,
\begingroup
\allowdisplaybreaks
\begin{align*}
\COST'(n) \;&=\;
          \summMmo \costm \Tmmttxx{m}{n}{\Xcal} \;+\;
            \costM\Tmmttxx{M}{n}{\Hcaldownmm{M}} \\
  &\leq\;
\sum_{m=2}^M \costmm{m} |\Hcalmm{m-1}|\,+\,
 \sum_{m=1}^{M-1} \costm |\Hcalm \cup \Hcalupm|
\left\lceil\frac{\eta^2}{{\gammam}^2}\betan \right\rceil.
\end{align*}
\endgroup
Since $N\leq \nCOSTtop$, we have for all $n\in\supp(N)$, $\COST'(n)$ is less
than the LHS of~\eqref{eqn:costonecondn} and hence less than $\COST/2$.
Therefore, the amount of cost spent at the $M$\ssth fidelity inside $\HcalM$ is
at least $\COST/2$ and since each such evaluation expends $\costM$,
we have $\Tmmttxx{M}{N}{\HcalM} \geq \nCOST/2$.
Therefore using~\eqref{eqn:RtildenBound} we have,
\begin{align*}
S(\COST) \leq \frac{1}{\Tmmttxx{M}{N}{\HcalM}} \RtildeN
\leq \sqrt{\frac{C_1 \beta_N\IGnn{\Tmmttxx{M}{N}{\HcalM}}(\HcalM)}{\Tmmttxx{M}{N}{\HcalM}}}
\leq \sqrt{\frac{2C_1 \beta_{\nCOSTtop}\IGnn{{\nCOST}}(\HcalM)}{\nCOST}}.
\end{align*}
Here, we have used 
$N\leq \nCOSTtop$ and that $\nCOST \geq \Tmmttxx{M}{N}{\HcalM} \geq \nCOST/2$.

% Proof of Corollary ends here ===================================================

\vspace{0.1in}

\noindent
\textbf{Proof of second result:}
Using~\eqref{eqn:tmnxcalboundone},
the total number of 
% By summing~\eqref{eqn:tmnxcalbound} for $m< M$,
queries at fidelities less than $M$ and the number of $M$\ssth fidelity
queries outside of $\HcalM$ can be bounded as follows,
\begingroup
\allowdisplaybreaks
\begin{align*}
\sum_{m=1}^{M-1}\sum_{x\in\Xcal} \Tmnx
\;+\; \Tmmttxx{M}{n}{\Hcaldownmm{M}}
\;\leq\;
|\Xcal| \;+\; \summMmo
|\Hcalm \cup \Hcalupm|\left\lceil\frac{\eta^2}{\gammamsq}\betan\right\rceil.
% \sum_{x\in\Hcalm}\left(\frac{5\eta^2}{\Deltamxsq}\betan + 1\right) +
% \sum_{m=1}^{M-1} |\Hcalupm|\left(\frac{\eta^2}{\gammamsq}\betan + 1\right).
\numberthis
\label{eqn:lowerFidelPlaysBound}
\end{align*}
\endgroup
% For this we first note that the LHS of~\eqref{eqn:lowerFidelPlaysBound} is
% reducible to the following for all $m\leq M-1$,
% \begin{align*}
% \sum_{x\in\Xcal} \Tmnx
% \leq \sum_{x\in\Hcaldownm} \Tmnx +
% \sum_{x\in\Hcalm}\left(\frac{5\eta^2}{\Deltamxsq}\betan + 1\right) +
% \sum_{x\in\Hcalupm}\left(\frac{\eta^2}{\gammamsq}\betan + 1\right).
% \end{align*}
% Now summing the above from $m=1$ to $M-1$ and rearranging the $\Tgmnx$ terms
% as follows gives us the bound~\eqref{eqn:lowerFidelPlaysBound}.
The first term of the RHS above follows via~\eqref{eqn:tgmnxbound} and the
following argument.
In particular, this does not use the additional condition on the discrete
algorithm -- we will use a similar argument in the continuous domain setting.
\begin{align*}
&\sum_{m=1}^{M}\sum_{x\in\Hcaldownm}\Tmnx
=
\sum_{m=1}^{M}\sum_{\ell=1}^{m-1} \sum_{x\in\Hcall}\Tmnx
\leq\;\sum_{m=1}^{M-1}\sum_{x\in\Hcalm} \Tgmnx
\;\leq\; |\Xcal|.
\numberthis \label{eqn:ncb2TgmnxBound}
\end{align*}
% Therefore, the following upper bounds the number of
% queries at fidelities less than $M$ after $n$ time steps,
% \[
% |\Xcal|n^\alpha \;+\; \summMmo
% \sum_{x\in\Hcalm}\left(\frac{5\eta^2}{\Deltamxsq}\betan + 1\right) +
% \sum_{m=1}^{M-1} |\Hcalupm|\left(\frac{\eta^2}{\gammamsq}\betan + 1\right).
% \]
% We can further argue that the number of $M$\ssth fidelity evaluations outside
% of $\HcalM$
Let the LHS of~\eqref{eqn:lowerFidelPlaysBound} be $A$ and the RHS be $B$ when $n=N$.
When $\COST>\COSTtwo$, by~\eqref{eqn:costtwocondn} and using the fact that
$N\leq \nCOSTtop$, we have $B < \nCOSTbot/2 < N/2$.
Since $N=A + \Tmmttxx{M}{N}{\HcalM}$, we have $\Tmmttxx{M}{N}{\HcalM} > N/2 > \nCOSTbot/2$.
Further, 
since the total expended budget after $N$ rounds $\COST(N)$ satisfies $\COST(N) \geq
 \Tmmttxx{M}{N}{\HcalM} \costM > \costM N /2$, we also have
$N<2\nCOST$.
Putting these results together we have for all $\COST>\COSTtwo$,
\begin{align*}
S(\COST)
\leq \sqrt{\frac{C_1 \beta_{N}\IGnn{\Tmmttxx{M}{N}{\HcalM}}(\HcalM)}
  {\Tmmttxx{M}{N}{\HcalM}}}
\leq \sqrt{\frac{2C_1 \beta_{2\nCOSTbot}\IGnn{{\nCOST}}(\HcalM)}{\nCOST}}.
\end{align*}

\vspace{-0.20in}

\hfill\BlackBox

\begin{remark}
\label{rmk:discgammamchoice}
\textbf{Choice of $\gammam$:}
\emph{
As described in the main text, the optimal choice for $\gammam$ 
 depends on the available budget and unknown problem dependent quantities.
However the choice $\gammam = \sqrt{\costm/\costmm{m+1}}\zetam$ ensures that
for any $x\in\Hcalm$, the bounds on the number of plays in Table~\ref{tb:TmnxBounds}
are on the same order for fidelities $m$ and below.
To see this, consider any $\ell < m$.
Then, 
\[\Deltamx
= \Deltammxx{\ell}{x} + \zetal - \zetam + \funcmm{\ell}(x) - \funcM(x) + \funcM(x) -
\funcm(x) \leq 3\gammamm{\ell} + 2\zetal \leq 5 \zetal.
\]
We therefore have,
\[
\costl\cdot \frac{\eta^2}{\gammammsq{\ell}} = 
\costmm{\ell+1}\frac{\eta^2}{{\zetal}^2} \leq
5 \bigg(\costm \cdot \frac{ 5 \eta^2}{\Deltamxsq}  \bigg)
\]
Above, by Table~\ref{tb:TmnxBounds},
the left most expression is an upper bound on the cost spent at fidelity
$\ell$ and the term inside the parantheses is an upper bound on the cost spent
at fidelity $m$.
Hence, the capital spent at the lower fidelities is within a constant factor of this
bound.
In the $K$-armed setting~\citep{kandasamy2016mfbandit}, we showed a
$\bigO(\eta^2/\Deltamxsq)$ lower bound on the number of plays at the $m$\ssth
fidelity as well;
such a result is not straightforward in the GP setting due to correlations between arms.
%\toworkon{@Gautam: please check wording.}
}
\end{remark}

\subsubsection{Proof of Lemma~\ref{lem:discreteUCB}}
\label{sec:discreteUCBProof}

This is a straightforward argument using Gaussian concentration and the union bound.
% But we would like to emphasise some subtle conditioning arguments with multiple
% fidelities.
Consider any given $m, t, x$.
\begingroup
\allowdisplaybreaks
\begin{align*}
& \PP\left(|\funcm(x)-\mumtmo(x)|>\betath\sigmamtmo(x)\right) \\
& \hspace{0.3in}
=\frac{1}{\Gtwoprob}\PPgp\left(|\funcm(x)-\mumtmo(x)|>\betath\sigmamtmo(x)\right) \\
& \hspace{0.3in}
=\frac{1}{\Gtwoprob}\EEgp\left[\EEgp\left[ \indfone\Big\{|\funcm(x)-\mumtmo(x)|>\betath\sigmamtmo(x)\Big\}
  \;\Big|\; \Dcalmtmo  \right]\right] \\
& \hspace{0.3in}
=\frac{1}{\Gtwoprob}\EEgp\left[\EEgp\left[ \indfone\Big\{|\funcm(x)-\mumtmo(x)|>\betath\sigmamtmo(x)\Big\}
  \;\Big|\; \Dcalmtmo  \right]\right] \\
& \hspace{0.3in}
=\frac{1}{\Gtwoprob}\EEgp\left[ \PP_{Z\sim\Ncal(0,1)}\left(|Z|>\betath\right) \right] \;
% & \hspace{0.3in}
\leq\; \frac{1}{\Gtwoprob}\exp\Big(\frac{\betat}{2}\Big)
\;=\; \frac{3\delta}{M|\Xcal|\pi^2t^2}.
\end{align*}
\endgroup
The first step uses Lemma~\ref{lem:GoGtbound}.
In the second step we have conditioned w.r.t $\Dcalmtmo$ which allows us to use 
Lemma~\ref{lem:gaussConcentration}. 
Recall that conditioning on all queries
will not be a Gaussian due to the $\zetam$ constraints.
% We will be using this conditioning argument repeatedly in our analysis.
The statement follows via a union bound over all $m\in\{1,\dots,M\}$, $x\in\Xcal$ and
all $t$ and noting that $\sum_{t}t^{-2} = \pi^2/6$.
\hfill\BlackBox

\subsubsection{Proof of Lemma~\ref{lem:TmnxBounds}}
\label{sec:TmnxBoundsProof}

First consider any $\ell<m$. Assume that we have already queried
$\big\lceil\eta^2\betan/\gammamsq\big\rceil$ times at any $t\leq n$. Since
the Gaussian variance after $s$ observations is
$\eta^2/s$ and that queries elsewhere will only decrease the conditional variance we
have,
$
\kernelmmtt{\ell}{t-1}(x,x) \leq
\eta^2/\Tmmttxx{\ell}{t-1}{x} 
< \gammamsq/\betan.
% \frac{\eta^2}{\Tmmttxx{\ell}{t-1}{x}} 
% < \frac{\gammamsq}{\betan}.
$
Therefore, $\betath\sigmammtt{\ell}{t-1}(x) < \betanh \sigmammtt{\ell}{t-1}(x)
< \gammam$ and by the design of our algorithm we will not play at the
$\ell$\ssth fidelity at time $t$ for all $t$ until $n$.
This establishes the first result.

To bound $\Tmnx$ we first observe, 
\begingroup
\allowdisplaybreaks
\begin{align*}
\indfonearg{\Tmnx > u} &\;\leq\;\;
\indfonearg{\exists t: u+1\leq t \leq n:\; \ucbttxx{t}{x} \textrm{ was maximum} 
  \indAnds \\
  & \hspace{0.55in}
  \betath\sigmammtt{\ell}{t-1}(x) < \gammam,\,\forall\,\ell<m \indAnd\,
  \betath\sigmamtmo(x) \geq \gammam \indAnd\, \\
  & \hspace{0.55in} \Tmmttxx{m}{t-1}{x} \geq u } \\
  &\;\leq\;\;
  \indfonearg{\exists t: u+1\leq t \leq n:\; \ucbttxx{t}{x} > \ucbttxx{t}{\xopt} 
      \indAnds \Tmmttxx{m}{t-1}{x} \geq u  } \\
  &\;\leq\;\;
  \indfonearg{\exists t: u+1\leq t \leq n:\; \ucbmtx > \funcopt 
      \indAnds \Tmmttxx{m}{t-1}{x} \geq u  }. 
\numberthis\label{eqn:TmnxguConditions}
\end{align*}
\endgroup
The first line just enumerates the conditions in our algorithm for it to have played
$x$ at time $t$ at fidelity $m$. In the second step we have relaxed some of those
conditions, noting in particular that if $\ucbttxx{t}{\cdot}$ was maximised at $x$
then it must be larger than $\ucbttxx{t}{\xopt}$. The last step uses the fact that
$\ucbmtx \geq \ucbtx$ and the assumption on $\ucbttxx{t}{\xopt}$.
Consider the event $\{\ucbmtx > \funcopt \indAnd \Tmmttxx{m}{t-1}{x} \geq u\}$.
We will choose $u=\ceil{5\eta^2\betan/\Deltamxsq}$ and
bound its probability via,
\begingroup
\allowdisplaybreaks
\begin{align*}
&\PP\Big(\ucbmtx > \funcopt \indAnd \Tmmttxx{m}{t-1}{x} \geq u\Big) \\
&\hspace{0.5in}=\;
 \frac{1}{\Gtwoprob}
  \PPgp\Big(\mumtmo(x) + \betath\sigmamtmo(x) + \zetam > \funcopt \indAnds 
      \Tmtmox\geq u\Big) \\
&\hspace{0.5in}=\;
 \frac{1}{\Gtwoprob}
  \PPgp\bigg(\mumtmo(x) - \funcm(x) \,>\,
  \underbrace{\funcopt-\funcm(x) -\zetam}_{\Deltamx} -\betath\sigmamtmo(x)
  \,\indAnd\,\\
&\hspace{1.6in}
 \Tmtmox > u\bigg) \\
&\hspace{0.5in}\leq\;
 \frac{1}{\Gtwoprob}
  \PPgp\Big(\mumtmo(x) - \funcm(x) >\, (\sqrt{5}-1) \betanh \sigmamtmo(x)\Big) \\
&\hspace{0.5in}\leq\;
 \frac{1}{\Gtwoprob}
  \PP_{Z\sim\Ncal(0,1)}\Big(Z\,> \frac{(\sqrt{5}-1)^2}{2}\betanh \Big) \\
%   \,\leq\,
&\hspace{0.5in}\leq\;
 \frac{1}{\Gtwoprob}
  \frac{1}{2}\exp\left(-\frac{3}{4}\betan\right) 
% &\hspace{0.5in}=\;
  \;=\;
 \frac{1}{\Gtwoprob}
  \frac{1}{2}\left(\frac{3\Gtwoprob\delta}{M|\Xcal|\pi^2}\right)^{\frac{3}{2}} n^{-3}
  \,\leq\,\frac{1}{2}\frac{3\delta}{M|\Xcal|\pi^2} n^{-3}
\end{align*}
\endgroup
Above in the third step we have used, if
$u\geq5\eta^2\betan/\Deltamxsq$, then $\Deltamx \geq \sqrt{5}\betanh\sigmamtmo(x)$ 
and that  $\betan\geq\betat$. The fourth step uses Lemma~\ref{lem:gaussConcentration} 
after
conditioning on $\Dcalmtmo$, the fifth step uses $(\sqrt{5}-1)^2>3/2$ and the last step
uses  $3\delta/|\Xcal|\pi^2 < 1$.
Using the union bound on~\eqref{eqn:TmnxguConditions}, we get $\PP(\Tmnx>u)
\leq \sum_{t=u+1}^n \PP(\ucbmtx > \funcopt \indAnd \Tmmttxx{m}{t-1}{x} \geq u)$.
Now~\eqref{eqn:TmnxguConditions} implies that
$\PP(\Tmnx>u) \leq \sum_{t=u+1}^n \PP(\ucbmtx > \funcopt \indAnd \Tmmttxx{m}{t-1}{x} \geq
u)$. The
second inequality of the lemma follows by noting that there are at most $n$ terms
in the summation.

Finally, for the third inequality we observe
\begin{align*}
\PP(\Tgmnx>u) \,\leq\, \PP\big( \exists t: u+1\leq t\leq n\,;\;
  \ucbmtx>\funcopt \indAnds \betath\sigmamtmo(x)<\gammam \big).
\numberthis\label{eqn:TgmnxguConditions}
\end{align*}
As before, we have used that if $x$ is to be queried at time $t$, then $\ucbtx$
should be at least larger than $\ucbttxx{t}{\xopt}$ which is larger than $\funcopt$
due to the assumption in the theorem. The second condition is necessary to ensure
that the switching procedure proceeds beyond the $m$\ssth fidelity. It is also
necessary to have $\betath\sigmammtt{\ell}{t-1}(x) <\gammal$ for $\ell<m$, 
but we have relaxed them.
We first bound the probability of the event
$\{\ucbmtx>\funcopt \indAnds \betath\sigmamtmo(x)<\gammam\}$.
\begingroup
\allowdisplaybreaks
\begin{align*}
&\PP\big(\ucbmtx\,>\,\funcopt \indAnd \betath\sigmamtmo(x)<\gammam\big) \\
&\hspace{0.5in}=\; 
\frac{1}{\Gtwoprob}
\PPgp\big(\ucbmtx\,>\,\funcopt \indAnd \betath\sigmamtmo(x)<\gammam\big) \\
&\hspace{0.5in}=\; 
\frac{1}{\Gtwoprob}
  \PPgp\big(\mumtmo(x)-\funcm(x)\,>\,\Deltamx - \betath\sigmamtmo(x)
 \indAnd \betath\sigmamtmo(x)<\gammam \big) \\
&\hspace{0.5in}\leq\; 
\frac{1}{\Gtwoprob}
  \PPgp\big(\mumtmo(x)-\funcm(x)\,>\, 2\gammam  - \betath\sigmamtmo(x)
 \indAnds \betath\sigmamtmo(x)<\gammam \big)    \\
&\hspace{0.5in}\leq\; 
\frac{1}{\Gtwoprob}
  \PPgp\big(\mumtmo(x)-\funcm(x)\,>\, \betath\sigmamtmo(x) \big)   \\
&\hspace{0.5in}\leq\; 
\frac{1}{\Gtwoprob}
  \PP_{Z\sim\Ncal(0,1)}\Big(Z\,> \betath \Big)
\;\leq\;
% &\hspace{0.5in}\leq\;
\frac{1}{\Gtwoprob}
   \frac{1}{2}\exp\left(-\frac{1}{2}\betat\right) \\
&\hspace{0.5in}=\;
\frac{1}{\Gtwoprob}
%   \,=\,
  \frac{1}{2}\left(\frac{3\Gtwoprob\delta}{M|\Xcal|\pi^2}\right)
t^{-2}
  \,\leq\,\frac{1}{2}\frac{3\delta}{M|\Xcal|\pi^2} t^{-2}
\end{align*}
\endgroup
Here, the second step uses that for all $x\in\Hcalm$, $\Deltamx>3\gammam>2\gammam$
and the  third step uses the second condition.
Using the union bound on~\eqref{eqn:TgmnxguConditions} and bounding the sum by an
integral gives us,
\begin{align*}
\PP(\Tgmnx>u) \;&\leq\; \sum_{t=u+1}^n
  \frac{1}{2}\frac{3\delta}{M|\Xcal|\pi^2} t^{-2}
\;\leq\;
  \frac{1}{2}\frac{3\delta}{M|\Xcal|\pi^2} \int_u^\infty t^{-2} \ud t 
  \\
  \;&\leq\; 
  \frac{1}{2}\;\frac{3\delta}{M|\Xcal|\pi^2}\; \frac{1}{u}.\\[-0.4in]
\end{align*}

\vspace{-0.1in}

\hfill\BlackBox

% \vspace{0.1in}
\subsection{Compact and Convex $\Xcal$}
\label{app:analysisCompact}

% \textbf{\emph{Proof of Theorem~\ref{thm:compactThm}. }}
To prove theorem~\ref{thm:compactThm} we will require a fairly delicate set up for
the continuous setting.
Given a sequence $\{\nun\}_{n\geq 0}$,
at time $n$ we will consider a
% $\frac{r\sqrt{d}}{2\nun^{\frac{1}{2d}}}$-covering of the space $\Xcal$ of size
$r\sqrt{d}/(2\nun^{1/2d})$-covering of the space $\Xcal$ of size
% $\frac{r\sqrt{d}}{2\nun^{1/2d}}$-covering of the space $\Xcal$ of size
$\nun^{1/2}$.
For instance, if $\Xcal = [0,r]^d$ a sufficient
discretisation would be an equally spaced grid having $\nun^{1/2d}$ points per
side.
Let $\{\ain\}_{i=1}^{\nalphatwo}$ be the points in the covering, 
$\Fn = \{\Ain\}_{i=1}^{\nalphatwo}$ be the ``cells'' in the covering, 
i.e. $\Ain$ is the set
of points which are closest to $\ain$ in $\Xcal$ and the union of 
all sets $\Ain$ in $\Fn$ is $\Xcal$.
Next we will define another partitioning of the space similar using this covering.
First let $\Fonen = \{ \Ain \in \Fn: \; \Ain \subset \Jcalmmee{1}{\maxtrg} \}$.
Next,
\begin{align*}
\Fmn = \left\{\,\Ain\in\Fn: \,\Ain\subset \Jcalcmtrg \,\indAnd\,
\Ain\notin \bigcup_{\ell=1}^{m-1} \Fln  \,\right\} 
\hspace{0.1in} \textrm{for } \;2\leq m\leq M-1.
\hspace{0.1in}
\label{eqn:FmnDefn}
\numberthis
\end{align*}
Note that $\Fmn\subset \Fmn$.
We define the following \emph{disjoint} subsets 
$\{\Fcalmn\}_{m=1}^{M-1}$ of $\Xcal$ via
$\Fcalmn = \bigcup_{\Ain\in\Fmn} \Ain$.
We have illustrated $\bigcup_{\ell=1}^{m-1}\Fcalln$ with respect to
$\Htcalm$ and $\Htncalm$ in Figure~\ref{fig:FcalHcal}.
% By noting that $\Htncalone = \Hcalone$ we make the following observation,
By observing that $\Htncalone = \Hcalone$ and 
that $\complement{\Htncalm \cup \Hcalupm} \subset \bigcup_{\ell=1}^{m-1}\Fcalln$
(see Figure~\ref{fig:FcalHcal})
we have the following,
\begin{align*}
\hspace{-0.15in}
\forall m\in\{1,\dots,M\},\quad\;
\TmnXcal \leq\,
\left(\sum_{\ell=1}^{m-1}\Tmnxx{\Fcalln}\right) 
\;+\, \Tmnxx{\Htncalm} \,+\, \Tmnxx{\Hcalupm}.
\numberthis
\label{eqn:TmnBreakdown}
\end{align*}
% This follows by noting $\complement{\Htncalm \cup \Hcalupm} \subset
% \bigcup_{\ell=1}^{m-1}\Fcalln$ (See Figure~\ref{fig:FcalHcal}). 
We are now ready to prove Theorem~\ref{thm:compactThm}.
We will denote the $\epsl$ covering number of a set $A\subset \Xcal$ in the $\|\cdot\|_2$ 
metric by $\Omegaii{\epsl}(A)$.
\insertFigFcalHcal

\vspace{0.1in}
\noindent
\textbf{\emph{Proof of Theorem~\ref{thm:compactThm}. }}
% To control $\Rtilden$ we will bound 
% control each of these terms individually.
As in the discrete case, we will first control the regret and the
number of lower fidelity
evaluations by controlling each term in~\eqref{eqn:TmnBreakdown}.

\vspace{0.1in}
  
\noindent
\textbf{Bounding the regret after $n$ evaluations:}
We will need the following lemma whose proof
is given in Section~\ref{sec:discreteUCBProof}.

\begin{lemma}
\label{lem:TmnHmBound}
For $\betat$ as given in Theorem~\ref{thm:compactThm}, the following holds with
probability $>1-5\delta/6$.
\[
\forall\,m\in\{1,\dots,M\},\;\;\forall\,t\geq 1,\hspace{0.2in}
\Deltamxt = \funcopt - \funcm(\xt) \,\leq\, 2\betat \sigmamtmo(\xt) \,+\, 1/t^2. %\frac{1}{t^2}.
\]
\end{lemma}

As in the discrete setting, we set
$\Zcal=\HtncalM$ in~\eqref{eqn:Rtilden} to bound $\Rtilden$.
Using $m=M$ in Lemma~\ref{lem:TmnHmBound} and using calculations similar to
the discrete case yields,
% and noting that when $\mt=M$, $\rt=\Deltammxx{M}{\xt}$.  Using 
\begin{align*}
\Rtilden \leq
\sum_{\substack{\mt=M\\\xt\in\goodset}}\left(2\betath\sigmaMtmo(\xt) + 
\frac{1}{t^2} \right)
\leq \sqrt{C_1\Tmnxx{\HtncalM}\betan\IGnn{\Tmnxx{\Htncalm}}(\Htncalm)} +
\frac{\pi^2}{6}.
\numberthis\label{eqn:Rtildenctsbound}
\end{align*}
Here $C_1 = 8/\log(1+\eta^{-2})$.
We have also used the fact $\sum_{t>0}t^{-2} = \frac{\pi^2}{6}$.

\vspace{0.1in}

\noindent
\textbf{Bounding the number of evaluations:}
The following lemma will be used to
 bound the number of plays in $\Htncalm \cup \Hcalupm$.
The proof is given in Section~\ref{sec:compactVarBound}.

\begin{lemma}
\label{lem:compactVarBound}
Let $\func\sim\GP(\zero,\kernel)$, $\func:\Xcal\rightarrow\RR$ and we observe
$y=f(x) + \epsilon$ where $\epsilon\sim\Ncal(0,\eta^2)$. Let
$A\subset\Xcal$ such that its $L_2$ diameter $\diam(A)\leq D$. Say we have $n$
queries $(\xt)_{t=1}^n$ of which $s$ points are in $A$. Then the posterior variance
of the GP, $\kernel'(x,x)$ at any $x\in A$ satisfies
\begin{align*}
\kernel'(x,x) \leq \begin{cases}
\;\;\Cse D^2 + \frac{\eta^2}{s}\hspace{0.3in} & \textrm{if $\kernel$ is the SE kernel,} \\
\;\;\Cmat D + \frac{\eta^2}{s} & \textrm{if $\kernel$ is the \matern kernel,}
\end{cases}
\end{align*}
for appropriate kernel dependent constants $\Cse,\Cmat$.
\end{lemma}

First consider the SE kernel.
At time $t$ consider any $\epsln =
\frac{\gammam}{\sqrt{8\Cse\betan}}$ covering $(B_i)_{i=1}^{\epsln}$ of
$\Htncalm\cup\Hcalupm$.
The number of queries inside any $B_i$ of this covering at time $n$ will be at most
$\left\lceil\frac{2\eta^2}{\gammamsq}\betan\right\rceil$.
To see this, assume we have already queried
this many times inside $B_i$ at time $t\leq n$. 
By Lemma~\ref{lem:compactVarBound}
the maximum variance in $A_i$ can be bounded by 
\[
\max_{x\in A_i} \kernelmmtt{m}{t-1}(x,x) \leq
\Cse(2\epsln)^2 \;+\; \frac{\eta^2}{\Tmmttxx{m}{t}{A_i}}
\leq \frac{\gammamsq}{\betan}.
\]
Therefore, $\betath\sigmamtmo(x) \leq \betanh\sigmamtmo(x) < \gammam$ and we 
will not query inside $A_i$ until time $n$. 
A similar result is obtained
for the \matern kernel by setting $\epsln =
\frac{\gammamsq}{4\Cmat\betan}$.
Therefore we have,
\begin{align*}
\Tmnxx{\Htncalm\cup\Hcalupm} \;&\leq\;
\Omegaen(\Hcalupm\cup\Hcalupm)\left\lceil\frac{2\eta^2}{\gammamsq}\betan\right\rceil
\label{eqn:htncalmbound}\numberthis
\\
\;&\leq\;
C_\kernel\eta^2 \betan^{p + 1}\frac{\vol(\Htncalm\cup\Hcalupm)}{{\gammam}^{2p}}.
\end{align*}
Here 
$C_\kernel = 2^{2+d/2}(d\Cse)^{\frac{d}{2}}$ and $p=1/2$ for the SE kernel while
$C_\kernel = 2^{2+d}(\Cmat)^{d}d^{d/2}$ and $p=1$ for the \matern{} kernel.
We have also used the fact that $\lceil k\rceil \leq 2k$ for large enough $k$
and the following bound for a $\delta$-packing in the Euclidean metric
$\Omega_\delta(A) \leq \vol(A)d^{d/2}/(2^{d/2} \delta^d)$.

% Therefore, the number of $m$\ssth fidelity queries is
% bounded by
% $\Omegaen(\Hcalupm)\left\lceil\frac{2\eta^2}{\gammamsq}\betan\right\rceil$.

Next, we will bound $\Tmnxx{\complement{\Htncalm\cup\Hcalupm}}$ by controlling. 
$\Tgmnxx{\Fcalmn}$.
To that end we provide the following Lemma whose
proof is given in Section~\ref{sec:TgmDmnBound}.
% \insertprespacing

\begin{lemma}
Consider any $\Ain \in \Fmn$ where $\Fmn$ is as defined in~\eqref{eqn:FmnDefn}.
Let $\betat$ be as given in Theorem~\ref{thm:compactThm}.
Then for all $n'\geq u\geq (3\eta)^{-2/3}$ we have,
\[
\PP(\TgmnpAin > u) \;\leq\; \frac{\delta}{\pi^2}\cdot \frac{1}{u}
\]
\label{lem:TgmDmnBound}
\end{lemma}
\insertpostspacing
Using the above result with $n' = \nCOSTtop$ gives us the result for all
$n'\leq\nCOSTtop$ since $\TgmnpAin$ is nondecreasing with $n$.
Setting $u=\max\{(3\eta)^{-2/3}, \nun^{1/2}\}$,
and applying the union bound
over all $m\in\{1,\dots,M\}$ and $\Ain\in\Fmn$, yields the following bound for all
$n'\leq\nCOSTtop$,
\begingroup
\allowdisplaybreaks
\begin{align*}
&\PP\left(\exists\,m\in\{1,\dots,M\},\, \TgmnpFcalmn > |\Fmn|\nun^{1/2}\right) %\\
\,\leq\,
% &\hspace{0.2in}\leq\;
\sum_{m=1}^M \PP\left( \TgmnpFcalmn > |\Fmn|\nun^{1/2} \right) \\%[-0.071in]
&\hspace{0.2in}\leq\;
\sum_{m=1}^M \sum_{\Ain\in\Fmn}\PP\left( \TgmnpAin > \nun^{1/2} \right) %\\
% &\hspace{0.2in}\leq\;
\;\leq\;
\sum_{m=1}^M |\Fmn| \frac{\delta}{\pi^2}\frac{1}{\nun^{1/2}}  \\
% \;\leq\;
&\hspace{0.2in}\leq\;
|\Fn|\frac{\delta}{\pi^2}\frac{1}{\nun^{1/2}} 
\;=\;
\frac{\delta}{\pi^2}
\leq
\frac{\delta}{6}
.%\frac{1}{\nun^{1\2}}.
\numberthis\label{eqn:Fcalmnbound}
\end{align*}
\endgroup
% Applying the union bound over $n$, we have $\TgmnFcalmn\leq n^\alpha$ for all $m$ 
% and all  $n\geq\max\{3, \,(2(\rho-\rho_0)\eta)^{2/3}\}^{2/\alpha}$ with probability
% $>1-\delta/6$.
% Therefore, using

Henceforth, all statements we make will make use of the bounds above and
will hold with probability $>1-\delta$ for all $n\in\supp(N)$.

\vspace{0.10in}
\noindent
\textbf{Proof of first result:}
Consider the cost $\COST'(n)$ spent at fidelities $1,\dots,M-1$ and at the
$M$\ssth fidelity outside of $\HtncalM$ after $n$ evaluations.
\begingroup
\allowdisplaybreaks
\begin{align*}
\COST'(n) \;&=\;
          \summMmo \costm \Tmmttxx{m}{n}{\Xcal} \;+\;
            \costM\Tmmttxx{M}{n}{\HtncalM} \\
 \;&=\;
    \sum_{m=1}^M \costm \left(\sum_{\ell=1}^{m-1}\Tmnxx{\Fcalln}\right) \,+\,
    \sum_{m=1}^{M-1} \costm \Tmmttxx{m}{n}{\Htncalm \cup \Hcalupm} \\
 \;&\leq\;
    \costM \nun + 
    C_\kernel\eta^2\betan^{p+1}\sum_{m=1}^{M-1} \costm
                      \frac{\vol(\Htncalm \cup \Hcalupm)}{{\gammam}^{2p}}
%                       \left\lceil\frac{\eta^2}{{\gammam}^2}\betan\right\rceil
% \\
%  \;&\leq\;
%     \sum_{m=2}^M \costmm{m} |\Hcalmm{m-1}|\,+\,
%     \eta^2\betan \sum_{m=1}^{M-1} \costmm{m+1} \frac{|\Hcalm \cup \Hcalupm|}{{\zetam}^2}
% %                       \frac{\eta^2}{{\gammam}^2}\betan
\end{align*}
\endgroup
The second step uses~\eqref{eqn:TmnBreakdown}.
The third step uses~\eqref{eqn:htncalmbound},~\eqref{eqn:Fcalmnbound},
 and the following argument,
\begin{align*}
\sum_{m=1}^M\left(\sum_{\ell=1}^{m-1}\Tmnxx{\Fcalln}\right)
\leq 
\summMmo\Tgmnxx{\Fcalln}
 \,\leq\, \summMmo |\Fmn| \nun^{1/2} \,\leq\, \nun^{1/2}|F_n| \,\leq\, \nun.
\label{eqn:lowfidelctsbound}\numberthis
\end{align*}
The remainder of the proof follows similar to the discrete case.
Noting that $\nCOSTbot \leq n \leq \nCOSTtop$ and that $\Htncalm$ is shrinking with $n$,
we can conclude that $\COST'(n)$ is less than the LHS of~\eqref{eqn:costtwocts}.
Therefore, $\Tmmttxx{M}{N}{\HtncalM} \geq \nCOST/2$ and hence,
\begin{align*}
S(\COST) \leq
\sqrt{\frac{C_1\beta_N\IGnn{\Tmmttxx{M}{N}{\Htncalm}}(\Htncalm)}%
        {\Tmmttxx{M}{N}{\HtncalM}}} +
\frac{\pi^2}{6\Tmmttxx{M}{N}{\HtncalM}}
\;\leq\;
\sqrt{\frac{2C_1\beta_{\nCOSTtop}\IGnn{\nCOST}(\Hcal_{\nCOST}^{(M)})}%
        {\nCOST}} +
\frac{\pi^2}{3\nCOST}
.
\end{align*}

\vspace{0.20in}
\noindent
\textbf{Proof of second result:}
As in the discrete case, we
bound the number of queries at fidelity $m<M$ and the $M$\ssth fidelity queries
outside $\HtncalM\cup \HcalM$ as follows.
\begin{align*}
\sum_{m=1}^{M-1}\TmnXcal \;+\; \Tmmttxx{M}{n}{\complement{\HtncalM}}
\;&\leq\;
\sum_{m=1}^M\left(\sum_{\ell=1}^{m-1}\Tmnxx{\Fcalln}\right)
\;+\, \summMmo\Tmnxx{\Htncalm\cup\Hcalupm} \\
% &\hspace{0.2in}\leq
&\leq\;
\nun \;+\;
C_\kernel\betan^{p+1}\summMmo \frac{\vol(\HtncalM\cup \HcalM)}{{\gammam}^{2p}}
\numberthis\label{eqn:ctslowfidelplays}
\end{align*}
The first step uses~\eqref{eqn:TmnBreakdown} while the
second step uses~\eqref{eqn:htncalmbound} and~\eqref{eqn:lowfidelctsbound}.
%  and the following argument,
% \[
% \sum_{m=1}^M\left(\sum_{\ell=1}^{m-1}\Tmnxx{\Fcalln}\right)
% \leq 
% \summMmo\Tgmnxx{\Fcalln}
%  \,\leq\, \summMmo |\Fmn| \nun^{1/2} \,\leq\, \nun^{1/2}|F_n| \,\leq\, \nun.
% \]
% The remainder of the proof follows similar to the discrete case.
Once again, similar to the discrete case we can argue
that for all $\COST>\COSTtwo$, the RHS $B$
of~\eqref{eqn:ctslowfidelplays} satisfies $B< \nCOST/2 < N/2$,
the $M$\ssth fidelity plays in $\HtncalM$ 
satisfies $\Tmmttxx{M}{N}{\HtncalM} > N/2 > \nCOST/2$,
and the number of plays satisfies $N\leq 2\nCOSTbot$.
Combining this with~\eqref{eqn:Rtildenctsbound} gives us the following for
all $n\leq \nCOST$,
% and the expended budget satisfies $\COST(N) \geq \costM
\begin{align*}
S(\COST) \leq
\sqrt{\frac{C_1\beta_N\IGnn{\Tmmttxx{M}{N}{\Htncalm}}(\Htncalm)}%
        {\Tmmttxx{M}{N}{\HtncalM}}} +
\frac{\pi^2}{6\Tmmttxx{M}{N}{\HtncalM}}
\;\leq\;
\sqrt{\frac{2C_1\beta_{2\nCOST}\IGnn{\nCOST}(\Hcal_{\nCOST}^{(M)})}%
        {\nCOST}} +
\frac{\pi^2}{3\nCOST}
.
\end{align*}

\vspace{-0.2in}

\hfill\BlackBox
% \end{proof}

% \insertpostspacing

\subsubsection{Proof of Lemma~\ref{lem:TmnHmBound}}
\label{sec:TmnHmBound}

The first part of the proof mimics the arguments in Lemmas 5.6, 5.7
of~\citet{srinivas10gpbandits}.
By Assumption~\ref{asm:kernelAssumption} for any given $m\in\{1,\dots,M\}$
and $i\in\{1,\dots,d\}$ we have,
\[
\PPgp\Bigg(\Big|\partialfrac{x_i}{\funcm(x)}\Big| \,>\, 
b\sqrt{\log\Big(\frac{6Mad}{\Gtwoprob\delta}\Big)}\;\Bigg)
\leq \frac{\Gtwoprob\delta}{6Md}
\]
Then, by the union bound and Lemma~\ref{lem:GoGtbound} we have,
\begin{align*}
&\PP\bigg( \forall\,m\in\{1,\dots,M\},\, \forall\,i\in\{1,\dots,d\},\,\forall
x\in\Xcal,\;\; \Big|\partialfrac{x_i}{\funcm(x)}\Big| \,<\, 
b\sqrt{\log\Big(\frac{6Mad}{\Gtwoprob\delta}}\Big)
\bigg) \\
&\hspace{0.2in}
\;\geq\; 1-\frac{\delta}{6}. 
% \numberthis \label{eqn:LoneLipschitzness}
\end{align*}
Now we construct a discretisation $\Ft$ of $\Xcal$ of size $(\nu_t)^d$ such
that we have for all $x\in\Xcal$, $\|x-[x]_t\|_1\leq rd/\nu_t$. Here $[x]_t$ is the
closest point to $x$ in the discretisation.
(Note that this is different from the discretisation appearing in
Theorem~\ref{thm:compactThm} even though we have used the same notation).
 By choosing $\nu_t = t^2brd\sqrt{\log(6Mad/(\Gtwoprob\delta))}$
and using the above we have
\begin{align}
\forall\,x\in\Xcal,\quad
|\funcm(x) - \funcm([x]_t)| \;\leq\; 
b\log(6Mad/\delta)\|x-[x]_t\|_1
 \;\leq\; 1/t^2 %\frac{1}{t^2}.
\label{eqn:TmnHmBoundIOne}
\end{align}
for all $\funcm$'s with probability $>1-\delta/6$.

Noting that $\betat \geq 2\log(M|\Ft|\pi^2t^2/2\delta)$ for the given choice of 
$\nu_t$ we have the following with probability $>1-\delta/3$.
\begin{align}
\forall\,t\geq1,\,\; \forall\,m\in\{1,\dots,M\},\; \forall\,a\in \Ft,\quad
|\funcm(a) - \mumtmo(a)| \,\leq\,\betath\sigmamtmo(a).
\label{eqn:TmnHmBoundITwo}
\end{align}
The proof mimics that of Lemma~\ref{lem:discreteUCB} using the same conditioning
argument. However, instead of a fixed set over all $t$, we change the set at which we
have confidence based on the discretisation. Similarly we can show that with
probability $>1-\delta/3$ we also have confidence on the decisions $\xt$ at all time
steps. Precisely,
\begin{align}
\forall\,t\geq1,\,\; \forall\,m\in\{1,\dots,M\},\quad
|\funcm(\xt) - \mumtmo(\xt)| \,\leq\,\betath\sigmamtmo(\xt).
\label{eqn:TmnHmBoundIThree}
\end{align}
Using~\eqref{eqn:TmnHmBoundIOne},\eqref{eqn:TmnHmBoundITwo} and 
\eqref{eqn:TmnHmBoundIThree} the following statements hold with probability
$>1-5\delta/6$.
First we can upper bound $\funcopt$ by,
\begin{align}
\funcopt 
\;\leq\; 
\funcm(\xopt) + \zetam
\;\leq\; 
\funcm([\xopt]_t) + \zetam + \frac{1}{t^2}
\;\leq\; 
\ucbmtxx{[\xopt]_t} + \frac{1}{t^2}.
\label{eqn:foptUCBBound}
\end{align}
Since the above holds for all $m$, we have
$\funcopt \leq \ucbtxx{[\xopt]_t} + 1/t^2$.
Now, using similar 
calculations as~\eqref{eqn:discreteInstRegretBound} we bound $\Deltamxt$.
\begin{align*}
\Deltamxt &=\; \funcopt - \funcm(\xt) - \zetam 
% \;\leq\; 
\\ &\leq\;
\ucbtxx{[\xopt]_t} + \frac{1}{t^2} - \funcm(\xt) - \zetam
\;\leq\; 
% \\ &\leq\;
\ucbtxx{\xt} - \funcm(\xt) - \zetam  + \frac{1}{t^2}
\\ &\leq\;
% \;\leq\; 
\ucbmtxx{\xt} - \mumtmo(\xt) +\betath\sigmamtmo(\xt) -\zetam  + \frac{1}{t^2} 
\;\leq\; 2\betath\sigmamtmo(\xt) + \frac{1}{t^2}.\\[-0.4in]
\end{align*}
\hfill\BlackBox

\subsubsection{Proof of Lemma~\ref{lem:compactVarBound}}
\label{sec:compactVarBound}

Since the posterior variance only decreases with more
observations, we can upper bound $\kernel'(x,x)$ for any $x\in A$ by considering its 
posterior variance with only the $s$ observations in $A$.
Further the maximum variance within $A$ occurs if we pick $2$ points
$x_1,x_2$ that are distance $D$ apart and have all observations at $x_1$; then
$x_2$ has the highest posterior variance.
Therefore, we will bound $\kernel'(x,x)$ for any $x\in A$ with $\kernel(x_2,x_2)$ in
the above scenario. Let $\kernelz = \kernel(x,x)$ and $\kernel(x,x') =
\kernelz\phi(\|x-x'\|_2)$, where $\phi(\cdot)\leq 1$ depends on the kernel.
Denote the gram matrix in the scenario described above by
$\Delta = \kernelz \one\one^\top + \eta^2 I$. 
Then using the Sherman-Morrison formula on
the posterior  variance~\eqref{eqn:gpPostEqns}, % can be bounded via,
\begingroup
\allowdisplaybreaks
\begin{align*}
\kernel'(x,x) \;&\leq\; \kernel'(x_2,x_2) \;=\;
  \kernel(x_2,x_2) - \left[\kernel(x_1, x_2)\one\right]^\top \Delta^{-1} 
  \left[\kernel(x_1, x_2)\one\right] \\
&= \kernelz - \kernelz^2\phi^2(D) \one^\top\left[ \kernelz\one\one^\top + 
  \eta^2I\right]^{-1}\one \\
&=\; \kernelz - \kernelz\phi^2(D) \one^\top
\left[   \frac{\kernelz}{\eta^2}I - 
  \frac{\left(\frac{\kernelz}{\eta^2}\right)^2\one\one^\top}
  {1 + \frac{\kernelz}{\eta^2}s } \right]\one 
\\
&=\; \kernelz - \kernelz\phi^2(D) \left(  \frac{\kernelz}{\eta^2}s - 
  \frac{\left(\frac{\kernelz}{\eta^2}\right)^2s^2}
  {1 + \frac{\kernelz}{\eta^2}s } \right) \\ 
&=\; \kernelz - \kernelz\phi^2(D) \frac{s}{\frac{\eta^2}{\kernelz} + s}
\;=\; \frac{1}{1 + \frac{\eta^2}{\kernelz s}}\left(\kernelz - \kernelz\phi^2(D) +
\frac{\eta^2}{s}\right) \\
&\leq\; \kernelz(1-\phi^2(D)) + \frac{\eta^2}{s}.
\end{align*}
\endgroup
For the SE kernel $\phi^2(D) = \exp\left(\frac{-D^2}{2h^2}\right)^2 = 
\exp\left(\frac{-D^2}{h^2}\right) \leq 1-\frac{D^2}{h^2}$. Plugging this into the
bound above retrieves the first result with $\Cse = \kernelz/h^2$.
For the \matern kernel we use a Lipschtiz constant $L_{Mat}$ of $\phi$.
Then $1-\phi^2(D) = (1-\phi(D))(1+\phi(D)) \leq 2(\phi(0)-\phi(D)) \leq 2L_{Mat}D$.
We get the second result with $\Cmat = 2\kernelz L_{Mat}$.
Since the SE kernel decays fast, we get a stronger result on its posterior variance
which translates to a better bound in our theorems.
\hfill\BlackBox

\subsubsection{Proof of Lemma~\ref{lem:TgmDmnBound}}
\label{sec:TgmDmnBound}

First, we will invoke the same discretisation used in the proof of
Lemma~\ref{lem:TmnHmBound} via which we have $\ucbtxx{[\xopt]_t} \geq \funcopt 
- 1/t^2$ ~\eqref{eqn:foptUCBBound}. 
(Therefore, Lemma~\ref{lem:TgmDmnBound} holds
only with probability $>1-\delta/6$, but this event has already been accounted
for in Lemma~\ref{lem:TmnHmBound}.)
Let $\bint = \argmax_{x\in\Ain} \ucbtx$ be the maximiser of the upper
confidence bound in $\Ain$ at time $t$.
Note that the discretisation is fixed ahead of time and $\bint$ is deterministic
given the data $\{(\xt,\mt,\yt)\}_{i=1}^{t-1}$ at time $t$.
Now using the relaxation
$\xt\in\Ain \implies \ucbtxx{\bint} > \ucbtxx{[\xopt]_t} \implies
\ucbmtxx{\bint} > \funcopt - 1/t^2$ and proceeding,
\begingroup
\allowdisplaybreaks
\begin{align*}
&\PP(\TgmnpAin>u) \;\leq\;
\frac{1}{\Gtwoprob}
\PPgp\big( \exists t: u+1\leq t \leq n, \;\;
  \ucbmtxx{\bint} > \funcopt - 1/t^2 \indAnds 
\numberthis \label{eqn:TgmDmnCalculations}
\\ &\hspace{3in} \betath\sigmamtmo(\bint) <\gammam\big) 
\\
&\hspace{0.2in}\leq\;
\frac{1}{\Gtwoprob}
  \sum_{t=u+1}^{n'}
\PPgp\big( \mumtmo(\bint) - \funcm(\bint) > 
  \Deltammxx{m}{\bint}
% \underbrace{\funcopt - \funcm(\bint) - \zetam}_{\Deltammxx{m}{\bint}} 
  - \betath\sigmamtmo(\bint) - 1/t^2
\indAnds 
\\ &\hspace{3in} %\betath\sigmamtmo(\bint) <\gammam\big) 
\betath\sigmamtmo(\bint) < \gammam\big) \\
&\hspace{0.2in}\leq\;
\frac{1}{\Gtwoprob}
  \sum_{t=u+1}^{n'}
\PPgp\big( \mumtmo(\bint) - \funcm(\bint) > 2\betath\sigmamtmo(\bint) -
1/t^2 \big) \\
&\hspace{0.2in}\leq\;
\frac{1}{\Gtwoprob}
  \sum_{t=u+1}^{n'} \PP_{Z\sim\Ncal(0,1)}\left(Z > \betath\right)
\;\leq\;
  \sum_{t=u+1}^{n'} 
\frac{1}{\Gtwoprob}
\frac{1}{2}\exp\left(\frac{-\betat}{2}\right) \\
&\hspace{0.2in}\leq\;
\frac{1}{\Gtwoprob}
  \frac{1}{2}\left(\frac{2\Gtwoprob\delta}{M\pi^2}\right)
  \sum_{t=u+1}^{n'} t^{-2} 
\;\leq\;
\frac{\delta}{M\pi^2}\frac{1}{u} 
% \;\leq\;
% \frac{\delta}{\pi^2} \frac{1}{u^{1+4/\alpha}}.
\end{align*}
\endgroup
In the second step we have rearranged the terms and used the definition of
$\Deltamx$. 
In the third step, as $\Ain\subset\Jcalcmtrg$,
we have $\Deltammxx{m}{\bint} > 3\gammam > 
3\betath\sigmamtmo(\bint)$.
The last step bounds the sum by an integral.
For the fourth step, we have used,
$t>u\geq 1/(3\eta)^{2/3},\,$
$\betat > 2\log(M\pi^2t^2/2\delta) >(3/2)^2$,
and $\sigmamtmo(\bint) > \eta/\sqrt{t}$ to conclude,
\begin{align*}
t > \frac{1}{(3\eta)^{2/3}}
\;&\implies\;
\frac{3t^{3/2}}{2} > \frac{1}{2\eta}
\;\implies\;
 t^{3/2}\betath > \frac{1}{2\eta}
\;\implies\;
 2 \betath \sigmamtmo > \frac{1}{t^2}.
\end{align*}

\vspace{-0.35in}

\hfill\BlackBox

% 
% \subsubsection{A $\polylog(n)$ bound for $\Tmnxx{\Htncalm}$}
% \label{sec:TmnHPolylog}
% \toworkon{To fill in.}

% \subsection{Continuous and Compact $\Xcal$}
% % 
% % \input{appCompactAnalysis2}

\section{Conclusion}
\label{sec:conclusion}

We introduced and studied the multi-fidelity bandit problem under Gaussian
Process assumptions.
%  which builds on our work on multi-fidelity $K$-armed
% bandits~\citep{kandasamy2016mfbandit}.
Our theorems demonstrate that \mfgpucbs explores the space using the
cheap lower fidelities, and uses
the higher fidelity queries on successively smaller regions, hence performing
better than single fidelity strategies.
Via experiments on synthetic functions, three hyper-parameter tuning tasks,
and an astrophysical maximum likelihood estimation problem,
we demonstrate the efficacy of our method and more generally, the
utility of the multi-fidelity framework.
Our Matlab implementation and experiments can be downloaded
from
\incmtt{\small{\href{https://github.com/kirthevasank/mf-gp-ucb}{github.com/kirthevasank/mf-gp-ucb}}}.

Going forward we wish to study multi-fidelity optimisation under different model
assumptions, and extend the algorithm when we have to deal with approximations
from structured fidelity spaces.

% Appendix

\vspace{0.2in}
\appendix
\section*{\Large Appendix}

\vspace{0.2in} \section{Addendum to Experiments}
\label{app:experiments}

\subsection{Some Implementation Details of other Baselines}
\label{app:baselines}

For \mfnaives we limited the number of first fidelity evalutions to
$\max\big(\frac{1}{2}\frac{\COST}{\costone}, 500\big)$ where $\COST$ was the total
budget used in the experiment. The $500$ limit was set to avoid unnecessary
computation -- for all of these problems, $500$ queries are not required to find the
maximum.
While there are other methods for multi-fidelity optimisation (discussed under
Related Work) none of them had made their code available nor were their methods
straightforward to implement - this includes \mfsko.

A straightforward way to incorporate lower fidelity information to \gpucbs and
\gpeis is to share the same kernel parameters.
This way, the kernel $\kernel$ can be learned by by jointly maximising the marginal
likelihood. 
While the idea seems natural, we got mixed results in practice. On some
problems this improved the performance of all GP methods (including 
\mfgpucb), but  on others all performed poorly.
One explanation is that while lower fidelities
approximate function values, they are not always best described by the same kernel.
The results presented do not use lower fidelities to learn $\kernel$ as it
was more robust. For \mfgpucb, each $\kernelm$ was learned
independently using only the queries at fidelity $m$.

In addition to the baselines presented in the figures, we also compared our method to
the following methods.
The first two are single fidelity and the last two are mutlti-fidelity methods.
\begin{itemize}[leftmargin=0.3in]
\vspace{-0.05in}
\item The probability of improvement (\gppi) criterion for BO~\citep{brochu12bo}.
We found that in general either \gpucbs or \gpeis performed better.
\vspace{-0.05in}
\item Querying uniformly at random at the highest fidelity  
and taking the maximum. On all problems this
performed worse than other methods.
\vspace{-0.05in}
\item A variant of \mfnaives where instead of \gpucbs we queried at the first fidelity
uniformly at random. On some problems this did better than querying
with \gpucb, probably since unlike \gpucbs it was not stuck at the maximum of
$\funcone$. However, generally it performed worse.
\vspace{-0.05in}
\item The multi-fidelity method from~\citet{forrester07cokriging} also based 
on GPs. We found that
this method did not perform as desired: in particular,
it barely queried beyond the first fidelity.
\end{itemize}

\subsection{Description of Synthetic Experiments}
\label{app:synthetic}
\label{app:synResults}

The following are the descriptions of the synthetic functions used.
The first three functions and their approximations were taken 
from~\cite{xiong13highAccuracy}.

\begin{itemize}[leftmargin=0.0in]

\item[]
\textbf{Currin exponential function:}
The domain is the two dimensional unit cube $\Xcal = [0,1]^2$.
The second and first fidelity functions are,
\begin{align*}
\functwo(x) &= \left(1-\exp\left(\frac{-1}{2x_2}\right)\right)
  \left(\frac{2300x_1^3 + 1900x_1^2 + 2092x_1 + 60}{100x_1^3 + 
  500x_1^2 + 4x_1 + 20}\right) ,\\
\funcone(x) &= \frac{1}{4}\functwo(x_1 + 0.05, x_2 + 0.05) +
               \frac{1}{4}\functwo(x_1 + 0.05, \max(0, x_2 - 0.05)) + \\
  &\hspace{0.4in}
               \frac{1}{4}\functwo(x_1 - 0.05, x_2 + 0.05) +
               \frac{1}{4}\functwo(x_1 - 0.05, \max(0, x_2 - 0.05)).
\end{align*}

\item[]
\textbf{Park function:}
The domain is $\Xcal = [0,1]^4$.
The second and first fidelity functions are,
\begin{align*}
\functwo(x) &= \frac{x_1}{2}\left( \sqrt{1 + (x_2+x_3^2)\frac{x_4}{x_1^2}} - 1\right)
+ (x_1 + 3x_4) \exp(1 + \sin(x_3)),\\
\funcone(x) &= \left( 1 + \frac{\sin(x_1)}{10}\right) \functwo(x) 
  - 2x_1^2 + x_2^2 + x_3^2 + 0.5.
\end{align*}

\item[]
\textbf{Borehole function:}
The second and first fidelity functions are,
\begin{align*}
\functwo(x) &= \frac{2\pi x_3(x_4-x_6)}
  {\log(x_2/x_1) \left(1 + \frac{2x_7x_3}{\log(x_2/x_1)x_1^2x_8} +
    \frac{x_3}{x_5} \right)}, \\
\funcone(x) &= \frac{5 x_3(x_4-x_6)}
  {\log(x_2/x_1) \left(1.5 + \frac{2x_7x_3}{\log(x_2/x_1)x_1^2x_8} +
    \frac{x_3}{x_5} \right)}. 
\end{align*}
The domain of the function is $[0.05, 0.15; 100, 50K; 63.07K, 115.6K;$
$990, 1110; 63.1, 116; 700, 820;$ $1120, 1680; 9855, 12045]$.
We first linearly  transform the variables to lie in $[0,1]^8$.

\item[]
\textbf{Hartmann-3D function:}
The $M$\ssth fidelity function is
$\funcM(x) = \sum_{i=1}^4 \alpha_i \exp\big( -\sum_{j=1}^3 A_{ij}
(x_j-P_{ij})^2\big)$ where $A, P\in\RR^{4\times 3}$ are fixed matrices given below and
$\alpha = [1.0, 1.2, 3.0, 3.2]$. For the lower fidelities we use the same form except
changing $\alpha$ to $\alpha^{(m)} = \alpha + (M-m)\delta$ where $\delta = [0.01, -0.01,
-0.1, 0.1]$ and $M=3$. The domain is $\Xcal = [0,1]^3$.
\[
A = 
\begin{bmatrix}
3 & 10 & 30 \\
0.1 & 10 & 35 \\
3 & 10 & 30 \\
0.1 & 10 & 35 
\end{bmatrix},
\quad
P = 10^{-4} \times
\begin{bmatrix}
3689 & 1170 & 2673 \\
4699 & 4387 & 7470 \\
1091 & 8732 & 5547 \\
381 & 5743 & 8828
\end{bmatrix}
\]

\item[]
\textbf{Hartmann-6D function:}
The $6$-D Hartmann function takes the same form as the $3$-D case except $A,P\in\RR^{4\times 6}$
are as given below.
We use the same modifications as above to obtain the lower fidelities using $M=4$.
\[
A = 
\begin{bmatrix}
10 & 3 & 17 & 3.5 & 1.7 & 8 \\
0.05 & 10 & 17 & 0.1 & 8 & 14 \\
3 & 3.5 & 1.7 & 10 & 17 & 8 \\
17 & 8 & 0.05 & 10 & 0.1 & 14 
\end{bmatrix},
\;\;
P = 10^{-4} \times
\begin{bmatrix}
1312 & 1696 & 5569 &  124 & 8283 & 5886 \\
2329 & 4135 & 8307 & 3736 & 1004 & 9991 \\
2348 & 1451 & 3522 & 2883 & 3047 & 6650 \\
4047 & 8828 & 8732 & 5743 & 1091 &  381 \\
\end{bmatrix}
\]

\end{itemize}

\section{Other Material}
\label{app:ancillary}

\subsection{Some Ancillary Results}
\label{app:ancillaryResults}

The following results were used in our analysis. The first is a standard
Gaussian concentration result and the second is an expression for the Information
Gain in a GP from~\citet{srinivas10gpbandits}.
\insertprespacing

\begin{lemma}[Gaussian Concentration]
\label{lem:gaussConcentration}
Let $Z\sim\Ncal(0,1)$. Then $\;\PP(Z>\epsilon) \leq \frac{1}{2}\exp(-\epsilon^2/2)$.
\end{lemma}
\insertprespacing

\begin{lemma}[Mutual Information in GP, \normalfont{\cite{srinivas10gpbandits}, Lemma 5.3}]
\label{lem:IGformula}
Let $\func\sim\GP(\zero,\kernel)$, $f:\Xcal\rightarrow\RR$ and we observe
$y=f(x) + \epsilon$ where $\epsilon\sim\Ncal(0,\eta^2)$. Let $A$ be a finite 
subset of $\Xcal$ and $f_A,y_A$ be the function values and observations on this set
respectively.
Using the basic Gaussian properties it can be shown that the mutual information
$I(y_A;f_A)$ is,
\[
I(y_A; \func_A) = \frac{1}{2} \sum_{t=1}^{n} \log(1 + \eta^{-2}
  \sigma^2_{t-1}(x_t) ).
\]
where $\sigma^2_{t-1}$ is the posterior GP variance after observing the first $t-1$
points. 
\end{lemma}

\vspace{0.1in}

\vspace{-0.15in}
\subsection{A Table of Notations and Abbreviations}
\label{app:notation}
\vspace{-0.05in}
The following table summarises the notation and abbreviations used in the manuscript.
The table continues to multiple pages.

\newcommand{\notTableEntry}[2]{#1 & #2 \\ \hline}
\newcommand{\tableExplWidth}{4.90in}

% \noindent
% \textbf{Notations}
% \begin{center}
\begin{longtable}{|c|p{\tableExplWidth}|}
\hline

\notTableEntry{\textbf{Notation}}{\textbf{Description}}

\notTableEntry{$\EEgp, \PPgp$}{Expectations and probabilities when
$\funcone,\dots,\funcM$ are sampled from $\GP(0,\kernel)$.}

\notTableEntry{$\EE, \PP$}{Expectations and probabilities under the prior, which
includes condition~\Gtwos after
$\funcone,\dots,\funcM$ are sampled from $\GP(0,\kernel)$.}

\notTableEntry{$\Gtwoprob$}{A lower bound on the probability that condition
\Gtwos holds when $\funcone,\dots,\funcM$ are sampled,
see~\eqref{eqn:gtwoprob}.}

\notTableEntry{$\gpsupbound$}{The function which controls the probability on 
the supremum of a GP, see Assumption~\ref{asm:supgp}.}

\notTableEntry{$M$}{The number of fidelities.}

\notTableEntry{$\func,\funcm$}{The payoff function and its $m$\superscript{th} 
fidelity approximation. $\funcM =\func$.}

\notTableEntry{$\COST$}{$\COST$ typically denotes the capital of some resource which
is expended upon each evaluation of at any fidelity.}

\notTableEntry{$\costm$}{The cost, i.e. amount of capital expended,
of querying at fidelity $m$.}

\notTableEntry{$N$}{The random number of queries at any fidelity within capital $\COST$.
\newline $N = \max\{n\geq 1: \sum_{t=1}^n \costmt \leq \COST\}$}

\notTableEntry{$\Xcal$}{The domain over which we are optimising $\func$.}

\notTableEntry{$\xopt,\funcopt$}{The optimum point and value of the 
$M$\superscript{th} fidelity function.}

\notTableEntry{$\complement{A}$\vphantom{$\Hcaldownm$}}
{The complement of a set $A\subset\Xcal$. 
$\complement{A} = \Xcal\backslash A$.}

\notTableEntry{$|A|$}{The cardinality of a set $A\subset\Xcal$ if it is countable.}

\notTableEntry{$\vee, \wedge$}{Logical \emph{Or} and \emph{And} respectively.}

\notTableEntry{$\lesssim, \gtrsim,\asymp$}{Inequalities and equality ignoring 
constant terms.}

\notTableEntry{$\qt, \rt$}{The instantaneous reward and regret respectively.
\newline
$\qt =\funcM(\xt)$ if $\mt=M$ and $-\infty$ if $\mt\neq M$.  $\rt = \funcopt - \qt$. 
% $\rt = \funcopt + B$ if $\mt\neq M$ and $\rt = \funcopt - \funcM(\xt)$ if $\mt= M$.
}
% See Equation~\eqref{eqn:instRegret}}

% \notTableEntry{$R(\COST)$}{The cumulative regret after spending capital $\COST$.
% See equation~\eqref{eqn:cumRegDefn}.}

\notTableEntry{$S(\COST)$}{The simple regret after spending capital $\COST$.
$S(\COST) = \funcopt - \min_{t=1,\dots,N}\func(\xt)$.}

\notTableEntry{$\zetam$}{The bound on the maximum difference between $\funcm$ and
$\funcM$,\newline $\|\funcM-\funcm\|_\infty\leq\zetam$.}

\notTableEntry{$\mumt$}{The mean of the $m$\superscript{th} fidelity GP $\funcm$ 
conditioned on $\Dcalmt$ at time $t$.}

\notTableEntry{$\kernelmt$}{The covariance of the $m$\superscript{th} fidelity GP 
$\funcm$  conditioned on $\Dcalmt$ at time $t$.}

\notTableEntry{$\sigmamt$}{The standard deviatiation of the $m$\superscript{th} 
fidelity GP $\funcm$ conditioned on $\Dcalmt$ \newline at time $t$.}

% \notTableEntry{$\mumt,\kernelmt,\sigmamt$}{The posterior mean, covariance and
% standard deviation respectively of the $m$\superscript{th} fidelity at time $t$.}

\notTableEntry{$\xt,\yt$}{The queried point and observation at time $t$.}

\notTableEntry{$\mt$}{The queried fidelity at time $t$.}

\notTableEntry{$\Dcalmn$}{The set of queries at the $m$\superscript{th} fidelity 
until time $n$ $\{(\xt,\yt)\}_{t:\mt=m}$.}

\notTableEntry{$\betat$}{
The coefficient trading off exploration and exploitation in the UCB.
\newline
See Theorems~\ref{thm:discreteThm} and ~\ref{thm:compactThm}.
}

\notTableEntry{$\ucbmtx$}{The upper confidence bound (UCB) provided by the
$m$\superscript{th} fidelity on $\funcM(x)$.  \newline
$\ucbmtx = \mumtmo(x) + \betath\sigmamtmo(x) + \zetam$.}

\notTableEntry{$\ucbtx$}{The combined UCB provided by all
fidelities on $\funcM(x)$. 
\newline $\ucbtx=\min_m \ucbmtx$.}

\notTableEntry{$\gammam$}{The parameter in \mfgpucbs for switching from the $m$\ssth
fidelity to the $(m+1)$\superscript{th}.\hspace{-0.1in}
}

\notTableEntry{$\Rtilden$ \vphantom{$\Hcaldownm$}}
{The $M$\ssth fidelity cumulative regret after $n$ rounds. See~\eqref{eqn:Rtilden}}
% ${\small \Rtilden}$ $= \sum_{t=1}^n\costmt\rt$. 
%\newline
% $\rt = \funcopt + B$ if $\mt\neq M$ and $\rt = \funcopt - \funcM(\xt)$ if $\mt= M$.

\notTableEntry{$\Tmnxx{A}$}{The number of queries at fidelity $m$ in
subset $A\subset\Xcal$ until time $n$. }

\notTableEntry{$\Tgmnxx{A}$}{Number of queries at fidelities greater than $m$ 
in any subset  $A\subset\Xcal$ until time $n$. }

% \end{longtable}
% 
% \begin{longtable}{|c|p{\tableExplWidth}|}
% \hline

\notTableEntry{$\nCOST$}{
$\nCOST=\floor{\COST/\costM}$.
Number of plays by a strategy querying only  at 
fidelity $M$ within capital $\COST$; also a lower bound on $N$, the number of plays
by a multi-fidelity strategy.
% \newline
}

\notTableEntry{$\nCOSTtop$}{An upper bound on $N$, the number of plays
by a multi-fidelity strategy within capital $\COST$.
% \newline
$\nCOSTtop=\floor{\COST/\costone}$.}

\notTableEntry{$\IG_n(A)$}{The maximum information gain of a set $A\subset\Xcal$
after $n$ queries in $A$. See Definition~\ref{def:infGain}.}

\notTableEntry{$\Deltamx$}{$\Deltamx = \funcopt - \funcm - \zetam$.}

\notTableEntry{$\Jcalme$}{The points in $\Xcal$ whose $\funcm$ value is within
$\zetam + \eta$ of the optimum $\funcopt$.
\newline
$\Jcalme = \{x\in\Xcal; \Deltamx\leq \eta\}$
.}

\notTableEntry{$\Hcalm$}{$(\Hcalm)_{m=1}^M$ 
is a partitioning of $\Xcal$. See
Equation~\eqref{eqn:HcalDefn}.
The analysis of \mfgpucbs hinges on these partitioning.
}

\notTableEntry{$\Hcalupm, \Hcaldownm$}{
The arms ``above"/``below" $\Hcalm$.
$\Hcalupm = \bigcup_{\ell=m+1}^M\Hcall$, $\;\Hcaldownm = \bigcup_{\ell=1}^{m-1}\Hcall$.
% \hspace{-0.1in}
}

% \notTableEntry{$\Htcalm$}{$(\Htcalm)_{m=1}^M$ 
% is a partitioning of $\Xcal$ in the continuous and compact.
% See Equation~\eqref{eqn:HtcalmDefn}.
% The analysis of \mfgpucbs hinges on these partitioning.
% }

\notTableEntry{$\Htncalm$}{
An $n$-dependent dilation of $\Htcalm$ in the continuous setting.
See Section~\ref{sec:analysisCts}.
}

\notTableEntry{$\Htcalupm, \Htcaldownm$}{
The arms ``above"/``below" $\Htcalm$.
$\Htcalupm = \bigcup_{\ell=m+1}^M\Htcall$, $\;\Htcaldownm = \bigcup_{\ell=1}^{m-1}\Htcall$.
\hspace{-0.1in}
}

\notTableEntry{$\Xcalg$}{The good set for $M=2$ fidelity
problems. $\;\Xcalg = \{x\in\Xcal; \funcopt-\funcone(x) \leq \zetaone\}$.}
% 
% \notTableEntry{$\Xcalrg,\Xcalrb$ \vphantom{$\Hcaldownm$}}
% {The inflations of $\Xcalg,\Xcalb$ in the discrete case. \newline 
% $\Xcalrg = \{x; \funcopt - \funcone(x) \leq \zetaone + \rho\gamma\} = \Hcaltwo$ 
% and $\Xcalrb = \Xcal\backslash\Xcalrg =  \Hcalone$.}

\notTableEntry{$\Xcalrg$ \vphantom{$\Hcaldownm$}}
{The inflated good set for \mfgpucb.
% \newline
$\;\Xcalrg = \{x; \funcopt - \funcone(x) \leq \zetaone +
3\gamma\}$.}

% \notTableEntry{$\Xcalrtang$}{The additional $n$-dependent inflation of $\Xcalrtg$.
% $\Xcalrtang = \Htncalone$.}

\notTableEntry{$\Omegaii{\varepsilon}(A)$}{The $\varepsilon$--covering number of a subset 
$A\subset\Xcal$ in the $\|\cdot\|_2$ metric.}

\notTableEntry{$\COSTone, \COSTtwo$}{The minimum capitals that need to be expended before
the bound on $S(\COST)$ hold in Theorems~\ref{thm:discreteThm} and~\ref{thm:compactThm}.}

& \\[-0.17in]
\hline

\notTableEntry{\textbf{Abbreviation}}{\textbf{Description}}

\notTableEntry{UCB}{Upper Confidence Bound}
\notTableEntry{BO}{Bayesian Optimisation}
\notTableEntry{\gpucb}{Gaussian Process Upper Confidence Bound~\citep{srinivas10gpbandits}}
\notTableEntry{\mfgpucb}{Multi-fidelity Gaussian Process Upper Confidence Bound}
\notTableEntry{\gpei}{(Gaussian Process) Expected Improvement~\citep{jones98expensive}}
\notTableEntry{\mfsko}{Multi-fidelity Sequential Kriging
Optimisation~\citep{huang06mfKriging}}
\notTableEntry{\mfnaive}{Naive multi-fidelity method described in
Section~\ref{sec:experiments}.}
\notTableEntry{\direct}{DIviding RECTangles~\citep{jones93direct}}
\notTableEntry{SE}{Squared Exponential (in reference to the kernel)}

\end{longtable}
% \end{center}

% \noindent
% \textbf{Abbreviations}
% % \begin{center}
% \begin{longtable}{|c|p{\tableExplWidth}|}
% \hline
% 
% \notTableEntry{\textbf{Notation}}{\textbf{Description}}
% 
% \notTableEntry{$M$}{The number of fidelities.}
% 
% \end{longtable}

% Acknowledgements should go at the end, before appendices and references
\acks{%
We wish to thank Bharath Sriperumbudur for the helpful email discussions.
This research is partly funded by DOE grant DESC0011114, NSF grant
IIS1563887, and the Darpa D3M program.
KK is supported by a Facebook fellowship and a Siebel scholarship.
This work was done when JO and GD were at Carnegie Mellon University.
}

\bibliography{kky,lai}
\bibliographystyle{abbrvnat}

\end{document}